%% file: main.tex
\documentclass[10pt, conference]{IEEEtran}

\usepackage{amsmath,amssymb,amsfonts}
\usepackage{graphicx}
\usepackage{pifont}

\usepackage{amsthm}
\usepackage{xcolor}
\usepackage[figuresright]{rotating}

\usepackage{graphicx}
\graphicspath{{/}{fig/}}

\usepackage{array}
\usepackage{textcomp}
\usepackage{xcolor}
\usepackage{multirow}
\usepackage{booktabs}

\usepackage{mathtools}
\usepackage{breqn}
\usepackage{float}

\usepackage{pgfplots}
\pgfplotsset{compat=1.7}
\usepgfplotslibrary{groupplots}

\usepackage[style=base]{caption}
\usepackage{subcaption}

\usepackage{multirow,tabularx}
\usepackage{hyperref}
\usepackage[ancient]{flushend}
\usepackage{algorithmic}
\usepackage[vlined, ruled, shortend]{algorithm2e}

\usepackage[capitalise]{cleveref}
\usepackage{svg}

\newlength\figureheight
\newlength\figurewidth
\SetAlCapNameFnt{\footnotesize}
\SetAlCapFnt{\footnotesize}

\DeclareMathOperator*{\argmin}{argmin}

\SetKwInput{kwInit}{Initilization}

\title{
    \huge
    Towards Large-Scale Relative Localization in Multi-Robot Systems with Dynamic UWB Role Allocation
}

\author{
    \IEEEauthorblockN{
        \vspace{1em}
        Paola Torrico Mor\'on\IEEEauthorrefmark{4},
        Jorge Pe\~na Queralta\IEEEauthorrefmark{4},
        Tomi Westerlund\IEEEauthorrefmark{4}
    }
    \IEEEauthorblockA{
        \normalsize
        \IEEEauthorrefmark{4}\href{https://tiers.utu.fi}{Turku Intelligent Embedded and Robotic Systems (TIERS) Lab, University of Turku, Finland}.\\
        Emails: \textsuperscript{1}\{pctomo, jopequ, tovewe\}@utu.fi\\[+6pt]
    }
}

\begin{document}

\maketitle
\thispagestyle{empty}
\pagestyle{empty}

\input{sec/00_Abstract.tex}
\IEEEpeerreviewmaketitle

\input{sec/01_Intro}

\input{sec/02_RelatedWorks}
\input{sec/03_ProblemDefinition}

\input{sec/04_Methodology}


\input{sec/05_Experiments}
\input{sec/06_Conclusion}


\section*{Acknowledgment}

This research work is supported by the R3Swarms project funded by the Secure Systems Research Center (SSRC), Technology Innovation Institute (TII). 

\bibliographystyle{unsrt}
\bibliography{bibliography}

\end{document}

%% file: sec/00_Abstract.tex

\begin{abstract}%
    \label{sec:abstract}%
    Ultra-wideband (UWB) ranging has emerged as a key radio technology for robot positioning and relative localization in multi-robot systems. Multiple works are now advancing towards more scalable systems, but challenges still remain. This paper proposes a novel approach to relative localization in multi-robot systems where the roles of the UWB nodes are dynamically allocated between active nodes (using time-of-flight for ranging estimation to other active nodes) and passive nodes (using time-difference-of-arrival for estimating range differences with respect to pairs of active nodes). We adaptively update UWB roles based on the location of the robots with respect to the convex envelope defined by active nodes, and introducing constraints in the form of localization frequency and accuracy requirements. We demonstrate the applicability of the proposed approach and show that the localization errors remain comparable to fixed-role systems. Then, we show how the navigation of an autonomous drone is affected by the changes in the localization system, obtaining significantly better trajectory tracking accuracy than when relying in passive localization only. Our results pave the way for UWB-based localization in large-scale multi-robot deployments, for either relative positioning or for applications in GNSS-denied environments. 
\end{abstract}

\begin{IEEEkeywords}

    UWB; ToF; TDoA; Localization; UWB Ranging

\end{IEEEkeywords}

%% file: sec/01_Intro.tex

\section{Introduction}\label{sec:introduction}
Autonomous mobile robots have been penetrating multiple industries and domains of our society. At the same time, their connectivity has been increasing and networked collaborative systems have gained importance within the field~\cite{hayat2016survey}. From construction~\cite{albeaino2021trends} to warehouses~\cite{plaksina2018development} to mining, the ability for robots to operate without a user input has become more important~\cite{yang2018grand}. In order to achieve a high level of autonomy and situational awareness, localization is one of the first problems to be solved~\cite{queralta2020collaborative}. Outdoors, GNSS sensors are widely used~\cite{stempfhuber2011precise}, but multipath propagation and other inherent limitations make this solution unreliable indoors~\cite{qi2020cooperative, chen2016network}. Localization in GNSS-denied environments often relies on onboard sensors such as IMUs~\cite{qi2020cooperative}, or odometry estimations from lidars~\cite{paneque2019multisensor, li2020multi} or cameras~\cite{qi2020cooperative, xu2020decentralized}. Onboard methods often have their own limitations in terms of long-term drift, or are affected by the environmental conditions, e.g., low visibility for cameras or lack of structure for lidars~\cite{schroeer2018realtime}.

In controlled deployments such as industrial environments, infrastructure can aid in localizing robots. Among the different approaches, radio technologies have gained momentum in recent years~\cite{xianjia2021applications}. Wi-Fi, Bluetooth ultra-wideband (UWB) technology all can aid in positioning by ranging between signal emitters and receivers, and often also including calculations regarding the angle of arrival of the signal~\cite{karaagac2017evaluation, khan2018angle}. UWB systems provide more accuracy than both Wi-Fi and Bluetooth~\cite{grosswindhager2019snaploc}, and a greater range than Bluetooth~\cite{chen2016network}. UWB technology 
features more immunity to multipath fading, better interference mitigation and improved timing~\cite{grosswindhager2019snaploc}. It also provides accurate enough localization indoor for mobile robots, including aerial vehicles, at a significantly lower cost than, e.g., motion capture (MOCAP) systems~\cite{queralta2020uwb}. Ranging in UWB systems is often based on either time-of-flight (ToF), with an active measurement between a given pair of nodes, or time difference of arrival (TDoA), with the possibility of the node being a passive listener~\cite{shule2020uwb}.

Albeit the many mobile robotic systems in the literature relying on UWB for localization~\cite{grosswindhager2019snaploc, heydariaan2020anguloc}, there are still a number of limitations in terms of deployment flexibility and scalability of these systems. First, a set of stationary nodes, or anchors, at known positions is often used as the basis for the localization of mobile nodes, or tags~\cite{shule2020uwb}. Only recently, there have been studies on the mobility of anchors~\cite{almansa2020autocalibration}, and on the design of multi-robot systems that are entirely mobile~\cite{xu2020decentralized, xu2021omni}. However, these still rely on ToF measurements limiting the scalability of he system, as node-to-node transmissions need to be scheduled within a time window defined by the desired ranging or localization frequency. Scalable TDoA-based systems, e.g., the Crazyflies' TDoA positioning system~\cite{zhao2021learning}, have the core limitations of relying in mostly fixed anchors and, independently on whether the anchors are fixed or not, becoming unstable and highly noisy as the mobile nodes move outside the convex envelope defined by the anchor positions. 

\begin{figure*}
    \centering
    \includegraphics[width=0.9\textwidth]{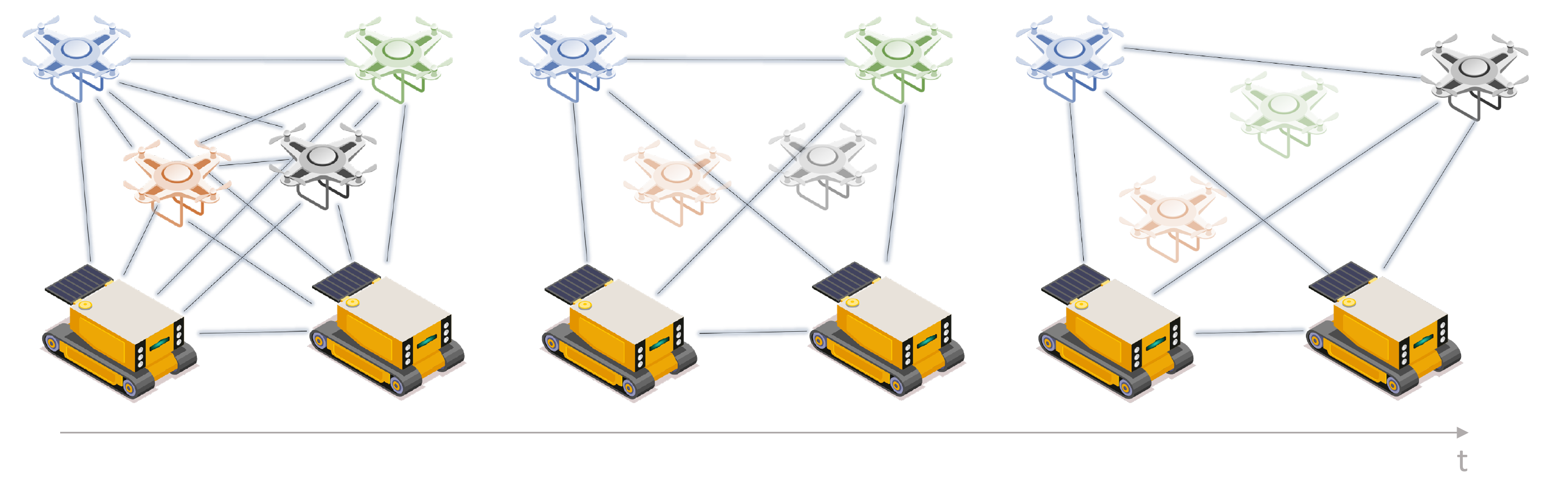}
    \caption{Cooperative localization approach based on UWB ranging measurements from multiple transceivers in different robots. The first step is to initialize all robot positions with all-to-all TOF ranging. The second step to select the active nodes and the passive listener nodes. Finally update the role of the nodes if the robot configuration changes}
    \label{fig:steps_solution}
    \vspace{-1em}
\end{figure*}

Motivated by the scalability limitation of UWB ranging in multi-robot systems, we approach the problem by relying on both ToF and TDoA ranging but by dynamically selecting the nodes that are actively using ToF for ranging. This approach allows us to exploit the high scalability of passive TDoA localization, where the main limitation is the physical space available within the ToF nodes. In the rest of this document, we refer to the nodes performing ToF ranging estimations as active nodes or anchor nodes, while the nodes localizing themselves using TDoA are referred to as passive nodes, or listener nodes. The proposed approach relies on UWB nodes that are programmed with custom firmware to act as either active or passive nodes, as well as in a backbone network of ROS 2 nodes in charge of the relative localization estimators, dynamic role allocation and, in general, data sharing within the robots. We decide on the role of each node by minimizing the global TDoA error based on the position of the nodes. In our experiments, we also utilize VIO estimations that are combined with the UWB positioning information for the navigation of autonomous robots. A conceptual illustration of this process is shown in Fig.~\ref{fig:steps_solution}

In summary, the core contribution of this paper is the design and implementation of a novel approach to scalable UWB-based localization for multi-robot systems. In relation to the state of the art, this is, to the best of our knowledge, the first one to approach the scalability problem in UWB localization by dynamically switching between ranging modalities. We demonstrate the usability of the method and its improved performance when compared to a standard TDoA approach with fixed anchors. We then show that the dynamic role allocation performs well based on ToF and TDoA ranging, comparing with a baseline that relies on an external MOCAP system. Our experiments show that ground a team of ground and aerial robots can perform accurate real-time relative localization and navigation with the proposed approach.

The remainder of this document is organized as follows. Section II introduces different ranging approaches utilizing UWB. Section III then describes the cooperative localization approach. In Section IV we introduce the methodology for simulations and experiments, along with results. Section V concludes the work and outlines future research directions.

%% file: sec/02_RelatedWorks.tex

\section{Background} \label{sec:related_work}

There is a clear trend in adopting UWB-based solutions for localizing mobile robots in GNSS-denied environments, with a significant increase in contributions to the existing literature in recent years~\cite{shule2020uwb}. 

Radio-based ranging and localization is not a new approach in mobile robotics. Nonetheless, the high accuracy of UWB compared to traditional Wi-Fi or Bluetooth approaches is enabling the wider adoption, together with lower interference in the bands it operates and enhanced robustness against multipath transmission~\cite{chen2016network}.

For localization purposes, different modalities of ranging exist. These modalities are common to the majority of wireless ranging systems~\cite{xianjia2021applications}. The two most used ones are ToF, also known as ToA, and TDoA or hyperbolic positioning~\cite{hamer2018selfcalibrating}. Both of these ranging modalities are widely used in research and in the industry, each one with its benefits and draw backs. Another modality is the Angle of Attack (AoA)~\cite{gao2009particle}, thought it is rarely mentioned and utilized for localization. A technique often employed by Bluetooth is RSSI~\cite{chen2016network}\cite{gao2009particle}. Here it is worth mentioning that Bluetooth can be used for positioning for short transmission distances of around a few meters~\cite{chen2016network}, while UWB can be utilized according to~\cite{shule2020uwb} up to  60m, and to 150m~\cite{guo2019ultrawideband}.

\subsection{ToF and TDoA}

Ranging estimation based on ToF approaches rely on measurements of the amount of time the signal takes from one node to another, i.e., the time of flight of the signal~\cite{qi2020cooperative}. ToF measurements are performed one to one between two devices, A and B. This is also called two-way ranging (TWR) which can be single-sided (SS-TWR) or double-sided (DS-TWR)~\cite{xianjia2021applications}. In short, in SS-TWR a device A initiates the transmission message. When B receives it, it immediately responds, returning the message with the internal delay of the device processing of the signal. Device A can then compute the round trip of the signal and with the delay of B the time of flight. On the other hand, in DS-TWR, both devices take turns initiating the communication. It is equivalent to two SS-TWR exchanges~\cite{shule2020uwb}. To obtain the position of a mobile node using ToF there needs to be at leas three anchor nodes to range to~\cite{guo2017ultrawideband}. Once the distance to each node is computed, the positioning of the node can be computed via multilateration.

The passive TDoA technique measures the difference in the propagation time between a pair of UWB transmitting nodes and one or more receivers~\cite{shule2020uwb}. 
Nodes near active transmitters performing ToF ranging passively intercept the messages and then are able to compute the difference of distance to a pair of nodes and overall its position. It is worth noting that this is not the only approach to TDoA-based localization, as synchronized anchors could act as receivers to a mobile node transmitting signals, without the need for responses. In this case, however, the location information is available at the anchor system only, unless transmitted to the tag.

\subsubsection{System scalability}

From the two different ranging modalities presented above, ToF lacks scalability owing to time scheduling constraints. The messages in the TWR approach need to be schedule among all the different transmitters and receivers to avoid collisions, and with a limited time bandwidth, each new node added to the system, decreases the communication frequency~\cite{chugunov2020toa}. Passive TDoA does not present this limitation. Passive nodes intercepting messages for their localization do not actively participate in communication, meaning that more can be added without decreasing the overall frequency of the system~\cite{vecchia2019talla}. 

Scalability can be considered a key parameter when a system is deployed in larger areas or for a large number of nodes~\cite{xianjia2021applications, queralta2021towards}. When talking about real time systems the rate of update in the nodes is very important, but the number of nodes in the system will limit the frequency of communication~\cite{corbalan2018concurrent}. With more UWB nodes in the system, if using ToF, the system will update less frequently having a negative impact in the real time operation~\cite{heydariaan2020anguloc}. Passive TDoA might provide a solution, since the listener nodes do not participate in the communication and thus do not need any time allocation~\cite{vecchia2019talla}. However, the drawbacks of this method is the need for clock synchronization~\cite{grosswindhager2019snaploc}  and the positioning error dramatically increase when leaving the convex envelope of the anchors~\cite{jansch2020robust}.

Another alternative solution with regards to scalability is concurrent ranging~\cite{xianjia2021applications}. A relevant solution in the literature is SnapLoc~\cite{grosswindhager2019snaploc}. The main idea of this solution is a localization system with nodes that can localize with very high update rates. The higher update rates are achieved by using concurrent ranging. Since there exits a previous knowledge of the anchor positions, the tag sends a broadcast message, each anchor then responds with individually delay in the nanoseconds range, avoiding then misclasification of the messages due to overlap when all the anchors respond to the tag~\cite{grosswindhager2019snaploc}. The node can then perform TDoA. Due to the need of fixed anchors in known positions, the system is not directly applicable to mobile deployments.

\subsection{UWB in Mobile Robots}

Multiple industries  utilize or are starting to implement autonomous mobile robots. Warehouse management, mining, or packet deliveries, among others. These applications might not provide a controlled environment for a system based on fixed anchors. Onboard odometry approaches like the use of lidar and VIO, might in time cause drifting and the integration of UWB can aid in correcting it and reducing long-term drift~\cite{xianjia2021applications, xianjia2021cooperative}.

In the field of UAVs, the centimeter-level accuracy that UWB can provide makes it a cost effective replacement for MOCAP systems and other higher-end solutions, when the accuracy is enough~\cite{guo2016ultrawideband}. A typical application for UWB to be used in UAVs is ground to air ranging~\cite{khawaja2019uwb}. Similar techniques can be used also in a swarm of robots~\cite{li2018accurate}.

Multiple approaches have implemented sensor fusion with UWB. The use of the Extended Kalman filter~\cite{liu2017cooperative}, Monte Carlo filter~\cite{paneque2019multisensor}, moving horizon estimator~\cite{pfeiffer2021computationally} and even particle filters~\cite{qi2020cooperative} are some of the examples. There is a clear trend in integrating visual sensors and especifically VIO estimators~\cite{xianjia2021applications}.

\begin{figure}
    \centering
    \includegraphics[width=0.48\textwidth]{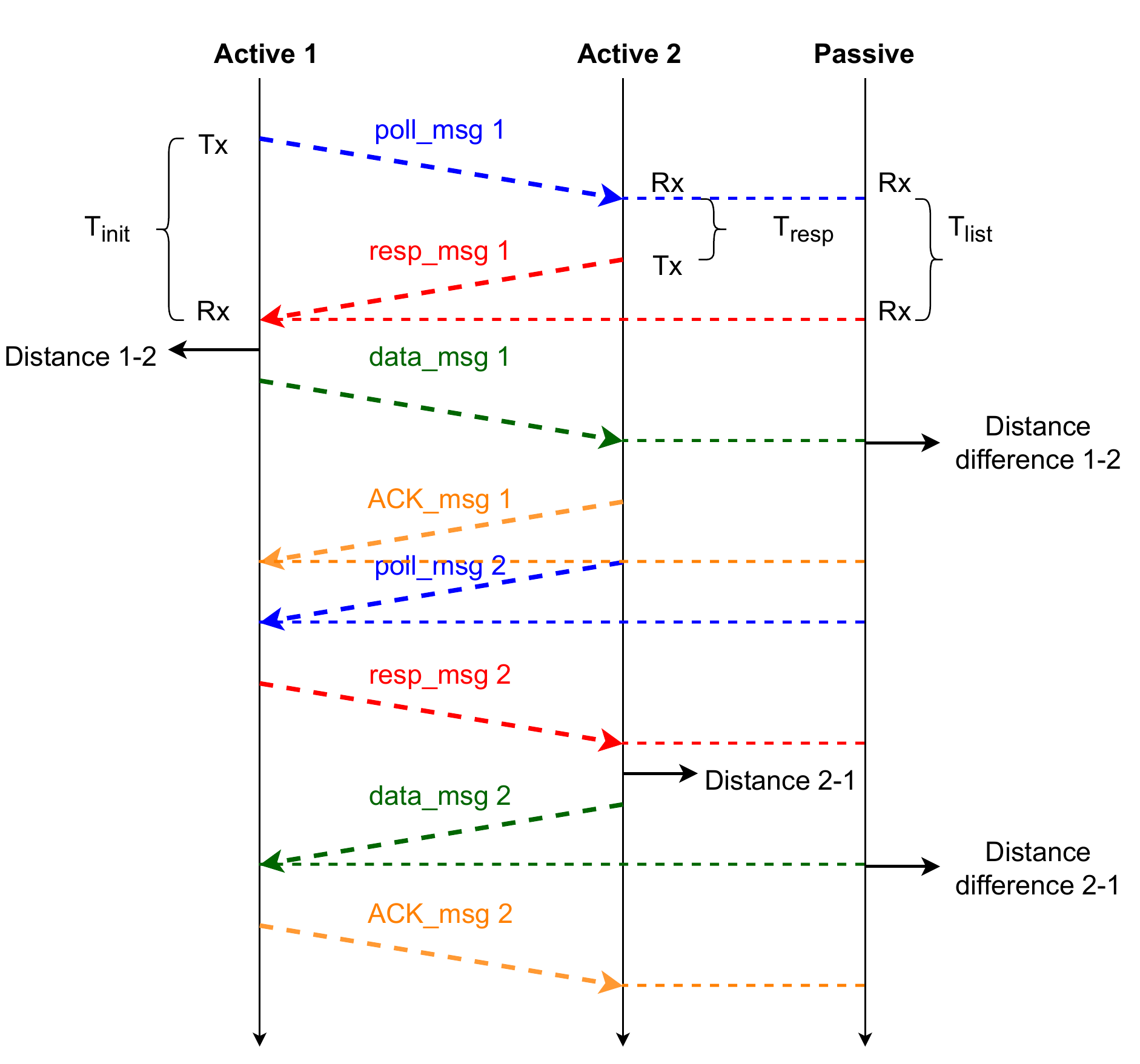}
    \caption{Implementation of the ranging scheme for the UWB nodes, which are driven via ROS nodes from the computer onboard for robots for reconfiguration between active and passive nodes.}
    \label{fig:uwb_implementation}
\end{figure}

%% file: sec/03_ProblemDefinition.tex

\section{Scalable UWB localization}

The majority of the systems implementing localization solutions using UWB are systems that have fixed nodes (anchors) and mobile nodes (tags). These type of solutions do not scale well, normally focusing in the accuracy of the measurements~\cite{grosswindhager2019snaploc}. The two main types of ranging used for UWB are ToF and TDoA, with the majority of them utilizing ToF, with tags being heavily involved in communication. ToF based systems are not very scalable because adding more tags to the system the frequency of the communication decreases, and thus the frequency of the localization for the overall system decreases as well. TDoA systems present a more scalable solution. They normally rely in synchronized anchors that communicate with mobile tags and compute their position. Both types of implementations require the allocation of specific time slots for node communication, making the addition of more nodes into the system, as mention before, reduce the frequency of localization in the entire system. In addition, both types of systems have spacial constraints due to the utilization of fixed anchors.

To create UWB based systems that can easily scale and also have all its nodes be able to mobilize, we propose the following solution. A system that will combine ToF and TDoA ranging approaches, having nodes being able to change the type of ranging they are performing based on its localization relative to the system, minimizing the overall error of the system. This solution will thus address the scalability and mobility problems presented before.

The proposed system will have a set number of nodes that are going to act as active nodes for the network. These active nodes will be performing TWR with ToF. The nodes are then going to localize themselves using multilateration and have a relative position with regards to a selected node in the system. These nodes will be able to obtain the distances between each other without synchronized clocks, and more importantly, to provide a solution for a multi-robot system, to be able to have all nodes move and calibrate on the fly. This part of the solution however does not solve the scalability problem.

To address the scalability of the system, the remaining nodes of the system will utilize TDoA to localize themselves. This set of nodes are from now going be referred as listeners due to the fact that implementing passive TDoA localization they will just receive the messages from the active nodes and not transmit any of their own. The TDoA solution implemented will not require to have synchronized nodes, either for the anchors or the listeners, having all the necessary information for the localization solution being passed on the messages from the active nodes. With this implementation, the addition of listeners can technically have no limitation, while in practically depends on computing power present on the distributed system. Listeners can be added to the network and receive the messages from active nodes and have the necessary information to localize themselves with respect to the reference frame being used.

For the solution, the UWB nodes used in the network can be incorporated either in terrestrial or aerial robots. The example in \cref{fig:steps_solution} assumes one node per robot. For the initialization of the network ToF is used, having all-to-all communication between each pair of nodes. Using multilateration, the position of all nodes is computed with the following assumptions:

\begin{itemize}
    \item The first active node, Active 1, is the initiator of the sequence of communication, and is the point of reference for the localization of the rest of the nodes, it is the origin of the coordinate system. 
    \item The direction from Active node 1 to Active node 2 defines the positive x-axis of the system.
    \item All the other nodes are in the positive y-axis.
\end{itemize}

Once the positions are initialized, a subset of nodes is selected based on a desired localization frequency across the whole network, while minimizing the positioning error in individual nodes. The selected subset of nodes functions as mobile anchors in  the system. The active nodes keep localizing themselves using ToF, communicating all-to-all continuously. The rest of nodes of the network start localizing themselves using passive TDoA, intercepting the the messages of the active nodes and using them to compute the difference of distance between each pair of nodes. 

Since all nodes are assumed to be mobile, the subset of UWB nodes that is chosen to be active may not be ideal for all possible configurations, so a role allocation algorithm dynamically changes the role of the UWB nodes as they move, maintaining a the specified number of active nodes for the network. Ideally, the active nodes should be located towards the outside of the system, because the error for the difference of distance for the listener nodes is smaller the closer to the middle of the convex envelope created by the active nodes. Thus, if one of the robots moves more towards the outside, it will change from an active node to a passive listener one.

%% file: sec/04_Methodology.tex

\section{Methodology}

Through this section, we look into the implementation details and experimental setup.

\subsection{UWB ToF Ranging for anchors}

The ranging between two nodes is estimated by implementing either SS-TWR or DS-TWR. The ToF can be computed by \eqref{eq:ss-twr}

\begin{equation}
    ToF=\frac{\left ( T_{init}- T_{resp}\right )}{2}
    \label{eq:ss-twr}
\end{equation}

where $T_{init}$ is the total time since the poll message was sent by the initiator node and the reception time of the response message; and $T_{resp}$ is the time for the replying node to process the poll message and send the reply. This time is sent as information in the response message, where the distance can then be obtained. This process is used between the nodes selected as active on the system. 

\subsection{UWB TDoA Ranging for anchors}

For the TDoA implementation without synchronized clocks a similar computation than before is performed on the listener nodes. Once the initiator nodes has received its response message and computed the respective distance, it sends a new message with the transmission time of the poll message and the reception time of the response message. The TDoA can then be computed by \eqref{eq:ss-twr}

\begin{equation}
    TDoA=T_{list}-T{resp}-ToF
    \label{eq:tdoa}
\end{equation}

where $T_{init}$ is the time in the listener node between the reception of the poll message and the reception of the response message used for the ToF ranging. With it the difference of distance to each pair of nodes can be obtained.

\subsection{UWB Ranging}


\Cref{fig:uwb_implementation} shows an exchange of messages between to active nodes performing ToF ranging, and the reception of the messages by one passive listener. Active node 1 initiates the TWR ranging by sending a poll message. Active node 2 response the polling with the necessary information for the distance computation, just as explained above. Then Active node 1 sends a data message with the the necessary information for the listener node to compute the TDoA and the difference of distance, also as explained before. These three messages are all what is needed for all active nodes to position themselves, as well as for the listeners. 

The active nodes in the network are in constant communication for the ToF ranging, and for the listeners positioning. They follow a fixed sequence known by all nodes in the network. A pair of nodes perform TWR, and the pass to the next pair. To ensure that the next pair of nodes continues the communication, a fourth message is added in the communication sequence, the acknowledgement message. This message is sent by the next node in charge of initiating the  TWR ranging. This ensures that the network follows the set communication sequence. If the acknowledgement message is not sent before a timeout, the data message is sent again until the response is received.

\input{algs/uwb}

\subsection{Positioning and role allocation}

The previous two approaches can be utilized directly for estimating the position of the nodes. However, as we have discussed, ToF has an inherent scalability problem, while TDoA is only stable for passive nodes moving within the convex envelope defined by the active nodes. Our objective is therefore to limit the number of active nodes to meet the localization frequency requirements, and then dynamically select which subset of nodes are active in order to minimize the positioning error of the passive listeners.

As the passive nodes relies on estimations of differences of the distance to pairs of active nodes, the actual position of the node is then found at the intersection of a series of hyperbolas. To minimize the error, we look into finding the set of nodes to be active such that we minimize the distance from the passive nodes to the centroid of all triangles defined by combinations of three of the active nodes.

We use the following notation for the remaining of this section to describe the position estimators. Let $[n] = \{1,\dots,n\}$ be the set of the first $n$ natural numbers. We assume we have a set of $n$ nodes with positions $\{p_i[t]\in\mathbb{R}^3\}_{i\in[n]}$ at a given time $t$. Let then $\mathcal{P}(\mathcal{X})$ represent the set of subsets of $\mathcal{X}$, and $\mathcal{P}_k(\mathcal{X})$ the set of subsets of cardinality $k$. The objective of the dynamic role allocation algorithm is to find at each time step $t$ the set of active nodes $A[t]=\{a_1,\dots,a_k\}_{a_i\in[n]} \in \mathcal{P}_k([n])$, with $k$ is based on the ranging speed and the minimum localization frequency. The set of passive listeners will then be $L[t] = [n] \setminus A[t]$. The set of active nodes $\mathcal{A}$ is then calculated by minimizing the following cost function:

\begin{equation}
    A[t+1] = \argmin_{A \in \mathcal{P}_k\left([n]\right)} \displaystyle\sum_{T\in\mathcal{P}_3(A)} \displaystyle\sum_{l\in[n] \setminus A} \left\lVert \dfrac{\sum_{t\in T}p_t[t]}{3} - p_l[t] \right\rVert^2
\end{equation}

where the positions $p_l[t-1]$ could have been calculated with either ToF or TDoA depending on whether the node was active at the previous time step, $l\in A[t-1]$, or not.

Given the set $A[t]$, we then calculate the positions of all active nodes $\{p_{a_i}\}$ using multilateration and following the assumptions described above, i.e., having the first two anchors defining the direction of the $x$ axis. The position of a passive listener $l\in [n] \setminus A$, $p_l$, is then calculated based on a least squares estimator that minimizes the following cost function:

\begin{equation}
    p_{l}[t] = \argmin_{p \in \mathbb{R}^3} \displaystyle\sum_{\substack{i\in A, \: j\in A \\ i \neq j}} \left( d_{ij}^{l_i} - \left( \left\lVert p - p_{j} \right\rVert - \left\lVert p - p_{i} \right\rVert \right) \right)^2
\end{equation}

Initially, $\{p_i[0]\}_{i\in[n]}$ are calculated using multilateration for all nodes as ToF ranging estimations are obtained between all pairs of nodes.

The process above is summarized in Algorithm~\ref{alg:localization_algorithm}.

\begin{figure}
    \centering
    \includegraphics[width=0.48\textwidth]{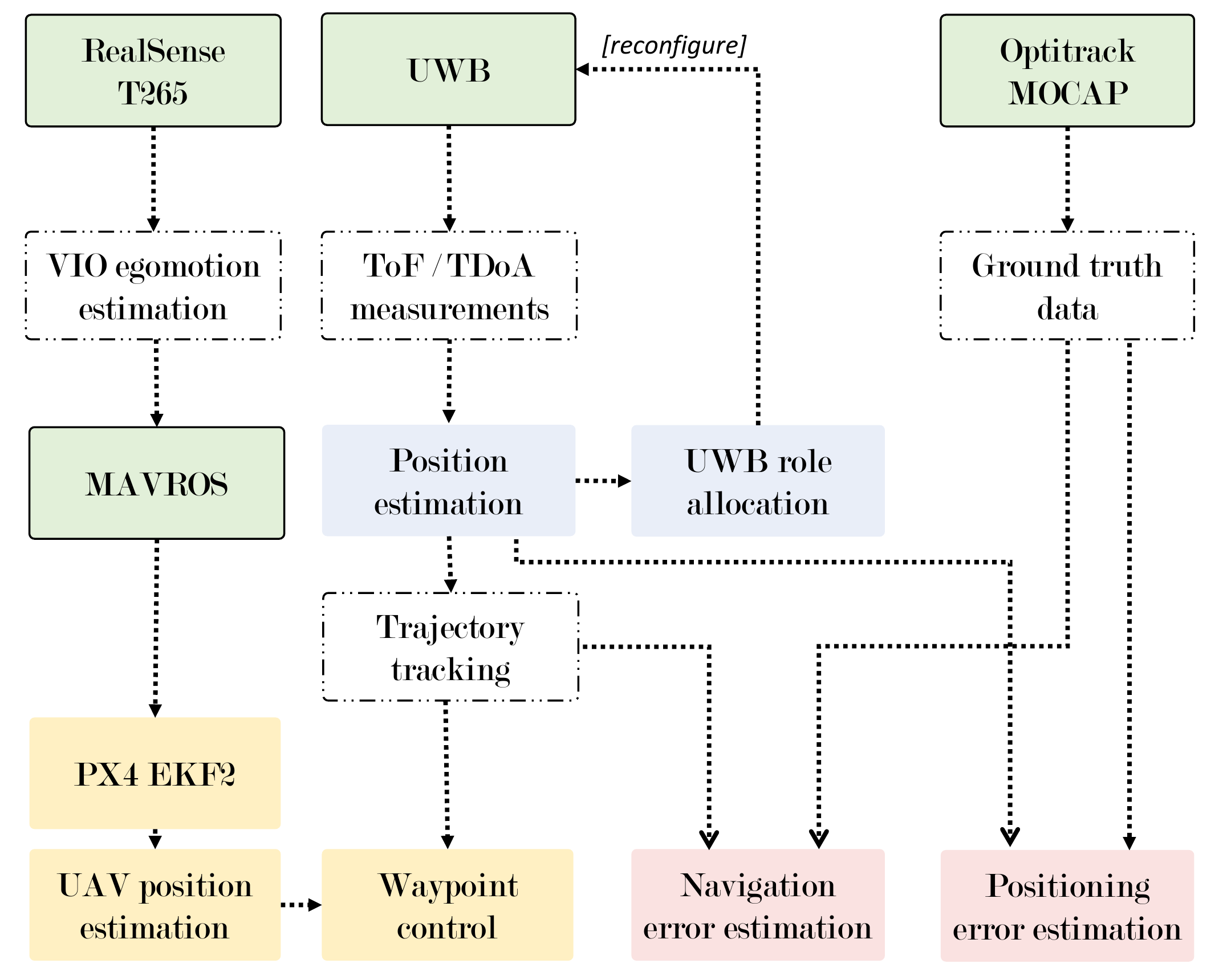}
    \caption{Set of ROS nodes and other modules in the experiment setup.}
    \label{fig:rosNodes}
\end{figure}

\subsection{UWB Nodes}

For this experiment setup, six UWB nodes were deployed. We define a limit of maximum four active nodes. The remaining two must then localize as passive listeners. The UWB devices used in the implementation are the Decawave’s DWM1001 module with custom firmware. We have designed and implemented a custom firmware for the DWM1001 modules to be able to act as active or passive nodes\footnote{https://github.com/TIERS/dynamic-uwb-firmware}. The devices broadcast the ranging information, either ToF or TDoA, from the entire UWB network, so that the position of all nodes can be calculated at any endpoint with all the information available.

\subsection{Robotic platforms}

A custom-built UAV was utilized for the experiments. It is based on the X500 quad-rotor frame, embedded with a Pixhawk 5X flight controller running PX4 firmware. We particularly rely on the EKF2 filter in PX4 for controlling the UAV in offboard mode. The onboard computer 
features
an Intel x5-Z8350 QuadCore processor @ 1.44Ghz running Ubuntu 18.04 and ROS Melodic. The UAV is also equipped with an Intel RealSense T265 camera for VIO-based egomotion estimation. Finally the UAV also has a Decawave DWM1001 UWB node to perform the localization throughout the experimentation. 

\subsection{Experiment Setup}

All nodes are deployed in an empty room of approximately $40\,m^2$ with a MOCAP system that provides the ground truth data. The UAV has one of the UWB nodes mounted and will provide the information of the UWB network to a backbone ROS network where the localization takes place.
The node localization algorithms and the role allocation are implemented through a series of ROS nodes, that distribute the information between the different systems, such as forwarding position information to MAVROS for the low-level control of the UAV.

\Cref{fig:rosNodes} shows the setup used in our experiments from the perspective of sensor drivers, ROS nodes and other algorithms. The UAV relies on fusing the VIO camera data with the flight controller's internal data to maintain stable flight, while the trajectory and motion planning is calculated entirely based on the UWB localization information. Through the experiments, we use three different position estimation methods (ToF, TDoA and dynamic mode) in addition to the MOCAP system for tracking two different trajectories.

%% file: algs/uwb.tex
\begin{algorithm}[t]
    \normalsize
    \small
	\caption{UWB role allocation and relative positioning}
	\label{alg:localization_algorithm}
	\KwIn{\\
	    \hspace{1em}Maximum number of active nodes: $k$ \\
	    \hspace{1em}Node positions at $t-1$: $\{p_i[t-1]\}\in\mathbb{R}^{3n}$; \\
	}
	\KwOut{\\
	    \hspace{1em} Set of active nodes : $A[t+1] \in \mathcal{P}_k([n])$; \\
	    \hspace{1em} Set of passive listeners : $L[t+1] = [n] \setminus A[t+1]$; \\
	    \hspace{1em} Active node positions : $\{p_{a}[t]\}_{a\in A}\in\mathbb{R}^{3k}$; \\
	    \hspace{1em} Passive listener positions : $\{p_{l}[t]\}_{l\in L} \in\mathbb{R}^{3(n-k)}$; \\
	}  
	\BlankLine
    \If {t = 0} {
        $\{d_{ij}\}_{i,j\in[n],i\neq j}$ $\leftarrow$ get\_tof\_ranges$\left([n]\right)$;\\
        $\{p_i[0]\}_{i\in[n]}$ $\leftarrow$ multilateration$\left(\{d_{ij}\}\right)$;\\
    }
    \Else {
        $\{d_{ij}\}_{i,j\in A,i\neq j}$ $\leftarrow$ get\_tof\_ranges$\left(A[t]\right)$;\\
        $\{d^{l}_{ij}\}_{l\in L,i,j\in A,i\neq j}$ $\leftarrow$ get\_tdoa\_ranges$\left(L[t]\right)$;\\[+0.42em]
        \tcp{Active node positions}
        $\{p_a[t]\}_{a\in A}$ $\leftarrow$ multilateration$\left(\{d_{ij}\}\right)$;\\[+0.42em]
        \tcp{Passive listener positions}
        $\{p_l[t]\}_{l\in L}$ $\leftarrow$ tdoa\_ls\_estimator$\left(\{d^{l}_{ij}\}\right)$;\\
    }
    \BlankLine
    \tcp{Role allocation}
    $A[t+1] \leftarrow \argmin_{A \in \mathcal{P}_k\left([n]\right)} tdoa\_cost\_fn\left(\{p_i[t]\}\right)$;\\
    $L[t+1] \leftarrow [n] \setminus A[t+1]$

\end{algorithm}


%% file: sec/05_Experiments.tex

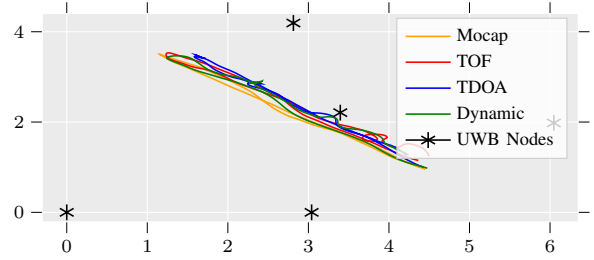
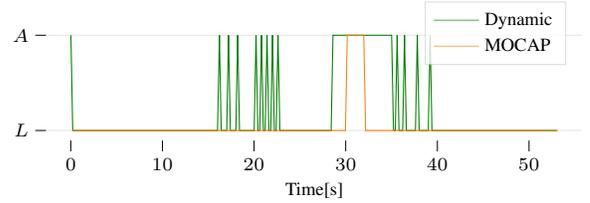
\begin{figure}
    \centering 
    \begin{subfigure}{0.48\textwidth}
        \centering
        \setlength\figureheight{0.5\textwidth}
        \setlength\figurewidth{\textwidth}
        \scriptsize{\input{tex/line_nodes_pose}}
         \caption{Two-dimensional path for autonomous UAV flights where a linear trajectory was tracked based on the different positioning methods.}
        \label{fig:line_XY_pose}
    \end{subfigure}
    
    \vspace{1em}
    \begin{subfigure}{0.48\textwidth}
        \centering
        \setlength\figureheight{0.4\textwidth}
        \setlength\figurewidth{\textwidth}
        \scriptsize{\input{tex/line_roles3}}
        \caption{Dynamic role allocation for the UAV UWB node during the line trajectory.}
        \label{fig:line_role}
    \end{subfigure}
    
    \caption{Path reconstruction with different positioning methods and UAV UWB node role for a linear trajectory.}
    \label{fig:line_results}

\end{figure}

\begin{figure}
    \centering 
    \begin{subfigure}{0.48\textwidth}
        \centering
        \setlength\figureheight{0.5\textwidth}
        \setlength\figurewidth{\textwidth}
        \scriptsize{\input{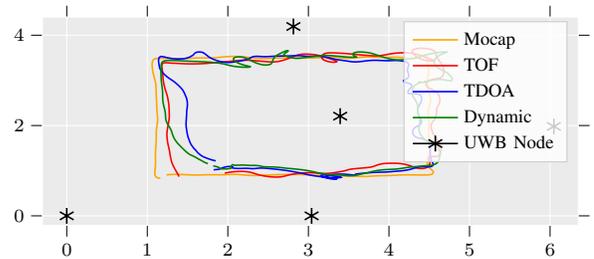}
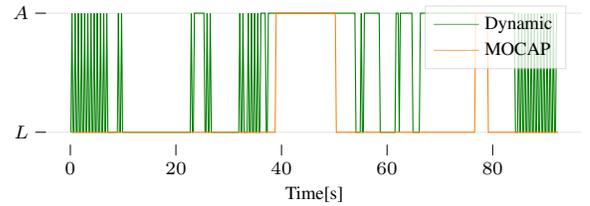}
        \caption{Two-dimensional path for autonomous UAV flights where a rectangular trajectory was tracked based on the different positioning methods.}
        \label{fig:box_XY_pose}
    \end{subfigure}
    \begin{subfigure}{0.48\textwidth}
        \centering
        \setlength\figureheight{0.4\textwidth}
        \setlength\figurewidth{\textwidth}
        \scriptsize{\input{tex/box_roles3}}
        \caption{Dynamic role allocation for the UAV UWB node during the rectangular trajectory}
        \label{fig:box_role}
    \end{subfigure}
    \caption{Path reconstruction with different positioning methods and UAV UWB node role for a rectangular trajectory.}
    \label{fig:rectangle_results}
    \vspace{-1em}
\end{figure}

\section{Experimental Results}

For the experiment setup described above two different flying patterns were tested. The first consisting on a straight line across the room shown in~\cref{fig:line_XY_pose} and the second being a rectangular trajectory shown in~\cref{fig:box_XY_pose}. For each of the patterns four different runs were performed with different positioning methods to compare the proposed solution with baseline ToF and TDoA approaches. 

The four runs consist of using the data provided by the MOCAP system to obtain a reference trajectory with the UAV, ToF ranging, TDoA ranging, and the dynamic approach proposed. Each run provides the data for the robot to move in order to follow the predefined trajectory. The UWB nodes set on the experiment room were place in order to have the node in the UAV be both inside and outside of the convex envelope of the active nodes, so the UAV UWB node would ideally switch in multiple occasions between acting as an active node and as a passive listener. 

\Cref{fig:line_XY_pose} and~\cref{fig:box_XY_pose} show the four runs of the UAV for the two predefined trajectories of the experiment. The roles for the dynamic approach during the execution can be seen in~\cref{fig:line_role} and~\cref{fig:box_role}, respectively. They show the comparison of the roles in the dynamic approach with regards to the ideal scenario provided by the MOCAP system. In both trajectories the UAV UWB node has a noisier behavior for the dynamic method. However, the periods of longer more stable change in the dynamic approach correspond to change in ideal one. It is worth noting that while the position estimation might differ significantly between methods, the fact that the flight stability is based on the VIO estimations and the data is forwarded to the EKF2 filter mean that the actual trajectory tracking error will be smaller than the real-time, unfiltered UWB positioning error, and that fast-switching between modes does not degrade performance noticeably.

\begin{figure}
    \begin{subfigure}{0.48\textwidth}
        \centering
        \setlength\figureheight{0.5\textwidth}
        \setlength\figurewidth{\textwidth}
        \scriptsize{\input{tex/line_nodes_pose2}}
        \caption{Raw position estimations for the same trajectories in subfigure (a), before filtering with VIO egomotion estimations at the UAV.}
        \label{fig:line_XY_pose2}
    \end{subfigure}
    
    \vspace{1em}
    \begin{subfigure}{0.48\textwidth}
        \centering
        \setlength\figureheight{0.5\textwidth}
        \setlength\figurewidth{0.95\textwidth}
        \scriptsize{\input{tex/line_boxplot}}
        \caption{Positioning error for raw UWB estimations before filtering for the line trajectory.}
        \label{fig:line_boxplot}
    \end{subfigure}
    
    \vspace{1em}
    \begin{subfigure}{0.48\textwidth}
        \centering
        \setlength\figureheight{0.5\textwidth}
        \setlength\figurewidth{0.95\textwidth}
        \scriptsize{\input{tex/line_Error_distance_NODE_4}}
        \caption{Positioning errors over time for the linear trajectory with the different UWB approaches.}
        \label{fig:line_error}
    \end{subfigure}
    
    \caption{Raw positioning estimations, raw positioning error for a linear trajectory.}
    
    \label{fig:line_results2}

\vspace{-1em}
\end{figure}
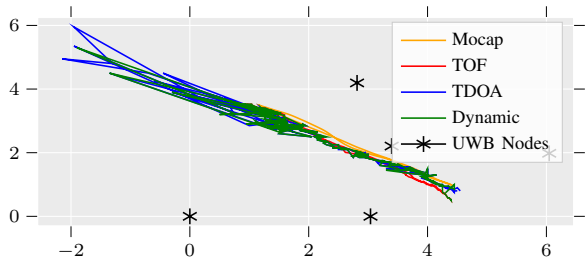
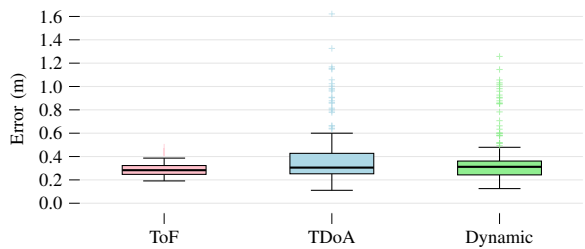
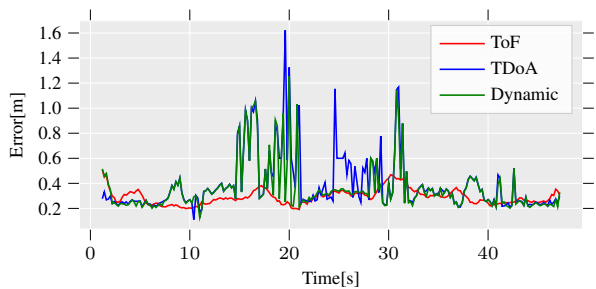

Continuing with the raw UWB error at independent time steps, we show in \Cref{fig:line_error} the distance error of the three localization approaches with reference to the ground truth for the linear trajectory. As expected the error of the dynamic implementation proposed matches the error of the ToF approach when the node is performing as an active node and the error of the TDoA approach when the nodes acts as a passive listener. The unfiltered measurements of the three approaches can be appreciated in~\cref{fig:line_XY_pose2}, representing the path that would be reconstructed from the raw UWB positioning data rather than the actual flight path that the UAV has followed and which is shown in~\cref{fig:line_XY_pose}.

\begin{figure*}
    \centering
    \begin{subfigure}{0.33\textwidth}
        \centering
        \setlength\figureheight{.8\textwidth}
        \setlength\figurewidth{.95\textwidth}
        \scriptsize{\input{tex/box_boxplot}}
        \caption{Positioning error for raw UWB estimations before filtering.}
        \label{fig:box_boxplot}
    \end{subfigure}
    \begin{subfigure}{0.32\textwidth}
        \centering
        \setlength\figureheight{0.825\textwidth}
        \setlength\figurewidth{0.95\textwidth}
        \scriptsize{\input{tex/box_boxplot_2}}
        \caption{Overall distribution of the trajectory tracking error.}
        \label{fig:box_boxplot2}
    \end{subfigure}
    \begin{subfigure}{0.33\textwidth}
        \centering
        \setlength\figureheight{0.8\textwidth}
        \setlength\figurewidth{0.95\textwidth}
        \scriptsize{\input{tex/box_boxplot_3}}
        \caption{Trajectory tracking error when switching roles.}
        \label{fig:box_boxplot3}
    \end{subfigure}
    \caption{Raw UWB positioning error, trajectory tracking error and navigation error in the rectangular trajectory.}
    \label{fig:square_tr_boxplots}
    \vspace{-1em}
\end{figure*}
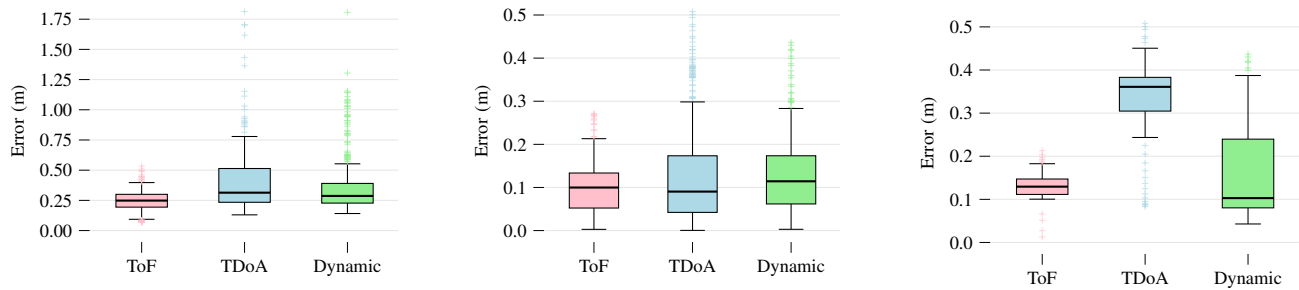

Finally, in~\cref{fig:square_tr_boxplots} we show the positioning and navigation error for the flights following the rectangular trajectory in~\cref{fig:rectangle_results}. \Cref{fig:box_boxplot} shows the raw positioning error for the different UWB positioning methods before the data is filtered together with VIO data in PX4. This is therefore the instantaneous error in individual, uncorrelated, UWB-based estimation. In~\cref{fig:box_boxplot2}, we plot the trajectory tracking error, which better showcases the ability of the robot to follow the predefined trajectory. The error is significantly reduced as individual estimations do not have a significant effect on the robot's flight. The results show that the dynamic approach is slightly more stable than a standard TDoA method. However, when we look closer in ~\cref{fig:box_boxplot3} at the trajectory tracking error from the part of the trajectory where the node switches from TDoA role to ToF and back, the proposed approach outperforms the baseline.

%% file: tex/line_nodes_pose.tex
\begin{tikzpicture}

\definecolor{color0}{rgb}{1,0.647058823529412,0}
\definecolor{color1}{rgb}{1,0.647058823529412,0}
\definecolor{color2}{rgb}{1,0.647058823529412,0}

\begin{axis}[
height=\figureheight,
width=\figurewidth,
axis background/.style={fill=white!92!black},
axis line style={white},
legend cell align={left},
legend style={fill opacity=0.8, draw opacity=1, text opacity=1, draw=white!80!black},
tick align=outside,
tick pos=both,
x grid style={white},
xmajorgrids,
xmin=-0.302363967895508, xmax=6.34964332580566,
xtick style={color=black},
y grid style={white},
ymajorgrids,
ymin=-0.20967013835907, ymax=4.40307290554047,
ytick style={color=black}
]

\addplot [semithick, color0]
table {%
4.40268182754517 0.986898362636566
4.37718868255615 1.00204133987427
4.3503098487854 1.01843881607056
4.32239866256714 1.03605592250824
4.29403734207153 1.05497562885284
4.2653865814209 1.07530617713928
4.23669481277466 1.09682559967041
4.20790672302246 1.1191178560257
4.17946577072144 1.14236390590668
4.15121269226074 1.16688311100006
4.12203216552734 1.19253385066986
4.09261608123779 1.21896243095398
4.06327247619629 1.2452632188797
4.03413009643555 1.27206444740295
4.00511121749878 1.29894769191742
3.97613859176636 1.3258193731308
3.94705748558044 1.35238683223724
3.91769218444824 1.37819087505341
3.88855886459351 1.40380108356476
3.8593955039978 1.42950987815857
3.83040547370911 1.45510792732239
3.80150127410889 1.48029637336731
3.77255201339722 1.50384747982025
3.74366092681885 1.52567183971405
3.71446132659912 1.54716658592224
3.68520951271057 1.56855309009552
3.65614581108093 1.58985579013824
3.62630105018616 1.61039710044861
3.59604454040527 1.63080132007599
3.56558227539062 1.65200173854828
3.5349018573761 1.67381811141968
3.5037693977356 1.69652569293976
3.47229409217834 1.71944570541382
3.44041275978088 1.7420836687088
3.4081757068634 1.76521468162537
3.37594819068909 1.78841757774353
3.34345889091492 1.81170237064362
3.31018376350403 1.83449327945709
3.27601194381714 1.8565559387207
3.24104285240173 1.87832140922546
3.20589089393616 1.90018904209137
3.17120742797852 1.92251563072205
3.13686156272888 1.94472873210907
3.10301113128662 1.96685969829559
3.06943941116333 1.98943650722504
3.03647303581238 2.01271915435791
3.00443243980408 2.03622055053711
2.9735004901886 2.05958437919617
2.94316744804382 2.08183693885803
2.91318106651306 2.1037278175354
2.88576126098633 2.1230936050415
2.85590958595276 2.14399766921997
2.82392382621765 2.16724252700806
2.79513001441956 2.18884491920471
2.76633238792419 2.21186518669128
2.73765110969543 2.2350332736969
2.70868396759033 2.2590696811676
2.67965841293335 2.28270244598389
2.65106654167175 2.30640578269958
2.6229875087738 2.33003735542297
2.59557461738586 2.35282969474792
2.56818842887878 2.37531161308289
2.54075884819031 2.39727687835693
2.51244044303894 2.41896748542786
2.48280668258667 2.44075322151184
2.45245361328125 2.46346950531006
2.42141318321228 2.48717951774597
2.39054727554321 2.51114702224731
2.36081099510193 2.53438067436218
2.33155941963196 2.55703949928284
2.30255913734436 2.57940268516541
2.27358889579773 2.60214281082153
2.24479794502258 2.62497854232788
2.21659636497498 2.64717578887939
2.18845462799072 2.66864514350891
2.1591796875 2.68936395645142
2.13058304786682 2.7105438709259
2.10114741325378 2.73174571990967
2.07148885726929 2.75335288047791
2.04269218444824 2.77582621574402
2.01376271247864 2.79885387420654
1.98547279834747 2.821448802948
1.95838737487793 2.84367346763611
1.93244159221649 2.86436414718628
1.9073885679245 2.88418459892273
1.88319253921509 2.90414619445801
1.85854256153107 2.92402243614197
1.83324205875397 2.94392919540405
1.80647552013397 2.96406674385071
1.78133106231689 2.98301291465759
1.75090730190277 3.00581955909729
1.72255349159241 3.02755093574524
1.69390654563904 3.04995656013489
1.6653459072113 3.07303762435913
1.63589191436768 3.09681582450867
1.60545098781586 3.12119364738464
1.57466840744019 3.14559006690979
1.54346418380737 3.17015790939331
1.51202750205994 3.19402432441711
1.48013579845428 3.21746492385864
1.44942235946655 3.23980188369751
1.4189121723175 3.26101136207581
1.38951981067657 3.28150296211243
1.36068439483643 3.30298924446106
1.3325047492981 3.32629561424255
1.30488717556 3.35142874717712
1.27794885635376 3.37757563591003
1.25111937522888 3.40381169319153
1.22424626350403 3.42945289611816
1.19723296165466 3.45466446876526
1.17107892036438 3.47866225242615
1.15274381637573 3.49681973457336
1.14919579029083 3.50429511070251
1.15759515762329 3.50115823745728
1.17440867424011 3.4904580116272
1.19552731513977 3.475905418396
1.21941745281219 3.46104168891907
1.24508416652679 3.44626331329346
1.27125871181488 3.43117833137512
1.29767549037933 3.41576910018921
1.32505488395691 3.39938116073608
1.35312640666962 3.38245368003845
1.38170480728149 3.36586236953735
1.41107106208801 3.34948134422302
1.44061708450317 3.33266258239746
1.47060334682465 3.31598091125488
1.50085318088531 3.29919695854187
1.53125536441803 3.28255653381348
1.56209278106689 3.26524043083191
1.59345805644989 3.24795484542847
1.62461733818054 3.22942590713501
1.65626847743988 3.20973920822144
1.68750047683716 3.18858242034912
1.71754670143127 3.16637754440308
1.74726068973541 3.1439573764801
1.77669978141785 3.12061262130737
1.80592584609985 3.09657645225525
1.83534777164459 3.07227158546448
1.86514210700989 3.04839563369751
1.89458608627319 3.02463960647583
1.92394304275513 3.00089693069458
1.95289051532745 2.97637724876404
1.98216187953949 2.95113778114319
2.01147246360779 2.92566871643066
2.04050898551941 2.90015292167664
2.06924510002136 2.87480783462524
2.09762668609619 2.85006260871887
2.1265287399292 2.82511615753174
2.15518832206726 2.7987744808197
2.18505930900574 2.77176332473755
2.21460437774658 2.74428677558899
2.24303555488586 2.71620273590088
2.27136325836182 2.6865348815918
2.29803490638733 2.65681290626526
2.32343697547913 2.62640619277954
2.34841442108154 2.59512209892273
2.37191414833069 2.5637526512146
2.394207239151 2.53305125236511
2.41683101654053 2.50223112106323
2.43872952461243 2.47208046913147
2.45998573303223 2.44291877746582
2.48106932640076 2.41523766517639
2.50179529190063 2.3877055644989
2.52349424362183 2.36004900932312
2.54635000228882 2.33266425132751
2.56925392150879 2.30615925788879
2.5917694568634 2.28063225746155
2.61509013175964 2.25548481941223
2.64160108566284 2.23071503639221
2.66832876205444 2.20557117462158
2.69691038131714 2.18015050888062
2.72584891319275 2.15488219261169
2.75589656829834 2.13113164901733
2.78750038146973 2.11033868789673
2.81976366043091 2.09133410453796
2.85169672966003 2.0729513168335
2.88317561149597 2.05336427688599
2.91412258148193 2.03364872932434
2.94459581375122 2.0131528377533
2.97585463523865 1.99245870113373
3.00693202018738 1.97182047367096
3.03836297988892 1.9518837928772
3.06949782371521 1.93295931816101
3.09965109825134 1.9162757396698
3.12900519371033 1.90081357955933
3.1582715511322 1.88599121570587
3.18788051605225 1.87162268161774
3.21783351898193 1.8571150302887
3.24794840812683 1.84302639961243
3.27872180938721 1.82801532745361
3.30975866317749 1.81187498569489
3.34059691429138 1.79416155815125
3.37142038345337 1.77500081062317
3.40125894546509 1.75466310977936
3.43030738830566 1.73359620571136
3.45815205574036 1.71161484718323
3.48527431488037 1.68978142738342
3.51151752471924 1.66813778877258
3.53709864616394 1.64698100090027
3.56275320053101 1.6258111000061
3.58831214904785 1.60441243648529
3.61450409889221 1.58268570899963
3.64064073562622 1.56068825721741
3.66722393035889 1.53901743888855
3.69424176216125 1.51709794998169
3.72168779373169 1.49510025978088
3.74917268753052 1.47242701053619
3.77647113800049 1.44908201694489
3.80365014076233 1.42537248134613
3.83110308647156 1.40186882019043
3.8584508895874 1.37822985649109
3.88592433929443 1.35429978370667
3.91345238685608 1.32986664772034
3.94114542007446 1.30503022670746
3.96883869171143 1.28017628192902
3.99609279632568 1.2559107542038
4.02343893051147 1.23314929008484
4.05081605911255 1.21148586273193
4.07815647125244 1.19119954109192
4.10546016693115 1.17226815223694
4.13397026062012 1.15503168106079
4.16349935531616 1.1388373374939
4.19379758834839 1.1232351064682
4.22444009780884 1.10774600505829
4.25523900985718 1.091268658638
4.28492069244385 1.07376670837402
4.31411027908325 1.0554621219635
4.34266901016235 1.03604662418365
4.37073564529419 1.01496279239655
4.39835119247437 0.993474662303925
4.42510890960693 0.97095662355423
4.44992733001709 0.948507487773895
};
\addlegendentry{Mocap}
\addplot [semithick, red]
table {%
4.49333000183105 1.24264025688171
4.48998641967773 1.27486598491669
4.48122978210449 1.30613744258881
4.4683575630188 1.33908426761627
4.45215034484863 1.37198114395142
4.43209838867188 1.40387344360352
4.40779781341553 1.43243193626404
4.37965631484985 1.45719230175018
4.35146331787109 1.47694301605225
4.31647825241089 1.49753284454346
4.28291463851929 1.51181256771088
4.24965858459473 1.51834094524384
4.2210545539856 1.51748132705688
4.193199634552 1.50804460048676
4.16853666305542 1.49087488651276
4.14924097061157 1.47061383724213
4.13079023361206 1.45112895965576
4.10717344284058 1.44005966186523
4.07873392105103 1.43950176239014
4.04766511917114 1.44789958000183
4.01686429977417 1.46267998218536
3.98891091346741 1.48310852050781
3.96606874465942 1.50517904758453
3.94019627571106 1.53249168395996
3.92149519920349 1.5573319196701
3.91469073295593 1.57925868034363
3.92003059387207 1.59821557998657
3.93389463424683 1.61480677127838
3.95305943489075 1.62979471683502
3.97066831588745 1.64548659324646
3.97751331329346 1.66124296188354
3.97202849388123 1.68020594120026
3.95863819122314 1.69539487361908
3.93551683425903 1.70989298820496
3.90921115875244 1.71892046928406
3.88040542602539 1.72439467906952
3.85285115242004 1.72634482383728
3.82203435897827 1.72751545906067
3.7908616065979 1.72684335708618
3.75738954544067 1.72339046001434
3.7277615070343 1.71607351303101
3.70168328285217 1.7031067609787
3.68399238586426 1.6869090795517
3.67269992828369 1.66281151771545
3.67240738868713 1.6395218372345
3.68076753616333 1.61797165870667
3.69479179382324 1.60117661952972
3.71679520606995 1.58675146102905
3.73848247528076 1.57877469062805
3.76618599891663 1.57417464256287
3.79256105422974 1.57519388198853
3.81688761711121 1.58038604259491
3.84175801277161 1.59235465526581
3.85416841506958 1.60886836051941
3.85553646087646 1.62898480892181
3.84931135177612 1.6488391160965
3.83489537239075 1.67249047756195
3.81697869300842 1.69398200511932
3.79655122756958 1.7150342464447
3.77470088005066 1.73577880859375
3.75368285179138 1.75499975681305
3.73002767562866 1.77619791030884
3.70324730873108 1.79804110527039
3.67704010009766 1.81633651256561
3.64859104156494 1.83216977119446
3.61873006820679 1.84566032886505
3.58792853355408 1.85985326766968
3.55674457550049 1.87561333179474
3.5246160030365 1.89158427715302
3.49132561683655 1.90523207187653
3.45964574813843 1.91688752174377
3.42199993133545 1.93064391613007
3.3873872756958 1.94441413879395
3.35336589813232 1.95923185348511
3.31993460655212 1.97518360614777
3.28675270080566 1.99244439601898
3.25662612915039 2.00819969177246
3.22200918197632 2.02692198753357
3.19133520126343 2.04447770118713
3.16169548034668 2.06237292289734
3.13301515579224 2.08105945587158
3.10503172874451 2.10132837295532
3.07862567901611 2.12413787841797
3.05457329750061 2.14964461326599
3.03175568580627 2.17671084403992
3.01033759117126 2.2021484375
2.98404049873352 2.23246097564697
2.95915961265564 2.26089906692505
2.93317413330078 2.28932285308838
2.9059956073761 2.31730437278748
2.88047575950623 2.34172582626343
2.8505482673645 2.37030172348022
2.8230767250061 2.39602589607239
2.79479575157166 2.42046117782593
2.76582837104797 2.44341468811035
2.73658680915833 2.4656617641449
2.70722150802612 2.48753380775452
2.67811584472656 2.50861215591431
2.64795327186584 2.52813100814819
2.61683201789856 2.54650783538818
2.58859944343567 2.56347012519836
2.55563521385193 2.58439159393311
2.52541661262512 2.60333156585693
2.49840092658997 2.6203601360321
2.4675087928772 2.64026832580566
2.43936467170715 2.65980386734009
2.41336274147034 2.67843985557556
2.38148283958435 2.69959783554077
2.35370993614197 2.71610045433044
2.32320713996887 2.73426580429077
2.29013252258301 2.75539374351501
2.2578604221344 2.77616024017334
2.22418880462646 2.7975594997406
2.19083452224731 2.81897664070129
2.1614682674408 2.83854675292969
2.12922644615173 2.86125826835632
2.10107493400574 2.88214516639709
2.07497072219849 2.90345668792725
2.05024647712708 2.92508721351624
2.02776575088501 2.94521450996399
2.00060653686523 2.96970272064209
1.97396123409271 2.99268817901611
1.94608819484711 3.01569223403931
1.91821658611298 3.03932690620422
1.89038372039795 3.06236624717712
1.86436522006989 3.08233547210693
1.83249068260193 3.10448455810547
1.80466842651367 3.12209248542786
1.77056801319122 3.14173698425293
1.74137425422668 3.15695953369141
1.70839619636536 3.17177653312683
1.67052149772644 3.18489456176758
1.6357753276825 3.19507050514221
1.60175120830536 3.20662927627563
1.57181298732758 3.21965527534485
1.54113030433655 3.23448252677917
1.51171636581421 3.25067615509033
1.48111522197723 3.27061247825623
1.45558488368988 3.28763031959534
1.42797517776489 3.30469870567322
1.39832782745361 3.3216438293457
1.37170004844666 3.3360230922699
1.34512197971344 3.35033988952637
1.31869292259216 3.365483045578
1.29554152488708 3.38125061988831
1.27209424972534 3.40214467048645
1.25113570690155 3.4289755821228
1.23806631565094 3.45703148841858
1.23280739784241 3.48703980445862
1.23526644706726 3.5112943649292
1.24519681930542 3.52815937995911
1.26255810260773 3.53337025642395
1.28242015838623 3.52913165092468
1.30395448207855 3.52008175849915
1.32477450370789 3.50958013534546
1.35062897205353 3.49557518959045
1.37292587757111 3.48235821723938
1.40021300315857 3.46604061126709
1.42398083209991 3.451819896698
1.45078575611115 3.43627023696899
1.47830879688263 3.42104291915894
1.50912249088287 3.40521812438965
1.53586030006409 3.39182329177856
1.568439245224 3.37607550621033
1.59903848171234 3.361083984375
1.62762928009033 3.34618711471558
1.66240632534027 3.32720255851746
1.69459736347198 3.30818748474121
1.72718167304993 3.28800296783447
1.75992691516876 3.26633858680725
1.7897253036499 3.24521613121033
1.82503759860992 3.21980834007263
1.85704469680786 3.1962296962738
1.88646161556244 3.17489457130432
1.92131018638611 3.14977192878723
1.95097124576569 3.12897276878357
1.98600661754608 3.10526132583618
2.01696634292603 3.08354020118713
2.04484009742737 3.06423282623291
2.07643008232117 3.04145312309265
2.10268044471741 3.02160358428955
2.13407397270203 2.99727654457092
2.16313982009888 2.97429919242859
2.19306778907776 2.9505181312561
2.22070002555847 2.92891001701355
2.25403451919556 2.90376567840576
2.28210496902466 2.88264608383179
2.31276106834412 2.85950040817261
2.34543108940125 2.83415722846985
2.37285804748535 2.81224274635315
2.40519404411316 2.78583669662476
2.43327045440674 2.76282262802124
2.46435642242432 2.73648810386658
2.49489426612854 2.70923018455505
2.525062084198 2.68073606491089
2.55425000190735 2.65138792991638
2.58493900299072 2.61904072761536
2.61182737350464 2.58927226066589
2.63691234588623 2.55979180335999
2.66093802452087 2.52993631362915
2.68458676338196 2.49959540367126
2.70413017272949 2.47097778320312
2.7254900932312 2.43930315971375
2.74570441246033 2.40818285942078
2.76587796211243 2.37775754928589
2.78676438331604 2.34703731536865
2.80988216400146 2.31450271606445
2.82990264892578 2.28775119781494
2.85223007202148 2.25924301147461
2.87641143798828 2.22932624816895
2.8994677066803 2.20218563079834
2.92185616493225 2.17786073684692
2.94788885116577 2.15247225761414
2.97776460647583 2.12698531150818
3.00445008277893 2.10676789283752
3.03643989562988 2.08370280265808
3.06270861625671 2.06513023376465
3.09101629257202 2.0449423789978
3.11924934387207 2.02451157569885
3.14712357521057 2.00383305549622
3.17707085609436 1.98085677623749
3.20443773269653 1.95925176143646
3.23187565803528 1.93787932395935
3.25936794281006 1.91641223430634
3.28757381439209 1.89513862133026
3.31430816650391 1.87612986564636
3.34426927566528 1.85602426528931
3.37701749801636 1.83455085754395
3.40549612045288 1.81633055210114
3.43988013267517 1.79453814029694
3.47209334373474 1.77377200126648
3.50465869903564 1.75218260288239
3.53798961639404 1.7293221950531
3.56913566589355 1.70722925662994
3.60620379447937 1.68065989017487
3.63970685005188 1.65637612342834
3.6698842048645 1.63469409942627
3.70554804801941 1.60962891578674
3.73530220985413 1.58916223049164
3.76996421813965 1.56452715396881
3.80109596252441 1.54189968109131
3.83079409599304 1.51856911182404
3.85878324508667 1.49551630020142
3.88355016708374 1.47511541843414
3.9103627204895 1.45332515239716
3.93753719329834 1.43190205097198
3.96495079994202 1.41092658042908
3.99237966537476 1.39058339595795
4.01935052871704 1.37090384960175
4.04792785644531 1.35016012191772
4.07139873504639 1.33321273326874
4.09574127197266 1.31541454792023
4.12098979949951 1.29659640789032
4.14257049560547 1.28036677837372
4.1666316986084 1.26251399517059
4.19393920898438 1.24290668964386
4.22022390365601 1.22517168521881
4.24756002426147 1.20827233791351
4.27368688583374 1.19341099262238
4.30343341827393 1.17810773849487
4.3367018699646 1.16260004043579
4.36497449874878 1.15025699138641
};
\addlegendentry{TOF}
\addplot [semithick, blue]
table {%
4.21784353256226 1.26790022850037
4.19831085205078 1.29664242267609
4.17531871795654 1.3272305727005
4.14950037002563 1.35982477664948
4.12515211105347 1.38976097106934
4.09936618804932 1.41808676719666
4.07291221618652 1.44480419158936
4.0473837852478 1.46671295166016
4.01743984222412 1.48692202568054
3.98332381248474 1.50551497936249
3.95129370689392 1.52155816555023
3.91880345344543 1.53847992420197
3.88760590553284 1.55682969093323
3.85971355438232 1.57594323158264
3.82820081710815 1.60069680213928
3.80105519294739 1.6216481924057
3.76866173744202 1.64554750919342
3.73789763450623 1.66582262516022
3.70633506774902 1.68522965908051
3.67437791824341 1.70407712459564
3.64419341087341 1.72101867198944
3.60821628570557 1.7396274805069
3.57460021972656 1.75640571117401
3.54407453536987 1.77260220050812
3.50882291793823 1.79368615150452
3.47931361198425 1.81268882751465
3.44503855705261 1.83578753471375
3.41476511955261 1.86004889011383
3.38684034347534 1.88712477684021
3.36384034156799 1.91713106632233
3.34843444824219 1.94957113265991
3.34222102165222 1.98226547241211
3.34132313728333 2.01049375534058
3.34103512763977 2.04073548316956
3.338547706604 2.07253479957581
3.32955884933472 2.09690690040588
3.31020832061768 2.12136149406433
3.28684520721436 2.14111471176147
3.26008319854736 2.15792989730835
3.23054981231689 2.17133140563965
3.20204329490662 2.18169498443604
3.16761517524719 2.19125771522522
3.13521194458008 2.19713449478149
3.10304427146912 2.20093393325806
3.0707995891571 2.20254993438721
3.04147386550903 2.20472264289856
3.0100359916687 2.21175312995911
2.97751259803772 2.22617101669312
2.94895434379578 2.24552798271179
2.92156052589417 2.26900577545166
2.89448714256287 2.29446601867676
2.86667919158936 2.32024478912354
2.84101343154907 2.34360671043396
2.8100528717041 2.37044787406921
2.78219079971313 2.39597821235657
2.75671672821045 2.42423391342163
2.73465824127197 2.45574879646301
2.7126157283783 2.48870611190796
2.69151449203491 2.52275037765503
2.67183589935303 2.55436635017395
2.64840912818909 2.58628797531128
2.61957192420959 2.61641001701355
2.59072494506836 2.64101600646973
2.55965709686279 2.66347026824951
2.52680087089539 2.68516373634338
2.4928777217865 2.70564365386963
2.45817399024963 2.72245097160339
2.42659711837769 2.73591923713684
2.39258980751038 2.75062441825867
2.35846829414368 2.76509976387024
2.32489371299744 2.78009867668152
2.29208946228027 2.79647827148438
2.26239514350891 2.81871843338013
2.24758195877075 2.84194231033325
2.24694919586182 2.85936999320984
2.2577919960022 2.87115168571472
2.27616429328918 2.87594151496887
2.30217266082764 2.8762264251709
2.32812976837158 2.87455892562866
2.35962581634521 2.86982846260071
2.38873314857483 2.86218810081482
2.40911746025085 2.85350775718689
2.41511654853821 2.84920024871826
2.40768194198608 2.85006284713745
2.39351344108582 2.85419297218323
2.37091755867004 2.86047053337097
2.34695029258728 2.86672210693359
2.32097434997559 2.87410736083984
2.2939932346344 2.8836498260498
2.26651120185852 2.89463806152344
2.23894691467285 2.90722107887268
2.21228909492493 2.9215350151062
2.18537187576294 2.93700122833252
2.15799450874329 2.95186233520508
2.129474401474 2.96415686607361
2.10039758682251 2.97526526451111
2.07087588310242 2.98739171028137
2.04154348373413 3.00239586830139
2.01493740081787 3.01820421218872
1.98530066013336 3.03724336624146
1.95352506637573 3.06035470962524
1.92453718185425 3.08329081535339
1.89626121520996 3.10726571083069
1.86814296245575 3.1317446231842
1.84407722949982 3.15512228012085
1.81769621372223 3.18445801734924
1.79590249061584 3.21371126174927
1.77668917179108 3.24165821075439
1.75371909141541 3.27469682693481
1.73118877410889 3.30341219902039
1.70979046821594 3.32739353179932
1.68219923973083 3.35247540473938
1.6581791639328 3.37170720100403
1.63246583938599 3.39163589477539
1.6119350194931 3.40859985351562
1.59891021251678 3.41964912414551
1.58824980258942 3.43255186080933
1.58377397060394 3.4453718662262
1.58942985534668 3.45086264610291
1.60319292545319 3.44879460334778
1.62197482585907 3.44218969345093
1.64245367050171 3.43427324295044
1.66884231567383 3.42516922950745
1.68741929531097 3.42155289649963
1.70206809043884 3.42143487930298
1.7098354101181 3.42140889167786
1.71256804466248 3.42048406600952
1.71337592601776 3.41893267631531
1.71233785152435 3.41705179214478
1.70998275279999 3.41531682014465
1.70003116130829 3.41823244094849
1.68329739570618 3.42702269554138
1.66213715076447 3.44066143035889
1.63826012611389 3.45712184906006
1.61409080028534 3.47366046905518
1.59811210632324 3.48477363586426
1.59504270553589 3.48840236663818
1.60663223266602 3.48279333114624
1.62400388717651 3.46973633766174
1.64508891105652 3.45166397094727
1.66877770423889 3.43200373649597
1.69518756866455 3.41450476646423
1.72279381752014 3.39810991287231
1.74887001514435 3.37829995155334
1.774258852005 3.35533261299133
1.79912352561951 3.33157777786255
1.82368016242981 3.30715179443359
1.84637475013733 3.28429961204529
1.87255823612213 3.259845495224
1.90240263938904 3.23458623886108
1.92858326435089 3.21405982971191
1.95787787437439 3.19256854057312
1.98991894721985 3.16880369186401
2.01795125007629 3.14863729476929
2.05089116096497 3.12395119667053
2.07847499847412 3.10190629959106
2.10796165466309 3.0763304233551
2.13626408576965 3.04955649375916
2.16408061981201 3.02213764190674
2.19210958480835 2.99476766586304
2.22271513938904 2.96507430076599
2.24862599372864 2.94048738479614
2.27946877479553 2.91246867179871
2.30548620223999 2.8896746635437
2.3341224193573 2.86557984352112
2.36301827430725 2.84083580970764
2.39387655258179 2.81331896781921
2.42238521575928 2.78748440742493
2.45013213157654 2.76108741760254
2.47545671463013 2.73690438270569
2.50225138664246 2.71087098121643
2.53077626228333 2.68291401863098
2.55434489250183 2.6596212387085
2.58213949203491 2.63316249847412
2.60538077354431 2.61204314231873
2.63131642341614 2.58982253074646
2.65985703468323 2.56522297859192
2.68658065795898 2.54217648506165
2.71267318725586 2.51837587356567
2.73915505409241 2.49450516700745
2.76556825637817 2.46947121620178
2.7929310798645 2.44306087493896
2.82071471214294 2.41479110717773
2.84858751296997 2.3860342502594
2.87547993659973 2.36028909683228
2.90793466567993 2.32990312576294
2.93804883956909 2.30108857154846
2.96604919433594 2.27443861961365
2.99873089790344 2.24264907836914
3.02562761306763 2.21574544906616
3.05418086051941 2.18708395957947
3.08186745643616 2.15951490402222
3.10872840881348 2.13324403762817
3.13676404953003 2.10594630241394
3.16332125663757 2.08185625076294
3.18812942504883 2.06027388572693
3.21601724624634 2.03781175613403
3.24453282356262 2.01621532440186
3.27528309822083 1.99308443069458
3.30126643180847 1.97375166416168
3.33162045478821 1.95066392421722
3.35712289810181 1.93089842796326
3.38558673858643 1.90867245197296
3.41459965705872 1.88640081882477
3.44375658035278 1.86436307430267
3.47337651252747 1.84245419502258
3.50284123420715 1.82090318202972
3.53557109832764 1.7984527349472
3.56262516975403 1.78011429309845
3.5917661190033 1.76038539409637
3.62302613258362 1.73958420753479
3.65162324905396 1.72012007236481
3.67942261695862 1.70019626617432
3.70691180229187 1.67946398258209
3.73167753219604 1.65986967086792
3.76077342033386 1.63557910919189
3.78732752799988 1.61222994327545
3.81211066246033 1.59026479721069
3.84124541282654 1.56361794471741
3.86840081214905 1.53787446022034
3.89585423469543 1.51091039180756
3.92140674591064 1.48487758636475
3.95134830474854 1.45365536212921
3.97856330871582 1.42468357086182
4.00548839569092 1.3954746723175
4.02983140945435 1.36855638027191
4.05643129348755 1.33965122699738
4.08427047729492 1.31041657924652
4.11170721054077 1.2810525894165
4.13913679122925 1.25161564350128
4.1675271987915 1.22204911708832
4.19588279724121 1.19342148303986
4.22689342498779 1.16325414180756
4.2528395652771 1.13865029811859
4.28333330154419 1.11050510406494
4.31165504455566 1.08529496192932
4.34090900421143 1.06119775772095
4.37081003189087 1.03862822055817
4.3991904258728 1.02002596855164
4.4302978515625 1.00027680397034
};
\addlegendentry{TDOA}
\addplot [semithick, green!50.1960784313725!black]
table {%
4.24253749847412 1.26419174671173
4.22189235687256 1.29199981689453
4.19828271865845 1.3180468082428
4.1727728843689 1.33755648136139
4.14795541763306 1.35126042366028
4.12246465682983 1.36899530887604
4.09690856933594 1.39182114601135
4.07083129882812 1.41800093650818
4.04275465011597 1.44514012336731
4.01269865036011 1.47048163414001
3.98241782188416 1.49424493312836
3.96023726463318 1.5154709815979
3.94717025756836 1.53509473800659
3.93988490104675 1.55634367465973
3.93629050254822 1.58003795146942
3.92967748641968 1.60762691497803
3.91772675514221 1.63809204101562
3.89942932128906 1.66812777519226
3.87558126449585 1.69600963592529
3.84742283821106 1.72108960151672
3.81699705123901 1.74395406246185
3.7843177318573 1.76487827301025
3.75062227249146 1.78315567970276
3.716144323349 1.79966950416565
3.68113708496094 1.81434631347656
3.64533352851868 1.82697331905365
3.60894298553467 1.83653521537781
3.57186913490295 1.8429833650589
3.53401255607605 1.84761655330658
3.49581813812256 1.85328352451324
3.45775771141052 1.86303389072418
3.42217111587524 1.8773707151413
3.39289593696594 1.89824879169464
3.37401080131531 1.92490482330322
3.36556696891785 1.95516085624695
3.36462044715881 1.9866818189621
3.36509561538696 2.01956057548523
3.36136722564697 2.04813456535339
3.34671139717102 2.07644963264465
3.32445740699768 2.09621429443359
3.29687261581421 2.10948181152344
3.2664577960968 2.11665868759155
3.23588848114014 2.11715316772461
3.20684552192688 2.11155009269714
3.17919373512268 2.10546922683716
3.15110301971436 2.10503435134888
3.12269115447998 2.11195182800293
3.09426093101501 2.1245641708374
3.06560707092285 2.14009761810303
3.03656411170959 2.15733909606934
3.00750803947449 2.17592763900757
2.97822093963623 2.19507527351379
2.94912481307983 2.21504974365234
2.92051267623901 2.23580574989319
2.89210796356201 2.25739359855652
2.86412811279297 2.27988982200623
2.8384850025177 2.30457806587219
2.81755495071411 2.33257722854614
2.79788589477539 2.36150026321411
2.77676773071289 2.39003944396973
2.75407886505127 2.41784429550171
2.73007392883301 2.44479990005493
2.70642971992493 2.46787858009338
2.6790246963501 2.4941930770874
2.65300726890564 2.51794672012329
2.62709403038025 2.54139852523804
2.60121393203735 2.56539750099182
2.57591795921326 2.59062814712524
2.55055665969849 2.61696290969849
2.52433013916016 2.64411282539368
2.49662089347839 2.67050290107727
2.46744632720947 2.69571256637573
2.43723249435425 2.7203893661499
2.4067280292511 2.744957447052
2.37740468978882 2.77064490318298
2.35185289382935 2.79753255844116
2.33674883842468 2.82015895843506
2.33656549453735 2.83608102798462
2.34124755859375 2.84833979606628
2.3473653793335 2.85925245285034
2.35939288139343 2.87197685241699
2.37606143951416 2.88460254669189
2.39685893058777 2.89240097999573
2.4125702381134 2.89585542678833
2.42149138450623 2.89967012405396
2.42527461051941 2.90522456169128
2.42663240432739 2.90667057037354
2.42602324485779 2.90390038490295
2.41630291938782 2.89822578430176
2.39808917045593 2.89097571372986
2.37415671348572 2.88465881347656
2.34674715995789 2.87938523292542
2.31721973419189 2.87498354911804
2.29025530815125 2.87147855758667
2.27004432678223 2.86836767196655
2.2702693939209 2.86679196357727
2.28560733795166 2.86586427688599
2.30927777290344 2.86324262619019
2.32911324501038 2.85945773124695
2.34307527542114 2.85564184188843
2.35148143768311 2.85336351394653
2.34902048110962 2.85438990592957
2.34584259986877 2.85746216773987
2.34684062004089 2.86501741409302
2.34473776817322 2.87603688240051
2.34146690368652 2.88208866119385
2.34369540214539 2.87861943244934
2.34639286994934 2.87158250808716
2.34939503669739 2.86779356002808
2.35142874717712 2.86830544471741
2.34596371650696 2.87161350250244
2.33185267448425 2.87683129310608
2.31208944320679 2.88431143760681
2.28878712654114 2.89365243911743
2.26353693008423 2.90457153320312
2.23746728897095 2.91716170310974
2.21290683746338 2.93387794494629
2.18881988525391 2.95391273498535
2.16250371932983 2.97236967086792
2.13425397872925 2.9880473613739
2.10462617874146 3.00039386749268
2.07408952713013 3.01084232330322
2.04613542556763 3.01943254470825
2.01307845115662 3.02770829200745
1.98306548595428 3.0349178314209
1.95384657382965 3.04395866394043
1.92400550842285 3.05539584159851
1.89412176609039 3.0690393447876
1.86625385284424 3.08407282829285
1.83275842666626 3.10341334342957
1.80164170265198 3.12438750267029
1.77176237106323 3.14847564697266
1.7429211139679 3.17504787445068
1.71570813655853 3.20370388031006
1.69033634662628 3.23390412330627
1.6673172712326 3.26478171348572
1.64473223686218 3.29459285736084
1.6212705373764 3.32307934761047
1.59758353233337 3.35172867774963
1.57341516017914 3.38083267211914
1.5466057062149 3.40811538696289
1.51652956008911 3.42998814582825
1.48371028900146 3.44621300697327
1.44904208183289 3.45745277404785
1.41393876075745 3.46432638168335
1.3784214258194 3.46736216545105
1.34363281726837 3.46620059013367
1.31278109550476 3.46215200424194
1.2795467376709 3.45278406143188
1.2552433013916 3.44097924232483
1.24222421646118 3.42812848091125
1.2419205904007 3.41459393501282
1.25103998184204 3.39987397193909
1.26664292812347 3.384197473526
1.2865275144577 3.3674521446228
1.30907928943634 3.35077095031738
1.33381462097168 3.33389234542847
1.35945463180542 3.31727695465088
1.38626766204834 3.30004072189331
1.41263067722321 3.28275656700134
1.43853843212128 3.26542282104492
1.46418428421021 3.24683499336243
1.4900918006897 3.22651743888855
1.51670598983765 3.20528244972229
1.54487550258636 3.18461966514587
1.57342624664307 3.16461110115051
1.60404717922211 3.14476609230042
1.63564300537109 3.12540912628174
1.66921615600586 3.10663294792175
1.70380568504333 3.08883905410767
1.73891079425812 3.07122039794922
1.7735732793808 3.05331063270569
1.8073992729187 3.03578591346741
1.83973395824432 3.01823282241821
1.87170386314392 3.00124478340149
1.90279459953308 2.98470163345337
1.9340523481369 2.96881508827209
1.965460896492 2.9532196521759
1.99860715866089 2.93738627433777
2.0326840877533 2.9205949306488
2.06660389900208 2.90186047554016
2.10099101066589 2.88172793388367
2.13520169258118 2.86102390289307
2.1692316532135 2.84021258354187
2.20216584205627 2.81889176368713
2.23548913002014 2.79785299301147
2.26847004890442 2.77665710449219
2.30131459236145 2.75520372390747
2.3337996006012 2.73343300819397
2.36594724655151 2.71115708351135
2.39791035652161 2.68826675415039
2.42810845375061 2.66451334953308
2.45684313774109 2.63933324813843
2.4849681854248 2.61253070831299
2.51267743110657 2.58390355110168
2.53972840309143 2.55358791351318
2.56598567962646 2.52223134040833
2.59103012084961 2.48974466323853
2.61497235298157 2.45645308494568
2.63856720924377 2.42197489738464
2.66281318664551 2.38611125946045
2.68652009963989 2.34947919845581
2.71042013168335 2.31378769874573
2.73521089553833 2.27901244163513
2.75942516326904 2.24831342697144
2.78648972511292 2.21568942070007
2.81688451766968 2.18144345283508
2.84698534011841 2.15150094032288
2.87797522544861 2.12246704101562
2.91000008583069 2.09435844421387
2.94355821609497 2.06730961799622
2.97748780250549 2.04102683067322
3.01165151596069 2.01625490188599
3.0455265045166 1.99348044395447
3.07931327819824 1.97118175029755
3.11275887489319 1.94913446903229
3.14561152458191 1.92758083343506
3.17832612991333 1.90695381164551
3.21111130714417 1.88642287254333
3.24427151679993 1.86575794219971
3.27733826637268 1.84552979469299
3.30760669708252 1.82671201229095
3.34321546554565 1.80440843105316
3.37567758560181 1.78438556194305
3.40800356864929 1.76496851444244
3.44013023376465 1.7453989982605
3.47177481651306 1.72540998458862
3.50254011154175 1.7051602602005
3.53307390213013 1.68468773365021
3.56086921691895 1.66581892967224
3.59338164329529 1.64344704151154
3.62289929389954 1.62227404117584
3.65268564224243 1.60015189647675
3.68249487876892 1.57704925537109
3.71238875389099 1.55258738994598
3.74266934394836 1.52685070037842
3.77288293838501 1.49986839294434
3.80252408981323 1.47189509868622
3.83174347877502 1.44292783737183
3.86074352264404 1.41336393356323
3.88962459564209 1.38380682468414
3.91844797134399 1.35464930534363
3.94793391227722 1.32569015026093
3.97726106643677 1.29724299907684
4.00613784790039 1.27019786834717
4.03472518920898 1.24438107013702
4.06287336349487 1.22093403339386
4.09080410003662 1.19859886169434
4.11966276168823 1.17627954483032
4.14874887466431 1.15431392192841
4.17886638641357 1.13297009468079
4.20940542221069 1.11318635940552
4.24059009552002 1.09466338157654
4.27115917205811 1.07849156856537
4.30146455764771 1.06368136405945
4.33268356323242 1.04998815059662
4.3643102645874 1.03464984893799
4.39562749862671 1.01952695846558
4.42628765106201 1.00445914268494
4.45405387878418 0.990319013595581
4.47668838500977 0.977408528327942
};
\addlegendentry{Dynamic}
\addplot [semithick, black, mark=asterisk, mark size=3, mark options={solid}]
table {%
0 0
};
\addlegendentry{UWB Nodes}
\addplot [semithick, black, mark=asterisk, mark size=3, mark options={solid}]
table {%
3.04 0
};
\addplot [semithick, black, mark=asterisk, mark size=3, mark options={solid}]
table {%
6.04727935791016 1.98065876960754
};
\addplot [semithick, black, mark=asterisk, mark size=3, mark options={solid}]
table {%
2.81377625465393 4.1934027671814
};
\addplot [semithick, black, mark=asterisk, mark size=3, mark options={solid}]
table {%
3.39600872993469 2.20860457420349
};
\end{axis}

\end{tikzpicture}

%% file: tex/line_roles3.tex
\begin{tikzpicture}

\definecolor{color0}{rgb}{0.12156862745098,0.466666666666667,0.705882352941177}
\definecolor{color1}{rgb}{1,0.498039215686275,0.0549019607843137}

\begin{axis}[
    height=\figureheight,
    width=\figurewidth,
    axis line style={white},
    tick align=outside,
    tick pos=left,
    x grid style={white},
    xtick style={color=black},
    y grid style={white!90!black},
    ymajorgrids,
    ytick style={color=black},
    scaled y ticks = false,
    legend cell align={left},
    legend style={
      fill opacity=0.8,
      draw opacity=0.666,
      text opacity=1,
      at={(0.97,0.97)},
      anchor=north east,
      draw=white!80!black
    },
    tick align=outside,
    tick pos=left,
    %
    %
    %
    %
    %
    %
    %
    %
xlabel={Time[s]},
xmajorgrids,
xmin=-2.64383374452591, xmax=55.8271645188332,
xtick style={color=black},
ymajorgrids,
ymin=-1.2, ymax=1.8,
ytick={-1,1},
yticklabels={$L$, $A$}
]
\addplot [green!50.1960784313725!black]
table {%
0.0139389038085938 1
0.2099289894104 -1
0.4131019115448 -1
0.610692977905273 -1
0.81158185005188 -1
1.00969386100769 -1
1.21090388298035 -1
1.41517400741577 -1
1.61356091499329 -1
1.81181502342224 -1
2.00968289375305 -1
2.20964694023132 -1
2.41426587104797 -1
2.6090350151062 -1
2.81001996994019 -1
3.01112198829651 -1
3.21070504188538 -1
3.41446900367737 -1
3.61236786842346 -1
3.81393599510193 -1
4.01112294197083 -1
4.21125888824463 -1
4.41187191009521 -1
4.61538505554199 -1
4.81142687797546 -1
5.00895404815674 -1
5.20967698097229 -1
5.4124698638916 -1
5.61258006095886 -1
5.81203699111938 -1
6.01078200340271 -1
6.20969891548157 -1
6.41016888618469 -1
6.61456704139709 -1
6.80991005897522 -1
7.00994896888733 -1
7.21042084693909 -1
7.4139289855957 -1
7.6117570400238 -1
7.81206488609314 -1
8.0152108669281 -1
8.21269583702087 -1
8.41253685951233 -1
8.65383696556091 -1
8.81301498413086 -1
9.01152801513672 -1
9.21503305435181 -1
9.40977597236633 -1
9.61222791671753 -1
9.8101658821106 -1
10.0098278522491 -1
10.2126288414001 -1
10.4121668338776 -1
10.6217279434204 -1
10.8104269504547 -1
11.010143995285 -1
11.2185909748077 -1
11.4109709262848 -1
11.6098279953003 -1
11.813206911087 -1
12.0134320259094 -1
12.2121000289917 -1
12.414675951004 -1
12.612203836441 -1
12.8098959922791 -1
13.0127580165863 -1
13.2145888805389 -1
13.4149329662323 -1
13.6222908496857 -1
13.8120110034943 -1
14.0135660171509 -1
14.2098269462585 -1
14.4124388694763 -1
14.6100718975067 -1
14.8149418830872 -1
15.0123748779297 -1
15.2110059261322 -1
15.4093379974365 -1
15.6124980449677 -1
15.810201883316 -1
16.0101299285889 -1
16.2190699577332 1
16.4133539199829 -1
16.6102108955383 -1
16.8101658821106 -1
17.0138740539551 -1
17.2098898887634 1
17.4133908748627 -1
17.6100249290466 -1
17.813628911972 -1
18.0124108791351 -1
18.2104849815369 1
18.4113478660583 -1
18.6116080284119 -1
18.8127310276031 -1
19.0108239650726 -1
19.213506937027 -1
19.4193489551544 -1
19.6100368499756 -1
19.8105640411377 -1
20.0097420215607 -1
20.2118968963623 1
20.4108018875122 -1
20.6112809181213 -1
20.8114318847656 1
21.0112998485565 -1
21.2102708816528 -1
21.4120509624481 1
21.6105630397797 -1
21.8138499259949 -1
22.0097739696503 1
22.2109129428864 -1
22.412122964859 -1
22.611319065094 1
22.8115658760071 -1
23.0136768817902 -1
23.2102649211884 -1
23.4100329875946 -1
23.6106338500977 -1
23.8111438751221 -1
24.0140829086304 -1
24.2097878456116 -1
24.4130878448486 -1
24.6108219623566 -1
24.8126590251923 -1
25.0131809711456 -1
25.212779045105 -1
25.4152638912201 -1
25.6160988807678 -1
25.8130609989166 -1
26.0134048461914 -1
26.2102448940277 -1
26.4116239547729 -1
26.6119868755341 -1
26.8131608963013 -1
27.0142028331757 -1
27.2128989696503 -1
27.4114220142365 -1
27.6092698574066 -1
27.8140389919281 -1
28.019387960434 -1
28.2108108997345 -1
28.4121999740601 -1
28.6107349395752 1
28.8109049797058 1
29.009819984436 1
29.213222026825 1
29.4116940498352 1
29.6097149848938 1
29.8114619255066 1
30.0104868412018 1
30.2110910415649 1
30.4156730175018 1
30.6222629547119 1
30.8107960224152 1
31.0179870128632 1
31.2110018730164 1
31.4102599620819 1
31.6118190288544 1
31.8095760345459 1
32.0169398784637 1
32.2127108573914 1
32.4097418785095 1
32.6138648986816 1
32.8137030601501 1
33.0101149082184 1
33.2122659683228 1
33.4114589691162 1
33.6112439632416 1
33.8102910518646 1
34.0123429298401 1
34.2153449058533 1
34.4099669456482 1
34.61066198349 1
34.8109040260315 1
35.0104858875275 1
35.2128689289093 -1
35.4120650291443 -1
35.6122539043427 1
35.8126530647278 -1
36.0154340267181 -1
36.2160668373108 -1
36.4154069423676 1
36.615483045578 -1
36.8120439052582 -1
37.0124268531799 -1
37.2116169929504 -1
37.4107580184937 -1
37.6098139286041 -1
37.8116629123688 1
38.0136299133301 -1
38.2114000320435 -1
38.4146828651428 -1
38.61119389534 -1
38.8120498657227 -1
39.0109279155731 -1
39.2124729156494 1
39.4105689525604 -1
39.6127910614014 -1
39.8124430179596 -1
40.0125088691711 -1
40.2148559093475 -1
40.4113969802856 -1
40.6119349002838 -1
40.8105778694153 -1
41.0101380348206 -1
41.211837053299 -1
41.412458896637 -1
41.6110198497772 -1
41.8128578662872 -1
42.0203759670258 -1
42.2127139568329 -1
42.4105229377747 -1
42.6124320030212 -1
42.818207025528 -1
43.0136640071869 -1
43.2122218608856 -1
43.410472869873 -1
43.6131389141083 -1
43.8114998340607 -1
44.0138909816742 -1
44.2163879871368 -1
44.4141359329224 -1
44.609561920166 -1
44.8144068717957 -1
45.0115330219269 -1
45.2122099399567 -1
45.4112799167633 -1
45.6110389232635 -1
45.8114080429077 -1
46.0147449970245 -1
46.2176229953766 -1
46.4150178432465 -1
46.6110439300537 -1
46.8111250400543 -1
47.0133469104767 -1
47.2145409584045 -1
47.4107539653778 -1
47.6100249290466 -1
47.8127019405365 -1
48.0129489898682 -1
48.2106988430023 -1
48.4111449718475 -1
48.6116988658905 -1
48.8127019405365 -1
49.0141048431396 -1
49.2127130031586 -1
49.4103739261627 -1
49.6146528720856 -1
49.8148880004883 -1
50.0143549442291 -1
50.2121820449829 -1
50.4120628833771 -1
50.6131608486176 -1
50.8125250339508 -1
51.0111248493195 -1
51.2111048698425 -1
51.4102909564972 -1
51.6127648353577 -1
51.8108539581299 -1
52.0105519294739 -1
52.2107560634613 -1
52.4147789478302 -1
52.6099720001221 -1
52.8111529350281 -1
53.0132298469543 -1
};
\addlegendentry{Dynamic}
\addplot [color1]
table {%
0.169052839279175 -1
0.369066953659058 -1
0.569289922714233 -1
0.769301891326904 -1
0.969346046447754 -1
1.16953897476196 -1
1.36944389343262 -1
1.56961584091187 -1
1.76951098442078 -1
1.96988105773926 -1
2.16937398910522 -1
2.36954402923584 -1
2.56936192512512 -1
2.77165794372559 -1
2.96917796134949 -1
3.16924405097961 -1
3.36970496177673 -1
3.5701789855957 -1
3.76919984817505 -1
3.96956205368042 -1
4.16967797279358 -1
4.36946392059326 -1
4.56969285011292 -1
4.76910495758057 -1
4.96952700614929 -1
5.16948103904724 -1
5.36927485466003 -1
5.56901502609253 -1
5.76932597160339 -1
5.96942806243896 -1
6.1693069934845 -1
6.36935591697693 -1
6.56944894790649 -1
6.76972603797913 -1
6.96900486946106 -1
7.16906785964966 -1
7.36926794052124 -1
7.57024192810059 -1
7.77219390869141 -1
7.96963500976562 -1
8.16936492919922 -1
8.37000799179077 -1
8.56941795349121 -1
8.77232003211975 -1
8.96931195259094 -1
9.16982889175415 -1
9.36973905563354 -1
9.56973385810852 -1
9.77013206481934 -1
9.969801902771 -1
10.1696028709412 -1
10.3700850009918 -1
10.569543838501 -1
10.769376039505 -1
10.9698288440704 -1
11.1697459220886 -1
11.3696830272675 -1
11.5696768760681 -1
11.7697010040283 -1
11.9695138931274 -1
12.1698739528656 -1
12.3697738647461 -1
12.5694029331207 -1
12.7691519260406 -1
12.9694139957428 -1
13.1694209575653 -1
13.3689439296722 -1
13.5691640377045 -1
13.7690320014954 -1
13.9691779613495 -1
14.1695909500122 -1
14.369292974472 -1
14.5698988437653 -1
14.769170999527 -1
14.9748289585114 -1
15.1693549156189 -1
15.3695669174194 -1
15.569620847702 -1
15.7693228721619 -1
15.9691140651703 -1
16.1701278686523 -1
16.3701529502869 -1
16.5696079730988 -1
16.7693829536438 -1
16.9691970348358 -1
17.1704130172729 -1
17.3698258399963 -1
17.5690770149231 -1
17.7693929672241 -1
17.9694919586182 -1
18.1692550182343 -1
18.369619846344 -1
18.5697400569916 -1
18.7693428993225 -1
18.9692070484161 -1
19.1693890094757 -1
19.3694038391113 -1
19.5691120624542 -1
19.7697539329529 -1
19.969085931778 -1
20.1694140434265 -1
20.3692829608917 -1
20.5691349506378 -1
20.7696979045868 -1
20.9694209098816 -1
21.1694338321686 -1
21.3697638511658 -1
21.5692429542542 -1
21.7693240642548 -1
21.9693419933319 -1
22.1691629886627 -1
22.3693690299988 -1
22.5695819854736 -1
22.7696499824524 -1
22.9692149162292 -1
23.1696400642395 -1
23.3698799610138 -1
23.5693459510803 -1
23.7694189548492 -1
23.9731538295746 -1
24.1695408821106 -1
24.3694360256195 -1
24.5731790065765 -1
24.7691049575806 -1
24.969388961792 -1
25.1693439483643 -1
25.3693628311157 -1
25.57017993927 -1
25.771880865097 -1
25.969388961792 -1
26.1693320274353 -1
26.3698379993439 -1
26.5693769454956 -1
26.7695770263672 -1
26.9695129394531 -1
27.1694378852844 -1
27.3700580596924 -1
27.5694470405579 -1
27.769690990448 -1
27.9696619510651 -1
28.1699960231781 -1
28.3699469566345 -1
28.5697560310364 -1
28.7697110176086 -1
28.9702179431915 -1
29.1699769496918 -1
29.3697719573975 -1
29.5697679519653 -1
29.7696058750153 -1
29.9693388938904 -1
30.1693439483643 1
30.3695859909058 1
30.5694949626923 1
30.7700250148773 1
30.9696819782257 1
31.1694090366364 1
31.3694360256195 1
31.5693080425262 1
31.7696049213409 1
31.9696340560913 1
32.1694209575653 -1
32.3694999217987 -1
32.5692038536072 -1
32.7696769237518 -1
32.9700658321381 -1
33.1740958690643 -1
33.3694539070129 -1
33.5697729587555 -1
33.7693939208984 -1
33.9690079689026 -1
34.1690299510956 -1
34.3802809715271 -1
34.5776510238647 -1
34.7693319320679 -1
34.9695999622345 -1
35.1692478656769 -1
35.3693079948425 -1
35.5695989131927 -1
35.7723429203033 -1
35.969132900238 -1
36.1693530082703 -1
36.3690578937531 -1
36.5694358348846 -1
36.7693610191345 -1
36.9695119857788 -1
37.1694910526276 -1
37.3691489696503 -1
37.5694348812103 -1
37.7691369056702 -1
37.9693179130554 -1
38.1692600250244 -1
38.3691229820251 -1
38.5693218708038 -1
38.7696349620819 -1
38.9694650173187 -1
39.1693730354309 -1
39.3690469264984 -1
39.5696940422058 -1
39.7692940235138 -1
39.9691829681396 -1
40.1698880195618 -1
40.3693130016327 -1
40.5694980621338 -1
40.771880865097 -1
40.9690170288086 -1
41.1693649291992 -1
41.3721840381622 -1
41.5695569515228 -1
41.7691049575806 -1
41.9691400527954 -1
42.1698479652405 -1
42.3693549633026 -1
42.5691120624542 -1
42.7693479061127 -1
42.9695248603821 -1
43.1695778369904 -1
43.3695268630981 -1
43.5693159103394 -1
43.7697780132294 -1
43.9696369171143 -1
44.1694209575653 -1
44.3700039386749 -1
44.5692460536957 -1
44.7692649364471 -1
44.9699659347534 -1
45.1693348884583 -1
45.3694319725037 -1
45.569718837738 -1
45.7697238922119 -1
45.9691450595856 -1
46.1696448326111 -1
46.3697700500488 -1
46.5696969032288 -1
46.7697360515594 -1
46.9700000286102 -1
47.1699459552765 -1
47.3696029186249 -1
47.5705449581146 -1
47.7691919803619 -1
47.9694800376892 -1
48.17009806633 -1
48.3703999519348 -1
48.5697400569916 -1
48.7701790332794 -1
48.9700028896332 -1
49.1701819896698 -1
49.3709149360657 -1
49.5699238777161 -1
49.7705140113831 -1
49.9700548648834 -1
50.1692318916321 -1
50.3692450523376 -1
50.5696699619293 -1
50.7691209316254 -1
50.9698169231415 -1
51.1694839000702 -1
51.3692879676819 -1
51.5691180229187 -1
51.7698829174042 -1
51.9695339202881 -1
52.1691999435425 -1
52.369206905365 -1
52.5697400569916 -1
52.7700970172882 -1
52.9692950248718 -1
53.1693918704987 -1
};
\addlegendentry{MOCAP}
\end{axis}

\end{tikzpicture}

%% file: tex/box_roles3.tex
\begin{tikzpicture}

\definecolor{color0}{rgb}{0.12156862745098,0.466666666666667,0.705882352941177}
\definecolor{color1}{rgb}{1,0.498039215686275,0.0549019607843137}

\begin{axis}[
    height=\figureheight,
    width=\figurewidth,
    axis line style={white},
    tick align=outside,
    tick pos=left,
    x grid style={white},
    xtick style={color=black},
    y grid style={white!90!black},
    ymajorgrids,
    ytick style={color=black},
    scaled y ticks = false,
    legend cell align={left},
    legend style={
      fill opacity=0.8,
      draw opacity=0.666,
      text opacity=1,
      at={(0.97,0.97)},
      anchor=north east,
      draw=white!80!black
    },
    tick align=outside,
    tick pos=left,
    %
    %
    %
    %
    %
    %
    %
    %
xlabel={Time[s]},
xmin=-4.50419826507568, xmax=97.0042273521423,
ymin=-1.2, ymax=1.2,
ytick={-1,1},
yticklabels={$L$, $A$}
]
\addplot [green!50.1960784313725!black]
table {%
0.109821081161499 -1
0.303874015808105 1
0.504191160202026 -1
0.70611310005188 -1
0.90673303604126 1
1.1071469783783 -1
1.30351305007935 -1
1.50445294380188 1
1.70380210876465 -1
1.90616703033447 -1
2.10580110549927 1
2.30263304710388 -1
2.50602912902832 -1
2.70392394065857 1
2.90708804130554 -1
3.10915994644165 -1
3.3068699836731 1
3.50452399253845 -1
3.70697712898254 -1
3.90368795394897 1
4.1071879863739 -1
4.30649614334106 -1
4.5094850063324 1
4.70572400093079 -1
4.90470814704895 -1
5.10376811027527 1
5.30470609664917 -1
5.50485706329346 -1
5.7144730091095 1
5.91227102279663 -1
6.10610008239746 -1
6.32274413108826 1
6.50356316566467 -1
6.7050199508667 -1
6.9031810760498 1
7.10606694221497 -1
7.30594396591187 -1
7.5042130947113 -1
7.70338702201843 -1
7.90371799468994 -1
8.10486102104187 -1
8.30572414398193 -1
8.50700712203979 -1
8.70667600631714 -1
8.90528297424316 -1
9.10622501373291 1
9.30341005325317 -1
9.50941610336304 -1
9.70531606674194 1
9.90394806861877 -1
10.1050460338593 -1
10.3038580417633 -1
10.517557144165 -1
10.7041871547699 -1
10.9117169380188 -1
11.1111121177673 -1
11.3124580383301 -1
11.5082819461823 -1
11.7043941020966 -1
11.9044671058655 -1
12.1092491149902 -1
12.3061590194702 -1
12.505156993866 -1
12.7145359516144 -1
12.9051930904388 -1
13.105532169342 -1
13.3039770126343 -1
13.5038731098175 -1
13.704980134964 -1
13.9069011211395 -1
14.1038949489594 -1
14.312164068222 -1
14.5085909366608 -1
14.707603931427 -1
14.9077091217041 -1
15.1102781295776 -1
15.3081810474396 -1
15.5049710273743 -1
15.7045631408691 -1
15.9044761657715 -1
16.1109609603882 -1
16.3072650432587 -1
16.5058281421661 -1
16.7063989639282 -1
16.9044179916382 -1
17.1034259796143 -1
17.3045260906219 -1
17.5040650367737 -1
17.7049930095673 -1
17.9047901630402 -1
18.1085519790649 -1
18.3257319927216 -1
18.5064740180969 -1
18.7079391479492 -1
18.9044671058655 -1
19.1072380542755 -1
19.306097984314 -1
19.5061130523682 -1
19.708419084549 -1
19.9046230316162 -1
20.1058821678162 -1
20.3076391220093 -1
20.5085730552673 -1
20.7130219936371 -1
20.9091010093689 -1
21.1085889339447 -1
21.3072259426117 -1
21.5063819885254 -1
21.7056131362915 -1
21.9046850204468 -1
22.1067471504211 -1
22.3069710731506 -1
22.5059750080109 -1
22.7060611248016 -1
22.9108030796051 1
23.1081709861755 -1
23.3061571121216 -1
23.5067579746246 1
23.7036111354828 1
23.9068360328674 1
24.1048099994659 1
24.3091461658478 1
24.5046679973602 1
24.7064731121063 1
24.9057130813599 1
25.1083109378815 1
25.3048329353333 1
25.5081541538239 -1
25.704950094223 -1
25.9083890914917 1
26.1074500083923 -1
26.3038041591644 -1
26.5071289539337 1
26.7056250572205 -1
26.904217004776 -1
27.1062800884247 -1
27.3070869445801 -1
27.5079340934753 -1
27.7035210132599 -1
27.903813123703 -1
28.1053211688995 -1
28.3064341545105 -1
28.5099251270294 -1
28.7042350769043 -1
28.9082429409027 -1
29.1045470237732 -1
29.3040101528168 -1
29.5173089504242 -1
29.7041521072388 -1
29.9057829380035 -1
30.1049540042877 -1
30.3177189826965 -1
30.5052061080933 -1
30.7050721645355 -1
30.9054510593414 -1
31.1075971126556 -1
31.3039700984955 -1
31.5050051212311 -1
31.7046940326691 -1
31.9045841693878 -1
32.1065440177917 1
32.3085949420929 -1
32.5087089538574 -1
32.7038531303406 1
32.9049489498138 -1
33.1070699691772 -1
33.3067049980164 -1
33.5045680999756 -1
33.7095201015472 1
33.9056749343872 -1
34.1094310283661 -1
34.305869102478 1
34.5055420398712 -1
34.7065989971161 -1
34.9053020477295 1
35.1056010723114 -1
35.3055341243744 -1
35.504723072052 1
35.707545042038 -1
35.9102711677551 -1
36.1055719852448 1
36.3057720661163 1
36.5090939998627 1
36.7047591209412 1
36.907772064209 1
37.1119589805603 -1
37.3083341121674 -1
37.5038549900055 1
37.7067670822144 1
37.9232411384583 1
38.1112270355225 1
38.3051600456238 1
38.5057289600372 1
38.7081711292267 1
38.9083931446075 1
39.1066401004791 1
39.3115041255951 1
39.51092004776 1
39.704391002655 1
39.9054350852966 1
40.1054611206055 1
40.3058669567108 1
40.5044760704041 1
40.706127166748 1
40.937301158905 1
41.1069951057434 1
41.3044281005859 1
41.5090351104736 1
41.7063839435577 1
41.9080140590668 1
42.1046481132507 1
42.3108341693878 1
42.509003162384 1
42.705817937851 1
42.9098520278931 1
43.1082799434662 1
43.3071000576019 1
43.5064680576324 1
43.7081940174103 1
43.9061281681061 1
44.1062369346619 1
44.3092019557953 1
44.5099060535431 1
44.7347919940948 1
44.9046449661255 1
45.108528137207 1
45.3138480186462 1
45.5100240707397 1
45.7051730155945 1
45.9072771072388 1
46.1058149337769 1
46.3063180446625 1
46.5067579746246 1
46.7048089504242 1
46.909863948822 1
47.1068069934845 1
47.3107991218567 1
47.5066630840302 1
47.7109410762787 1
47.9090621471405 1
48.108570098877 1
48.3055610656738 1
48.5066511631012 1
48.7053799629211 1
48.9233050346375 1
49.1048810482025 1
49.3057360649109 1
49.5093729496002 1
49.7078449726105 1
49.9062991142273 1
50.1076221466064 1
50.3092150688171 1
50.5074880123138 1
50.7086310386658 1
50.9057800769806 1
51.1065239906311 1
51.3179841041565 1
51.510134935379 1
51.7077610492706 1
51.9118521213531 1
52.1126799583435 1
52.3124520778656 1
52.5094089508057 1
52.7040779590607 1
52.9089529514313 1
53.1065890789032 1
53.3059091567993 1
53.506098985672 1
53.7075810432434 1
53.9050951004028 1
54.1074600219727 -1
54.3149001598358 -1
54.5077879428864 -1
54.7066760063171 -1
54.9063580036163 -1
55.1087291240692 1
55.3051731586456 -1
55.5094330310822 -1
55.7080149650574 1
55.9063611030579 1
56.1061379909515 1
56.3065180778503 1
56.5050201416016 1
56.7082040309906 1
56.9068131446838 1
57.1058051586151 1
57.3089330196381 1
57.5076661109924 1
57.7049419879913 1
57.9056911468506 1
58.1069531440735 1
58.3060960769653 1
58.507050037384 1
58.7061281204224 -1
58.9087729454041 -1
59.1117630004883 -1
59.3103079795837 -1
59.5131220817566 -1
59.7082419395447 -1
59.9062480926514 -1
60.1051290035248 -1
60.31325507164 -1
60.5091519355774 -1
60.7080111503601 -1
60.9077999591827 -1
61.1081700325012 -1
61.3057689666748 -1
61.5067870616913 -1
61.7103321552277 1
61.9082310199738 1
62.1086781024933 1
62.3086230754852 -1
62.5084619522095 1
62.7067511081696 1
62.9072620868683 1
63.108381986618 1
63.3104720115662 1
63.5149171352386 1
63.7115180492401 1
63.9081900119781 1
64.1044199466705 1
64.3095390796661 1
64.5078520774841 1
64.7065660953522 1
64.9098031520844 -1
65.1210739612579 -1
65.3076660633087 -1
65.507425069809 -1
65.7084989547729 -1
65.913388967514 -1
66.1066391468048 -1
66.3179891109467 1
66.5114691257477 1
66.7102439403534 1
66.9059901237488 1
67.1175510883331 1
67.3076529502869 1
67.5115959644318 1
67.7058429718018 1
67.9112470149994 1
68.1130859851837 1
68.3064970970154 1
68.5106589794159 1
68.7046830654144 1
68.9075150489807 1
69.1090619564056 1
69.3099241256714 1
69.5058481693268 1
69.7060890197754 1
69.9058001041412 1
70.1202001571655 1
70.3082089424133 1
70.5100100040436 1
70.7064490318298 1
70.9093890190125 1
71.1082670688629 1
71.3128991127014 1
71.5102310180664 1
71.7077560424805 1
71.9067680835724 1
72.107143163681 1
72.3105731010437 1
72.5072801113129 1
72.7114551067352 1
72.9092130661011 1
73.1110849380493 1
73.3112089633942 1
73.5077021121979 1
73.7152960300446 1
73.9085099697113 1
74.1079709529877 1
74.3051221370697 1
74.5093369483948 1
74.7090539932251 1
74.9086589813232 1
75.1128289699554 1
75.3107199668884 1
75.5076539516449 1
75.7075769901276 1
75.9097609519958 1
76.1056790351868 1
76.3061311244965 1
76.5096881389618 1
76.7050790786743 1
76.9100770950317 1
77.1054580211639 1
77.3080761432648 1
77.5098531246185 1
77.7068600654602 1
77.9063560962677 1
78.1074481010437 1
78.311439037323 1
78.5075590610504 1
78.7079780101776 1
78.9091761112213 1
79.1077320575714 1
79.3095600605011 1
79.5108561515808 1
79.7058069705963 1
79.911318063736 1
80.1072521209717 1
80.3154511451721 1
80.5087220668793 1
80.7078931331635 1
80.9103381633759 1
81.1103200912476 1
81.3087680339813 1
81.5091280937195 1
81.7117080688477 1
81.9106440544128 1
82.108992099762 1
82.312108039856 1
82.5084390640259 1
82.7112100124359 1
82.9066050052643 1
83.1111040115356 1
83.3091361522675 1
83.5068249702454 1
83.7096199989319 1
83.9066491127014 1
84.1105959415436 1
84.3243701457977 -1
84.5086269378662 -1
84.7176101207733 -1
84.9108710289001 1
85.1123631000519 -1
85.3191621303558 -1
85.5072610378265 1
85.708389043808 -1
85.9085831642151 -1
86.1097421646118 1
86.3104829788208 -1
86.5102331638336 -1
86.7080330848694 1
86.908350944519 -1
87.1124219894409 -1
87.3250689506531 1
87.5072889328003 -1
87.7093911170959 -1
87.907655954361 1
88.1063041687012 -1
88.3099181652069 -1
88.5096311569214 1
88.7085001468658 -1
88.908075094223 -1
89.1119530200958 1
89.322811126709 -1
89.5074119567871 -1
89.7084219455719 1
89.9070200920105 -1
90.1066751480103 -1
90.3155710697174 1
90.505795955658 -1
90.707258939743 -1
90.9055049419403 1
91.1091830730438 -1
91.3133800029755 -1
91.5056231021881 1
91.709114074707 -1
91.9115920066833 -1
92.106348991394 1
};
\addlegendentry{Dynamic}
\addplot [color1]
table {%
0.18969202041626 -1
0.38935112953186 -1
0.589781045913696 -1
0.789828062057495 -1
0.989954948425293 -1
1.18965005874634 -1
1.38964915275574 -1
1.58951997756958 -1
1.78951907157898 -1
1.98977613449097 -1
2.18966007232666 -1
2.38977909088135 -1
2.58975100517273 -1
2.78927111625671 -1
2.9894380569458 -1
3.18941903114319 -1
3.38972401618958 -1
3.58993697166443 -1
3.78989696502686 -1
3.99045705795288 -1
4.19059896469116 -1
4.39097499847412 -1
4.58948516845703 -1
4.79004502296448 -1
4.98991894721985 -1
5.18990612030029 -1
5.38993716239929 -1
5.59153413772583 -1
5.78982710838318 -1
5.99032807350159 -1
6.19017100334167 -1
6.38995409011841 -1
6.58984899520874 -1
6.78978800773621 -1
6.98951196670532 -1
7.18991112709045 -1
7.39025402069092 -1
7.58999800682068 -1
7.78992295265198 -1
7.99036812782288 -1
8.19126200675964 -1
8.39043998718262 -1
8.59096908569336 -1
8.79076409339905 -1
8.99107193946838 -1
9.19104099273682 -1
9.39006614685059 -1
9.59000897407532 -1
9.790118932724 -1
9.98995995521545 -1
10.1893689632416 -1
10.3901641368866 -1
10.5898630619049 -1
10.7896549701691 -1
10.9897119998932 -1
11.1896281242371 -1
11.3897271156311 -1
11.5898740291595 -1
11.7899310588837 -1
11.989953994751 -1
12.1895151138306 -1
12.3895721435547 -1
12.5903789997101 -1
12.7895510196686 -1
12.9899210929871 -1
13.1894071102142 -1
13.3897559642792 -1
13.5897209644318 -1
13.7903289794922 -1
13.9895601272583 -1
14.1894900798798 -1
14.3896450996399 -1
14.5897760391235 -1
14.7893340587616 -1
14.9900829792023 -1
15.1896030902863 -1
15.3894050121307 -1
15.5897531509399 -1
15.7895770072937 -1
15.9897220134735 -1
16.1898410320282 -1
16.3898661136627 -1
16.5893559455872 -1
16.7897281646729 -1
16.9898860454559 -1
17.1893730163574 -1
17.3894481658936 -1
17.5917479991913 -1
17.7894501686096 -1
17.9896790981293 -1
18.1894769668579 -1
18.3895220756531 -1
18.5894751548767 -1
18.7896430492401 -1
18.9896709918976 -1
19.1893899440765 -1
19.3945829868317 -1
19.5897271633148 -1
19.7895390987396 -1
19.9894881248474 -1
20.1899421215057 -1
20.3895571231842 -1
20.589292049408 -1
20.7899129390717 -1
20.9898359775543 -1
21.1893630027771 -1
21.3897490501404 -1
21.5898411273956 -1
21.7897641658783 -1
21.9896681308746 -1
22.1897950172424 -1
22.3907129764557 -1
22.5918719768524 -1
22.789510011673 -1
22.9894111156464 -1
23.1894149780273 -1
23.3895010948181 -1
23.5895221233368 -1
23.7898471355438 -1
23.9895370006561 -1
24.1895370483398 -1
24.3895659446716 -1
24.589378118515 -1
24.7898800373077 -1
24.9894630908966 -1
25.1895730495453 -1
25.3897390365601 -1
25.5892729759216 -1
25.7900371551514 -1
25.9897830486298 -1
26.1897661685944 -1
26.3893620967865 -1
26.5899369716644 -1
26.7898459434509 -1
26.9898040294647 -1
27.189572095871 -1
27.3895561695099 -1
27.592255115509 -1
27.7896909713745 -1
27.9895551204681 -1
28.1892809867859 -1
28.3900220394135 -1
28.5897991657257 -1
28.7897591590881 -1
28.9899020195007 -1
29.1894190311432 -1
29.3895349502563 -1
29.5903489589691 -1
29.7896800041199 -1
29.9899830818176 -1
30.190052986145 -1
30.3899500370026 -1
30.5896670818329 -1
30.7896289825439 -1
30.9895761013031 -1
31.1897721290588 -1
31.3905169963837 -1
31.5900211334229 -1
31.7900071144104 -1
31.9896631240845 -1
32.1897051334381 -1
32.3900799751282 -1
32.5896680355072 -1
32.7897500991821 -1
32.9901621341705 -1
33.1896431446075 -1
33.3900020122528 -1
33.5897941589355 -1
33.7895359992981 -1
33.9896960258484 -1
34.189346075058 -1
34.3899109363556 -1
34.589604139328 -1
34.7895660400391 -1
34.9894981384277 -1
35.1898109912872 -1
35.389641046524 -1
35.5895550251007 -1
35.7899060249329 -1
35.9894051551819 -1
36.1894781589508 -1
36.3895659446716 -1
36.5899331569672 -1
36.7896990776062 -1
36.9897530078888 -1
37.1901299953461 -1
37.3895380496979 -1
37.5957500934601 -1
37.7898480892181 -1
37.9894831180573 -1
38.1896131038666 -1
38.3897709846497 -1
38.5896589756012 -1
38.789510011673 -1
38.9893269538879 1
39.1896669864655 1
39.3901081085205 1
39.5900030136108 1
39.7898271083832 1
39.9897880554199 1
40.189512014389 1
40.3894791603088 1
40.5900089740753 1
40.7903380393982 1
40.9896650314331 1
41.1895890235901 1
41.3895030021667 1
41.5895130634308 1
41.7892999649048 1
41.9898099899292 1
42.1894309520721 1
42.3897609710693 1
42.5898430347443 1
42.7898631095886 1
42.9895730018616 1
43.1894550323486 1
43.3902249336243 1
43.5894131660461 1
43.7896270751953 1
43.989413022995 1
44.1897490024567 1
44.3896470069885 1
44.5894379615784 1
44.7899799346924 1
44.9901490211487 1
45.1897840499878 1
45.3894000053406 1
45.5896379947662 1
45.7900400161743 1
45.9901609420776 1
46.1898159980774 1
46.3898320198059 1
46.5894939899445 1
46.7898149490356 1
46.9898190498352 1
47.1929450035095 1
47.3896479606628 1
47.5904810428619 1
47.7898650169373 1
47.989844083786 1
48.1902780532837 1
48.3896460533142 1
48.5896730422974 1
48.7906489372253 1
48.9938671588898 1
49.189975976944 1
49.3902690410614 1
49.5898251533508 1
49.790118932724 1
49.9905371665955 1
50.1904449462891 1
50.3901340961456 -1
50.5903379917145 -1
50.7981860637665 -1
50.9902210235596 -1
51.1897671222687 -1
51.3905899524689 -1
51.590451002121 -1
51.7902100086212 -1
51.9896969795227 -1
52.1896281242371 -1
52.3904480934143 -1
52.5896921157837 -1
52.7897300720215 -1
52.989511013031 -1
53.1895899772644 -1
53.3900730609894 -1
53.5894689559937 -1
53.7896931171417 -1
53.9898951053619 -1
54.1895439624786 -1
54.3901381492615 -1
54.5899300575256 -1
54.7895569801331 -1
54.9895720481873 -1
55.1895451545715 -1
55.3895461559296 -1
55.5896911621094 -1
55.7899329662323 -1
55.9897999763489 -1
56.1895649433136 -1
56.3902080059052 -1
56.5897870063782 -1
56.7898211479187 -1
56.98948097229 -1
57.1894991397858 -1
57.3896291255951 -1
57.5895881652832 -1
57.7895729541779 -1
57.9895370006561 -1
58.1898050308228 -1
58.3897120952606 -1
58.5898630619049 -1
58.7898809909821 -1
58.9897601604462 -1
59.1896779537201 -1
59.3901431560516 -1
59.5898990631104 -1
59.7893509864807 -1
59.9896631240845 -1
60.1895251274109 -1
60.3911089897156 -1
60.5897009372711 -1
60.7897140979767 -1
60.9894161224365 -1
61.1894171237946 -1
61.3896520137787 -1
61.5897619724274 -1
61.7892200946808 -1
61.989669084549 -1
62.1893670558929 -1
62.3895101547241 -1
62.5897350311279 -1
62.7894470691681 -1
62.9893519878387 -1
63.1896200180054 -1
63.3894600868225 -1
63.5897121429443 -1
63.7894849777222 -1
63.98952293396 -1
64.1896080970764 -1
64.3894519805908 -1
64.5905611515045 -1
64.7896549701691 -1
64.9896650314331 -1
65.1896989345551 -1
65.3899159431458 -1
65.5897691249847 -1
65.7894780635834 -1
65.98992395401 -1
66.1895599365234 -1
66.3903839588165 -1
66.589457988739 -1
66.7894749641418 -1
66.9897770881653 -1
67.1898131370544 -1
67.3897271156311 -1
67.5893220901489 -1
67.790118932724 -1
67.9911341667175 -1
68.1898291110992 -1
68.3894829750061 -1
68.5896410942078 -1
68.7900531291962 -1
68.9898509979248 -1
69.1902711391449 -1
69.3895139694214 -1
69.5900640487671 -1
69.7896461486816 -1
69.9951281547546 -1
70.189651966095 -1
70.3899819850922 -1
70.5893750190735 -1
70.7898750305176 -1
70.9900839328766 -1
71.1897060871124 -1
71.3897550106049 -1
71.5897779464722 -1
71.7896809577942 -1
71.9951951503754 -1
72.1898920536041 -1
72.3895270824432 -1
72.589940071106 -1
72.7906260490417 -1
72.990079164505 -1
73.190299987793 -1
73.3906810283661 -1
73.5904591083527 -1
73.7897479534149 -1
73.9895730018616 -1
74.1895370483398 -1
74.3898150920868 -1
74.5896019935608 -1
74.7899730205536 -1
74.9897651672363 -1
75.1902101039886 -1
75.3897800445557 -1
75.5895600318909 -1
75.7896130084991 -1
75.9895861148834 -1
76.1893579959869 -1
76.3898241519928 -1
76.5895719528198 -1
76.7893569469452 1
76.9894740581512 1
77.1897370815277 1
77.3902080059052 1
77.5898859500885 1
77.790050983429 1
77.9894421100616 1
78.189432144165 1
78.3897540569305 1
78.5896601676941 1
78.7894690036774 1
78.990040063858 1
79.1898419857025 -1
79.3895139694214 -1
79.5899930000305 -1
79.7896101474762 -1
79.9895491600037 -1
80.1899299621582 -1
80.3899121284485 -1
80.5896079540253 -1
80.789754152298 -1
80.9898600578308 -1
81.1898670196533 -1
81.3900849819183 -1
81.5897600650787 -1
81.7896840572357 -1
81.989856004715 -1
82.1899189949036 -1
82.389680147171 -1
82.5895259380341 -1
82.7896540164948 -1
82.98943400383 -1
83.189759016037 -1
83.3898539543152 -1
83.5897991657257 -1
83.7897109985352 -1
83.9897060394287 -1
84.1898140907288 -1
84.3895690441132 -1
84.5896320343018 -1
84.7897040843964 -1
84.9902729988098 -1
85.1895990371704 -1
85.389662027359 -1
85.5898520946503 -1
85.7893280982971 -1
85.9896111488342 -1
86.1898000240326 -1
86.3896069526672 -1
86.5896110534668 -1
86.7897801399231 -1
86.9896860122681 -1
87.189483165741 -1
87.3895480632782 -1
87.5900571346283 -1
87.7894020080566 -1
87.9900741577148 -1
88.1901919841766 -1
88.3900380134583 -1
88.5899729728699 -1
88.7900121212006 -1
88.9901871681213 -1
89.1898920536041 -1
89.3899660110474 -1
89.5909540653229 -1
89.7898609638214 -1
89.9896860122681 -1
90.1897990703583 -1
90.3903820514679 -1
90.5907330513 -1
90.7905969619751 -1
90.9904050827026 -1
91.1903340816498 -1
91.3907589912415 -1
91.5901279449463 -1
91.7899589538574 -1
91.9898200035095 -1
92.1902220249176 -1
92.3902080059052 -1
};
\addlegendentry{MOCAP}
\end{axis}

\end{tikzpicture}

%% file: tex/line_nodes_pose2.tex
\begin{tikzpicture}

\definecolor{color0}{rgb}{1,0.647058823529412,0}



\begin{axis}[
height=\figureheight,
width=\figurewidth,
axis background/.style={fill=white!92!black},
axis line style={white},
legend cell align={left},
legend style={fill opacity=0.8, draw opacity=1, text opacity=1, draw=white!80!black},
tick align=outside,
tick pos=both,
x grid style={white},
xmajorgrids,
xmin=-2.55986396789551, xmax=6.45714332580566,
xtick style={color=black},
y grid style={white},
ymajorgrids,
ymin=-0.2975, ymax=6.2475,
ytick style={color=black}
]

\addplot [semithick, color0]
table {%
4.40268182754517 0.986898362636566
4.37718868255615 1.00204133987427
4.3503098487854 1.01843881607056
4.32239866256714 1.03605592250824
4.29403734207153 1.05497562885284
4.2653865814209 1.07530617713928
4.23669481277466 1.09682559967041
4.20790672302246 1.1191178560257
4.17946577072144 1.14236390590668
4.15121269226074 1.16688311100006
4.12203216552734 1.19253385066986
4.09261608123779 1.21896243095398
4.06327247619629 1.2452632188797
4.03413009643555 1.27206444740295
4.00511121749878 1.29894769191742
3.97613859176636 1.3258193731308
3.94705748558044 1.35238683223724
3.91769218444824 1.37819087505341
3.88855886459351 1.40380108356476
3.8593955039978 1.42950987815857
3.83040547370911 1.45510792732239
3.80150127410889 1.48029637336731
3.77255201339722 1.50384747982025
3.74366092681885 1.52567183971405
3.71446132659912 1.54716658592224
3.68520951271057 1.56855309009552
3.65614581108093 1.58985579013824
3.62630105018616 1.61039710044861
3.59604454040527 1.63080132007599
3.56558227539062 1.65200173854828
3.5349018573761 1.67381811141968
3.5037693977356 1.69652569293976
3.47229409217834 1.71944570541382
3.44041275978088 1.7420836687088
3.4081757068634 1.76521468162537
3.37594819068909 1.78841757774353
3.34345889091492 1.81170237064362
3.31018376350403 1.83449327945709
3.27601194381714 1.8565559387207
3.24104285240173 1.87832140922546
3.20589089393616 1.90018904209137
3.17120742797852 1.92251563072205
3.13686156272888 1.94472873210907
3.10301113128662 1.96685969829559
3.06943941116333 1.98943650722504
3.03647303581238 2.01271915435791
3.00443243980408 2.03622055053711
2.9735004901886 2.05958437919617
2.94316744804382 2.08183693885803
2.91318106651306 2.1037278175354
2.88576126098633 2.1230936050415
2.85590958595276 2.14399766921997
2.82392382621765 2.16724252700806
2.79513001441956 2.18884491920471
2.76633238792419 2.21186518669128
2.73765110969543 2.2350332736969
2.70868396759033 2.2590696811676
2.67965841293335 2.28270244598389
2.65106654167175 2.30640578269958
2.6229875087738 2.33003735542297
2.59557461738586 2.35282969474792
2.56818842887878 2.37531161308289
2.54075884819031 2.39727687835693
2.51244044303894 2.41896748542786
2.48280668258667 2.44075322151184
2.45245361328125 2.46346950531006
2.42141318321228 2.48717951774597
2.39054727554321 2.51114702224731
2.36081099510193 2.53438067436218
2.33155941963196 2.55703949928284
2.30255913734436 2.57940268516541
2.27358889579773 2.60214281082153
2.24479794502258 2.62497854232788
2.21659636497498 2.64717578887939
2.18845462799072 2.66864514350891
2.1591796875 2.68936395645142
2.13058304786682 2.7105438709259
2.10114741325378 2.73174571990967
2.07148885726929 2.75335288047791
2.04269218444824 2.77582621574402
2.01376271247864 2.79885387420654
1.98547279834747 2.821448802948
1.95838737487793 2.84367346763611
1.93244159221649 2.86436414718628
1.9073885679245 2.88418459892273
1.88319253921509 2.90414619445801
1.85854256153107 2.92402243614197
1.83324205875397 2.94392919540405
1.80647552013397 2.96406674385071
1.78133106231689 2.98301291465759
1.75090730190277 3.00581955909729
1.72255349159241 3.02755093574524
1.69390654563904 3.04995656013489
1.6653459072113 3.07303762435913
1.63589191436768 3.09681582450867
1.60545098781586 3.12119364738464
1.57466840744019 3.14559006690979
1.54346418380737 3.17015790939331
1.51202750205994 3.19402432441711
1.48013579845428 3.21746492385864
1.44942235946655 3.23980188369751
1.4189121723175 3.26101136207581
1.38951981067657 3.28150296211243
1.36068439483643 3.30298924446106
1.3325047492981 3.32629561424255
1.30488717556 3.35142874717712
1.27794885635376 3.37757563591003
1.25111937522888 3.40381169319153
1.22424626350403 3.42945289611816
1.19723296165466 3.45466446876526
1.17107892036438 3.47866225242615
1.15274381637573 3.49681973457336
1.14919579029083 3.50429511070251
1.15759515762329 3.50115823745728
1.17440867424011 3.4904580116272
1.19552731513977 3.475905418396
1.21941745281219 3.46104168891907
1.24508416652679 3.44626331329346
1.27125871181488 3.43117833137512
1.29767549037933 3.41576910018921
1.32505488395691 3.39938116073608
1.35312640666962 3.38245368003845
1.38170480728149 3.36586236953735
1.41107106208801 3.34948134422302
1.44061708450317 3.33266258239746
1.47060334682465 3.31598091125488
1.50085318088531 3.29919695854187
1.53125536441803 3.28255653381348
1.56209278106689 3.26524043083191
1.59345805644989 3.24795484542847
1.62461733818054 3.22942590713501
1.65626847743988 3.20973920822144
1.68750047683716 3.18858242034912
1.71754670143127 3.16637754440308
1.74726068973541 3.1439573764801
1.77669978141785 3.12061262130737
1.80592584609985 3.09657645225525
1.83534777164459 3.07227158546448
1.86514210700989 3.04839563369751
1.89458608627319 3.02463960647583
1.92394304275513 3.00089693069458
1.95289051532745 2.97637724876404
1.98216187953949 2.95113778114319
2.01147246360779 2.92566871643066
2.04050898551941 2.90015292167664
2.06924510002136 2.87480783462524
2.09762668609619 2.85006260871887
2.1265287399292 2.82511615753174
2.15518832206726 2.7987744808197
2.18505930900574 2.77176332473755
2.21460437774658 2.74428677558899
2.24303555488586 2.71620273590088
2.27136325836182 2.6865348815918
2.29803490638733 2.65681290626526
2.32343697547913 2.62640619277954
2.34841442108154 2.59512209892273
2.37191414833069 2.5637526512146
2.394207239151 2.53305125236511
2.41683101654053 2.50223112106323
2.43872952461243 2.47208046913147
2.45998573303223 2.44291877746582
2.48106932640076 2.41523766517639
2.50179529190063 2.3877055644989
2.52349424362183 2.36004900932312
2.54635000228882 2.33266425132751
2.56925392150879 2.30615925788879
2.5917694568634 2.28063225746155
2.61509013175964 2.25548481941223
2.64160108566284 2.23071503639221
2.66832876205444 2.20557117462158
2.69691038131714 2.18015050888062
2.72584891319275 2.15488219261169
2.75589656829834 2.13113164901733
2.78750038146973 2.11033868789673
2.81976366043091 2.09133410453796
2.85169672966003 2.0729513168335
2.88317561149597 2.05336427688599
2.91412258148193 2.03364872932434
2.94459581375122 2.0131528377533
2.97585463523865 1.99245870113373
3.00693202018738 1.97182047367096
3.03836297988892 1.9518837928772
3.06949782371521 1.93295931816101
3.09965109825134 1.9162757396698
3.12900519371033 1.90081357955933
3.1582715511322 1.88599121570587
3.18788051605225 1.87162268161774
3.21783351898193 1.8571150302887
3.24794840812683 1.84302639961243
3.27872180938721 1.82801532745361
3.30975866317749 1.81187498569489
3.34059691429138 1.79416155815125
3.37142038345337 1.77500081062317
3.40125894546509 1.75466310977936
3.43030738830566 1.73359620571136
3.45815205574036 1.71161484718323
3.48527431488037 1.68978142738342
3.51151752471924 1.66813778877258
3.53709864616394 1.64698100090027
3.56275320053101 1.6258111000061
3.58831214904785 1.60441243648529
3.61450409889221 1.58268570899963
3.64064073562622 1.56068825721741
3.66722393035889 1.53901743888855
3.69424176216125 1.51709794998169
3.72168779373169 1.49510025978088
3.74917268753052 1.47242701053619
3.77647113800049 1.44908201694489
3.80365014076233 1.42537248134613
3.83110308647156 1.40186882019043
3.8584508895874 1.37822985649109
3.88592433929443 1.35429978370667
3.91345238685608 1.32986664772034
3.94114542007446 1.30503022670746
3.96883869171143 1.28017628192902
3.99609279632568 1.2559107542038
4.02343893051147 1.23314929008484
4.05081605911255 1.21148586273193
4.07815647125244 1.19119954109192
4.10546016693115 1.17226815223694
4.13397026062012 1.15503168106079
4.16349935531616 1.1388373374939
4.19379758834839 1.1232351064682
4.22444009780884 1.10774600505829
4.25523900985718 1.091268658638
4.28492069244385 1.07376670837402
4.31411027908325 1.0554621219635
4.34266901016235 1.03604662418365
4.37073564529419 1.01496279239655
4.39835119247437 0.993474662303925
4.42510890960693 0.97095662355423
4.44992733001709 0.948507487773895
};
\addlegendentry{Mocap}
\addplot [semithick, red]
table {%
4.39233347527004 0.549658727520683
4.39625520145949 0.502829621153845
4.36380885386803 0.582072475791939
4.3424619931537 0.575444951064336
4.3014084948844 0.689688722552006
4.25133465971475 0.757592101917157
4.21184402412732 0.82708812499037
4.17519697965255 0.866297403712083
4.13691296731535 0.946878688239124
4.10615626357549 0.962724519640541
4.09672847605564 0.977243062373474
4.07495672468232 0.955427602578806
4.06140700785809 0.951886556519841
4.02596385338102 0.98609850505768
3.99664797814341 1.00186117837406
3.97916632922732 1.00610518353038
3.97244916408869 1.03511024173429
3.92297214638003 1.07332798906699
3.90243001489937 1.07963951251896
3.88054591072828 1.08968737674313
3.83900761884501 1.14039425725661
3.76202108745538 1.20700941569399
3.73624635427602 1.26916794567196
3.6894450423124 1.34146804120029
3.65063838580067 1.38515038320203
3.61191164231343 1.41631035131457
3.58792290880397 1.43923002508514
3.56665107048774 1.44382054254891
3.55651823561361 1.4421363440762
3.52718063961389 1.47049388056755
3.50084172565594 1.5120377266179
3.46573948198195 1.55078541535282
3.42393302802422 1.58049851353498
3.38373593270409 1.60642610666905
3.32511353745506 1.63902724608978
3.30460007351176 1.64034078409786
3.27518729177041 1.64953224146932
3.24808972647585 1.68423400524458
3.21468140268221 1.70558032804212
3.219180310149 1.714004306661
3.18001773400562 1.75422964579053
3.14773027971929 1.78278125764548
3.11577071574046 1.80537543364479
3.06849020012118 1.84189856085192
3.01294926199798 1.8827013477134
2.95963195937769 1.91873235248746
2.93083553614722 1.94859069647522
2.88324241986167 1.97277602181994
2.84972245792374 1.99674747266896
2.81737970992398 2.01455658039037
2.77544278322043 2.0280802736874
2.73083874496983 2.05798157623787
2.69845884550986 2.09004532200187
2.66088479644716 2.12470564833472
2.62914063067968 2.14234095707652
2.59340975817501 2.17921184003078
2.54203729043508 2.20779847083594
2.48998018098932 2.23005361057678
2.4670137604661 2.24980612608956
2.42896645624441 2.28617740797212
2.39492105211446 2.31036482586864
2.36913690673615 2.3361849057963
2.34681809775885 2.35273011087339
2.30889125062422 2.36839046911433
2.28291472684031 2.38825078520517
2.25347022136434 2.40496458238483
2.2249221662261 2.41734402443161
2.20552521094182 2.44042035267879
2.18270296821881 2.4770617674845
2.1367063261188 2.50607238545429
2.10713919813246 2.53609081180949
2.08359699687357 2.56315909911697
2.04504430137306 2.59424163414729
2.00444838120453 2.61601222533402
1.98091666477581 2.64851256529873
1.93977124433697 2.67835805910673
1.89204483344813 2.71687473849862
1.83303030292035 2.75150500520084
1.7968618959541 2.78567008250684
1.75182490288336 2.81463391167304
1.71047814228695 2.84011088641627
1.67024580775136 2.85925874817933
1.66007404987469 2.87438978923136
1.63672843554835 2.88371871399298
1.62423007992044 2.9000343306568
1.62329349755474 2.91886730500393
1.62876621931134 2.934105624911
1.61683128099589 2.96325260965817
1.61122385220595 2.98926832010186
1.60935516683023 2.9993938428272
1.59638856734939 3.0220344037492
1.58278796241063 3.04937186590373
1.56622797876233 3.06871466327145
1.52268305814407 3.09246680003189
1.48596928984878 3.13499073549927
1.4668118465371 3.157533752361
1.49636737649294 3.1896390272288
1.46873073654563 3.20199456926807
1.43860253254855 3.22328851918488
1.40345003285367 3.22926283650211
1.3798753018955 3.25466796485292
1.25665995891868 3.26082641963644
1.22977807472865 3.29279765591872
1.20379083571806 3.3255239839667
1.19517003840827 3.34584812655055
1.11405843540326 3.35034399520876
1.10887890475396 3.37505255940749
1.05021229732213 3.39962901937704
1.02329380651629 3.42461016102891
0.960901186963254 3.45854464733202
0.910768291105166 3.47470859718706
0.898064531620833 3.49510068429471
0.902882702937215 3.51864976875187
0.89694359442793 3.52444108483674
0.89694359442793 3.52444108483674
0.938483416768002 3.53337539991776
0.920637142787109 3.5298679531223
0.933530978776819 3.52220817888737
0.96046508973428 3.52900773580438
0.997699971541131 3.52256353060964
1.03765346637449 3.52359408068216
1.03831265707385 3.50068982994068
1.06251306506575 3.48078044699904
1.10813254117965 3.44163011826999
1.14050839261856 3.41787999269842
1.17169241613393 3.41152300771292
1.23038692775596 3.40592363373628
1.27399300399699 3.39875218022672
1.27458903029194 3.38561315806937
1.30603984180679 3.37307207981334
1.33661581662312 3.34381529108074
1.37533490970645 3.33964680554712
1.39330458966492 3.31052745653187
1.44362938087784 3.28590598103242
1.47824903365564 3.26336515248994
1.53739230966769 3.23943191065486
1.57220346612653 3.20390626514711
1.59237692611306 3.18804363510721
1.58964634428954 3.17152970692219
1.59336337524651 3.15171500685016
1.57407678134021 3.13249755070243
1.60097268049167 3.11395308816554
1.62137875559534 3.090181355882
1.62826204444826 3.07314017538264
1.63868294307915 3.05853548448124
1.65418320852185 3.03875699144758
1.65101547377412 3.0153191730854
1.68525158523382 2.98646509867231
1.72556205054995 2.95162844669684
1.76813776259555 2.91916641820214
1.79182078847392 2.88627365492635
1.83234428997358 2.8468788659167
1.85359469748766 2.81631746011932
1.89754859948315 2.79607609763193
1.94258860263282 2.74877589902158
2.00488210069838 2.7139926688362
2.03469688983605 2.69243514056688
2.08076400572705 2.65772829681394
2.11667906614591 2.61900646237825
2.12929793346135 2.59295185029411
2.16035499312253 2.55026880622407
2.19892186180079 2.50759695845
2.21982121318226 2.46801499545141
2.24383924251215 2.42972758258372
2.28471373827366 2.39699186796304
2.31102247637359 2.37102713353201
2.31354691807443 2.34393726629343
2.33965580534639 2.3123746366415
2.35560882503605 2.27570118237288
2.37662092533957 2.24744670278698
2.39007829485847 2.22095118633927
2.45338322194218 2.1973147803793
2.48240183672155 2.1769706665012
2.51945092844864 2.16788025165948
2.54703881541993 2.15399027189458
2.56802775567835 2.12885854121507
2.57711084536079 2.10528420477441
2.58687932669736 2.0807445165425
2.60271491637234 2.04085719340641
2.61670278120144 2.0097500116007
2.66995139980755 1.99502403647495
2.70482175656812 1.98553360874173
2.76794258618923 1.96308296227863
2.80358539670312 1.95753620759801
2.85468907904698 1.93594276364207
2.89986593043532 1.91018882615986
2.97066790605163 1.85508477985663
2.99685580000452 1.85139453678245
3.04265476201435 1.8233783898902
3.07836320185872 1.80674520942658
3.09846242576543 1.7752351135609
3.09728457453731 1.77733855648127
3.13114427276675 1.74733375325856
3.1705622350164 1.73805397113963
3.20348079361869 1.71931273669757
3.24861386079203 1.70520838975411
3.30768841477663 1.66373036854418
3.34077840068819 1.62479943652156
3.36142921958462 1.59691632072185
3.39908016901711 1.55934084846708
3.40981502629234 1.53568536285611
3.43005933338189 1.51920219458333
3.47128418688288 1.48893601167816
3.50024895175442 1.44153100358371
3.52079426665269 1.42320567251588
3.55398362798993 1.40555643162449
3.58886611725127 1.36416027527706
3.5952346793807 1.35881098076469
3.60829633870728 1.35823250043242
3.61911177627153 1.32904590507956
3.6427210287303 1.29274167173075
3.635145443653 1.30231100763623
3.68263583903179 1.24138909222019
3.73668534602282 1.21350813627231
3.78213821377603 1.16761586881753
3.81881273789536 1.13573445529813
3.86591852651773 1.10053539029487
3.88045475070639 1.11515291734872
3.89681792811152 1.08010534218957
3.92082145952156 1.09094494589905
3.93112348535987 1.09874512133877
3.9725202985652 1.04907630274763
4.00859888202568 1.014457476765
4.0421530280328 0.986122082495998
4.06308639566768 1.00764712457641
4.10323223165357 0.950801099503094
4.13854202904819 0.901870356706541
4.17698766769206 0.827382767732813
4.21054857858108 0.813085463627334
4.25844649389339 0.720513932116259
4.29024684072501 0.713765834386634
4.30080537049255 0.736253637643834
};
\addlegendentry{TOF}
\addplot [semithick, blue]
table {%
4.5 0.8
4.45 0.8
4.45 0.75
4.4 0.85
4.35 0.85
4.35 0.85
4.25 0.95
4.2 0.95
4.2 1
4.1 1.1
4.1 1.05
4.1 1.05
4.05 1.1
4.05 1.1
3.95 1.2
4 1.15
4 1.15
4 1.5
4 1.25
4 1.3
3.7 1.4
3.8 1.4
3.9 1.4
3.9 1.4
3.8 1.45
3.8 1.45
3.65 1.55
3.65 1.5
3.6 1.55
3.5 1.6
3.45 1.6
3.4 1.65
3.35 1.7
3.3 1.7
3.25 1.7
3.15 1.75
3.05 1.8
3.05 1.8
2.95 1.85
2.95 1.85
2.85 1.9
2.95 1.9
3 1.9
3 1.95
2.95 1.95
2.9 2
2.9 2
2.9 2
2.75 2.05
2.75 2.05
2.75 2.05
2.7 2.1
2.6 2.15
2.55 2.15
2.5 2.2
2.5 2.2
2.5 2.25
2.45 2.25
2.4 2.25
2.35 2.3
2.3 2.35
2.25 2.35
2.3 2.35
2.25 2.35
2.2 2.4
2.15 2.4
2.15 2.45
2.1 2.5
2.2 2.5
1.65 2.85
1.5 2.6
2.05 2.5
1.65 2.7
1.3 2.8
1.35 2.8
1.65 2.75
1.2 2.9
1.2 2.9
1.1 3
1.2 2.9
1.85 2.75
1.8 2.75
1.75 2.85
1.5 2.95
1.65 2.95
1.3 3.05
1.45 2.85
1.45 2.95
1 2.85
1 2.85
-0.65 4.25
-1.95 5.35
0.8 3.2
0.2 3.65
1.15 3.25
0.5 3.75
-0.6 4.25
1.15 3.3
1.25 3.25
0.75 3.2
0.6 3.7
1.35 3.15
1.25 3.2
1.3 3.15
1.3 3.15
1.25 3.2
1.15 3.3
1.05 3.35
1.1 3.35
1.1 3.3
0.9 3.45
0.95 3.4
0.85 3.5
0.8 3.55
1 3.4
1 3.4
1 3.4
1.1 3.35
0.15 3.75
-1.95 5.95
-1.3 4.8
-2.15 4.95
-0.7 4.5
0.8 3.55
1 3.25
0.9 3.4
0.95 3.35
1.45 3.1
1.05 3.4
1.2 3.3
1.4 3.15
1.55 3.1
1.25 2.95
1.35 2.9
-0.45 4.5
1.65 3
-1.35 4.5
1.45 2.7
1.45 2.85
-0.3 3.9
1.65 2.95
1.2 3.2
1.9 2.85
1.9 2.9
1.95 2.85
1.7 2.85
1.9 2.8
1.85 2.8
1.55 2.65
1.2 3.4
1.2 3.4
1.75 2.65
1.4 2.85
2.15 2.65
1.85 2.7
2.2 2.55
2.2 2.55
2.25 2.5
2.2 2.5
2.2 2.45
2.15 2.45
2.15 2.4
2.15 2.4
2.25 2.35
2.25 2.35
2.25 2.3
2.3 2.25
2.3 2.25
2.4 2.2
2.4 2.2
2.45 2.15
2.4 2.15
2.45 2.15
2.5 2.15
2.5 2.1
2.55 2.1
2.65 2.1
2.75 1.85
2.8 2
2.8 2
2.95 1.95
2.95 2
2.9 1.95
2.8 1.95
2.8 1.95
2.75 1.9
2.75 1.9
2.8 1.9
2.85 1.85
2.85 1.85
2.95 1.85
3.05 1.8
3.05 1.8
3.2 1.75
3.25 1.75
3.25 1.75
3.3 1.7
3.4 1.7
3.35 1.6
3.5 1.6
3.15 1.7
3.2 1.6
3.55 1.55
3.7 1.55
3.75 1.5
3.8 1.45
3.75 1.45
3.65 1.5
3.3 1.5
3.7 1.35
3.95 1.3
3.95 1.25
4 1.25
3.95 1.25
4.15 1.3
3.9 1.25
3.9 1.2
4 1.15
4.05 1.1
4.1 1.05
4.1 1.05
4.1 1.05
4.1 1.05
4.15 1
4.2 1
4.25 1
4.3 0.9
4.35 0.95
4.35 0.9
4.45 0.85
4.5 0.9
4.55 0.8
};
\addlegendentry{TDOA}
\addplot [semithick, green!50.1960784313725!black]
table {%
4.39233347527004 0.549658727520683
4.39625520145949 0.502829621153845
4.36380885386803 0.582072475791939
4.3424619931537 0.575444951064336
4.3014084948844 0.689688722552006
4.25133465971475 0.757592101917157
4.25 0.95
4.2 0.95
4.2 1
4.1 1.1
4.1 1.05
4.05 1.1
4.05 1.1
4.05 1.15
3.95 1.2
4 1.15
4 1.15
4 1.5
3.95 1.3
4 1.3
3.7 1.45
3.8 1.4
3.9 1.4
3.9 1.4
3.8 1.45
3.8 1.45
3.7 1.55
3.65 1.5
3.6 1.55
3.5 1.65
3.45 1.6
3.45 1.65
3.35 1.7
3.3 1.7
3.25 1.7
3.15 1.75
3.05 1.8
3.05 1.8
2.95 1.85
2.95 1.85
2.85 1.9
2.95 1.9
3 1.9
3 1.95
2.95 1.95
2.9 2
2.9 2
2.8 2.05
2.75 2.05
2.8 2.05
2.8 2.05
2.7 2.1
2.6 2.15
2.55 2.15
2.5 2.2
2.5 2.2
2.5 2.25
2.45 2.25
2.4 2.25
2.35 2.3
2.3 2.35
2.25 2.35
2.3 2.35
2.25 2.35
2.2 2.4
2.15 2.4
2.15 2.45
2.1 2.5
2.2 2.5
1.7 2.9
1.5 2.6
2.05 2.5
1.65 2.7
1.3 2.8
1.35 2.8
1.65 2.75
1.2 2.9
1.15 2.9
1.1 3
1.25 2.85
1.85 2.75
1.8 2.75
1.75 2.85
1.5 2.95
1.65 2.95
1.25 3.05
1.45 2.9
1.61683128099589 2.96325260965817
0.95 2.9
0.95 2.9
1.59638856734939 3.0220344037492
-1.9 5.3
0.8 3.2
1.52268305814407 3.09246680003189
1.15 3.25
0.55 3.7
1.49636737649294 3.1896390272288
1.46873073654563 3.20199456926807
1.25 3.25
0.5 3.2
1.3798753018955 3.25466796485292
1.35 3.15
1.25 3.2
1.3 3.15
1.19517003840827 3.34584812655055
1.11405843540326 3.35034399520876
1.10887890475396 3.37505255940749
1.05021229732213 3.39962901937704
1.02329380651629 3.42461016102891
0.960901186963254 3.45854464733202
0.910768291105166 3.47470859718706
0.898064531620833 3.49510068429471
0.902882702937215 3.51864976875187
0.89694359442793 3.52444108483674
0.89694359442793 3.52444108483674
0.938483416768002 3.53337539991776
0.920637142787109 3.5298679531223
0.933530978776819 3.52220817888737
0.96046508973428 3.52900773580438
0.997699971541131 3.52256353060964
1.03765346637449 3.52359408068216
1.03831265707385 3.50068982994068
1.06251306506575 3.48078044699904
1.10813254117965 3.44163011826999
1.14050839261856 3.41787999269842
1.17169241613393 3.41152300771292
1.23038692775596 3.40592363373628
1.27399300399699 3.39875218022672
1.27458903029194 3.38561315806937
1.30603984180679 3.37307207981334
1.33661581662312 3.34381529108074
1.37533490970645 3.33964680554712
1.39330458966492 3.31052745653187
1.44362938087784 3.28590598103242
1.47824903365564 3.26336515248994
1.53739230966769 3.23943191065486
-1.35 4.5
1.5 2.65
1.58964634428954 3.17152970692219
-0.3 3.9
1.6 3
1.60097268049167 3.11395308816554
1.9 2.85
1.9 2.9
1.95 2.85
1.7 2.85
1.9 2.8
1.85 2.8
1.55 2.65
1.2 3.4
1.3 2.95
1.83234428997358 2.8468788659167
1.4 2.85
2.15 2.65
1.85 2.7
2.2 2.55
2.2 2.55
2.3 2.5
2.2 2.5
2.2 2.45
2.15 2.45
2.15 2.4
2.15 2.4
2.25 2.35
2.3 2.35
2.25 2.3
2.3 2.25
2.3 2.25
2.4 2.2
2.4 2.2
2.45 2.15
2.4 2.15
2.5 2.15
2.5 2.15
2.5 2.15
2.55 2.1
2.65 2.1
2.75 1.85
2.8 2
2.8 2
2.95 1.95
2.9 1.95
2.9 1.95
2.8 1.95
2.8 1.95
2.75 1.9
2.75 1.9
2.8 1.9
2.85 1.85
2.85 1.85
2.95 1.85
3.05 1.8
3.05 1.8
3.2 1.75
3.25 1.75
3.25 1.75
3.3 1.7
3.4 1.7
3.35 1.6
3.5 1.6
3.25 1.75
3.2 1.6
3.5 1.5
3.55 1.45
3.75 1.5
3.75 1.5
3.75 1.45
3.65 1.45
3.3 1.45
3.7 1.35
3.95 1.3
3.95 1.25
4 1.25
3.95 1.25
4.15 1.3
3.9 1.25
3.95 1.2
4 1.15
4.1 1.1
4.1 1.05
4.1 1.05
4.1 1.05
4.1 1.05
4.15 1
4.2 1
4.25 1
4.25 0.95
4.35 0.95
4.35 0.95
4.45 0.85
4.45 1
4.30080537049255 0.736253637643834
};
\addlegendentry{Dynamic}
\addplot [semithick, black, mark=asterisk, mark size=3, mark options={solid}]
table {%
0 0
};
\addlegendentry{UWB Nodes}
\addplot [semithick, black, mark=asterisk, mark size=3, mark options={solid}]
table {%
3.04 0
};
\addplot [semithick, black, mark=asterisk, mark size=3, mark options={solid}]
table {%
6.04727935791016 1.98065876960754
};
\addplot [semithick, black, mark=asterisk, mark size=3, mark options={solid}]
table {%
2.81377625465393 4.1934027671814
};
\addplot [semithick, black, mark=asterisk, mark size=3, mark options={solid}]
table {%
3.39600872993469 2.20860457420349
};
\end{axis}

\end{tikzpicture}

%% file: tex/line_boxplot.tex
\begin{tikzpicture}


\definecolor{color0}{rgb}{1,0.752941176470588,0.796078431372549}
\definecolor{color1}{rgb}{0.67843137254902,0.847058823529412,0.901960784313726}
\definecolor{color2}{rgb}{0.564705882352941,0.933333333333333,0.564705882352941}

\begin{axis}[
    height=\figureheight,
    width=\figurewidth,
    axis line style={white},
    tick align=outside,
    tick pos=left,
    x grid style={white!69.0196078431373!black},
    xtick style={color=black},
    y grid style={white!90!black},
    ymajorgrids,
    ytick style={color=black},
    scaled y ticks = false,
    tick align=outside,
    tick pos=left,
    %
    %
    %
    %
    %
    %
xmin=0.5, xmax=3.5,
xtick={1,2,3},
xticklabels={ToF,TDoA,Dynamic},
ylabel={Error (m)},
ymin=-0.0810928646347238, ymax=1.7029501573292,
ytick={-0.2,0,0.2,0.4,0.6,0.8,1,1.2,1.4,1.6,1.8},
yticklabels={0.0,0.0,0.2,0.4,0.6,0.8,1.0,1.2,1.4,1.6,}
]
\addplot [black, forget plot]
table {%
1 0.247243124483298
1 0.191212574373395
};
\addplot [black, forget plot]
table {%
1 0.322990825943308
1 0.386533740853635
};
\addplot [black, forget plot]
table {%
0.875 0.191212574373395
1.125 0.191212574373395
};
\addplot [black, forget plot]
table {%
0.875 0.386533740853635
1.125 0.386533740853635
};
\addplot [mark=0, mark size=1.2, color0, forget plot, fill opacity=0.2, draw opacity=0.666]
table {%
1 0.50843352280211
1 0.449294663151415
1 0.469906520763141
1 0.407502682453082
1 0.427884880781041
1 0.43992003650037
1 0.468516344318961
1 0.462990366930103
1 0.449556162191258
1 0.43511328154
1 0.437995732427785
1 0.427533009299628
1 0.434013175516709
1 0.42129132762959
1 0.401115638432171
};
\addplot [black, forget plot]
table {%
2 0.252352353865499
2 0.110696512656144
};
\addplot [black, forget plot]
table {%
2 0.427369318096555
2 0.6
};
\addplot [black, forget plot]
table {%
1.875 0.110696512656144
2.125 0.110696512656144
};
\addplot [black, forget plot]
table {%
1.875 0.6
2.125 0.6
};
\addplot [mark=0, mark=+, mark size=1.2, color1, only marks, forget plot, fill opacity=0.2, draw opacity=0.666]
table {%
2 0.80452657705633
2 0.856765914952527
2 0.664145830456908
2 0.983030935587097
2 0.903421490150587
2 1.00318554739304
2 0.971342524876511
2 1.05740816414537
2 0.907367220866827
2 0.661559047818523
2 0.863668326248512
2 0.81452148874719
2 0.960952286534252
2 1.62185729269448
2 1.32599849589977
2 0.789878142110486
2 1.02398364365677
2 1.15430291852295
2 0.642629387128155
2 0.77745107543428
2 0.636341881041304
2 1.1476894968596
2 1.16835101453195
2 0.877967507940664
};
\addplot [black, forget plot]
table {%
3 0.24285625227095
3 0.125375455985277
};
\addplot [black, forget plot]
table {%
3 0.361297016940425
3 0.479173711291266
};
\addplot [black, forget plot]
table {%
2.875 0.125375455985277
3.125 0.125375455985277
};
\addplot [black, forget plot]
table {%
2.875 0.479173711291266
3.125 0.479173711291266
};
\addplot [mark=0, mark=+, mark size=1.2, color2, only marks, forget plot, fill opacity=0.2, draw opacity=0.666]
table {%
3 0.515365510818456
3 0.782476325370448
3 0.855228551616133
3 0.663223987369369
3 0.981589332212016
3 0.901620844143095
3 0.579556076079072
3 1.00257649079162
3 1.01843834721538
3 1.05614795130529
3 0.851122176425931
3 0.514613410489002
3 0.707475477917438
3 0.907156627829184
3 0.858868829933425
3 0.6
3 0.959503095996631
3 0.58332541891273
3 1.25736576445386
3 1.03188138939702
3 0.6
3 0.57818405617285
3 0.6
3 0.634002698701113
3 1.14562325442657
3 0.926599166668513
3 0.877488435975554
3 0.493343098018053
3 0.518389168177187
};
\addplot [semithick, color0]
table {%
0 0
};
\addplot [semithick, color1]
table {%
0 0
};
\addplot [semithick, color2]
table {%
0 0
};
\path [draw=black, fill=color0]
(axis cs:0.75,0.247243124483298)
--(axis cs:1.25,0.247243124483298)
--(axis cs:1.25,0.322990825943308)
--(axis cs:0.75,0.322990825943308)
--(axis cs:0.75,0.247243124483298)
--cycle;
\path [draw=black, fill=color1]
(axis cs:1.75,0.252352353865499)
--(axis cs:2.25,0.252352353865499)
--(axis cs:2.25,0.427369318096555)
--(axis cs:1.75,0.427369318096555)
--(axis cs:1.75,0.252352353865499)
--cycle;
\path [draw=black, fill=color2]
(axis cs:2.75,0.24285625227095)
--(axis cs:3.25,0.24285625227095)
--(axis cs:3.25,0.361297016940425)
--(axis cs:2.75,0.361297016940425)
--(axis cs:2.75,0.24285625227095)
--cycle;
\addplot [thick, black, forget plot]
table {%
0.75 0.282986827348523
1.25 0.282986827348523
};
\addplot [thick, black, forget plot]
table {%
1.75 0.305322024202579
2.25 0.305322024202579
};
\addplot [thick, black, forget plot]
table {%
2.75 0.311885981229509
3.25 0.311885981229509
};
\end{axis}

\end{tikzpicture}

%% file: tex/line_Error_distance_NODE_4.tex
\begin{tikzpicture}

\definecolor{color0}{rgb}{1,0.647058823529412,0}

\begin{axis}[
height=\figureheight,
width=\figurewidth,
axis background/.style={fill=white!92!black},
axis line style={white},
legend cell align={left},
legend style={fill opacity=0.8, draw opacity=1, text opacity=1, draw=white!80!black},
tick align=outside,
tick pos=both,
x grid style={white},
xlabel={Time[s]},
xmajorgrids,
xmin=-1.09841945171356, xmax=49.5015223264694,
xtick style={color=black},
y grid style={white},
ylabel={Error[m]},
ymajorgrids,
ymin=0.0351384736542271, ymax=1.69741533169639,
ytick style={color=black},
ytick={0,0.2,0.4,0.6,0.8,1,1.2,1.4,1.6,1.8},
yticklabels={,0.2,0.4,0.6,0.8,1.0,1.2,1.4,1.6,}
]
\addplot [semithick, red]
table {%
1.20157790184021 0.50843352280211
1.40152406692505 0.449294663151415
1.60156798362732 0.469906520763141
1.80152893066406 0.384374539081865
2.00156497955322 0.339614034709799
2.2015368938446 0.297972870907738
2.40154695510864 0.28663099758075
2.60147094726562 0.249178891497994
2.80152297019958 0.249784521290428
3.00154590606689 0.256720400687233
3.2015209197998 0.289716630610892
3.40163493156433 0.311052272890823
3.6015989780426 0.306053104459155
3.80152106285095 0.318888080632073
4.00155997276306 0.333804504293233
4.20151686668396 0.328981451879843
4.40157389640808 0.3213839141622
4.60153102874756 0.338859730752496
4.80152606964111 0.351558021739679
5.00152206420898 0.329818299605487
5.20151901245117 0.304276591188377
5.40156388282776 0.270755357893581
5.60152101516724 0.242483191451026
5.80151700973511 0.231670616112563
6.00154590606689 0.222636829779346
6.20160388946533 0.226634947806841
6.40152096748352 0.23610050480427
6.60151791572571 0.242214994309533
6.80139708518982 0.240532986354211
7.00155186653137 0.22123741246579
7.20159292221069 0.215163124382983
7.40152192115784 0.212416135263492
7.60157704353333 0.212978557430722
7.80156207084656 0.222310997402905
8.00152587890625 0.225336988243311
8.20156407356262 0.231578765973817
8.40155601501465 0.220046403709517
8.60144805908203 0.227891782568457
8.80152106285095 0.220491923727947
9.00152087211609 0.208896098007695
9.20155191421509 0.204537853711451
9.4015679359436 0.206547728093765
9.60158085823059 0.198576495907187
9.80155396461487 0.201478375352641
10.0015599727631 0.204089991683231
10.2015218734741 0.200802826881128
10.4016048908234 0.208738506191721
10.6015219688416 0.215323598204535
10.8015179634094 0.212363170212427
11.0015649795532 0.224008981595376
11.2015550136566 0.252137371731618
11.4015688896179 0.239842591236022
11.6015808582306 0.233323356485584
11.8015599250793 0.23734369318535
12.0015599727631 0.239221779099195
12.2015240192413 0.255409988308377
12.4015610218048 0.270476709993918
12.6014440059662 0.267458798444639
12.8015148639679 0.275527876623811
13.0015108585358 0.283543057693995
13.2015199661255 0.279660730122973
13.4015309810638 0.272713592885924
13.6015150547028 0.283729230281795
13.8015089035034 0.275338088761498
14.0015230178833 0.281382521087248
14.2015240192413 0.282276274938797
14.4015629291534 0.272141064549638
14.6016058921814 0.262542894820052
14.8015539646149 0.277803351568163
15.0016119480133 0.274678332543699
15.201623916626 0.272100233470995
15.4015500545502 0.278859161883495
15.601557970047 0.290162901093108
15.80157995224 0.282986827348523
16.0015950202942 0.294582814944435
16.20157289505 0.313674974180757
16.401526927948 0.341321419298366
16.6015539169312 0.348362962396916
16.8015599250793 0.362790489992567
17.0015399456024 0.373197758998435
17.2015309333801 0.381580402103612
17.4015579223633 0.368084849941523
17.6015238761902 0.36398011862319
17.8015310764313 0.347307761006742
18.0015690326691 0.325865204803854
18.2015619277954 0.30203401991172
18.4015538692474 0.290547818267902
18.6015629768372 0.285259417500351
18.8015320301056 0.256708604803044
19.0015320777893 0.260813174664321
19.2015578746796 0.231316344218089
19.4015419483185 0.229079949694156
19.6015799045563 0.233638297383167
19.8013379573822 0.252591320179805
20.0015239715576 0.239392825180609
20.201514005661 0.204094700227417
20.4015259742737 0.197417168717999
20.6015300750732 0.20020636241849
20.8015298843384 0.197332019058638
21.0015280246735 0.191212574373395
21.2015490531921 0.249637820320775
21.4015259742737 0.250101959511272
21.6015188694 0.247855698304622
21.8015079498291 0.23341998998434
22.001373052597 0.273289345081819
22.2015209197998 0.253718179558422
22.4015738964081 0.279878373369019
22.601569890976 0.278030146301711
22.8020169734955 0.310996396538281
23.0015769004822 0.330454151013113
23.2015368938446 0.312713870472515
23.4015209674835 0.291834175682912
23.6015510559082 0.300146530671704
23.8015608787537 0.305494609406054
24.0015199184418 0.297389753225027
24.2015700340271 0.32171316353446
24.4015760421753 0.334857908296883
24.6015560626984 0.337362828566791
24.8015530109406 0.330442718659038
25.0015568733215 0.322886084375051
25.2015650272369 0.342352439629513
25.4015760421753 0.345195218198434
25.6015269756317 0.33027121318422
25.8015248775482 0.322368123174933
26.0015149116516 0.32265282526402
26.2015118598938 0.300437017865408
26.4015100002289 0.29665321619841
26.6015138626099 0.320636558354171
26.8015129566193 0.323095567511565
27.0015108585358 0.316740370605986
27.2015180587769 0.31552658005698
27.4015109539032 0.323358434002259
27.6015119552612 0.306017057612706
27.8015110492706 0.302875038986007
28.0015199184418 0.285769818827795
28.2015390396118 0.270979853247198
28.401526927948 0.281275872260249
28.6015529632568 0.309372007520399
28.8015120029449 0.332654860560655
29.0015449523926 0.372890353491461
29.201514005661 0.378658024480897
29.401524066925 0.386533740853635
29.6015288829803 0.407502682453082
29.8015248775482 0.427884880781041
30.001522064209 0.43992003650037
30.2015218734741 0.468516344318961
30.4016280174255 0.462990366930103
30.601567029953 0.449556162191258
30.8015389442444 0.43511328154
31.0015308856964 0.437995732427785
31.2015249729156 0.427533009299628
31.4015228748322 0.434013175516709
31.6015400886536 0.42129132762959
31.8015170097351 0.401115638432171
32.0015139579773 0.365241055782094
32.2015199661255 0.365536112891663
32.4015200138092 0.344685026072086
32.6015129089355 0.332614123025595
32.8015170097351 0.342295702713532
33.001522064209 0.330581664923548
33.2014420032501 0.31212775338871
33.4015109539032 0.312150077336333
33.6020269393921 0.306829738532985
33.8015909194946 0.288796468823871
34.0015108585358 0.287180311047657
34.2015109062195 0.303161549348171
34.4015119075775 0.300943279735313
34.6015119552612 0.307598580885184
34.8018360137939 0.308604030257852
35.0015280246735 0.325441957212816
35.201523065567 0.292010621070235
35.4015519618988 0.290746512219162
35.6015539169312 0.286555892953731
35.8015189170837 0.291384568901122
36.0015180110931 0.302442424369143
36.2015190124512 0.320260913941512
36.4015839099884 0.336213351785733
36.601557970047 0.35148438616388
36.8015649318695 0.366136049574582
37.0015618801117 0.343722964263256
37.201523065567 0.340709858304189
37.4015300273895 0.313752277966377
37.6015410423279 0.311119267246518
37.801521062851 0.29443603596169
38.0015208721161 0.278732508987624
38.201523065567 0.248366073500883
38.401556968689 0.248924167691645
38.6015639305115 0.238541533226457
38.8015630245209 0.237025255011328
39.0015249252319 0.24735230006567
39.2015390396118 0.268792935910316
39.4015500545502 0.264672206085977
39.6015570163727 0.266220103932915
39.8015298843384 0.258494246129604
40.0015258789062 0.241280394916159
40.2015330791473 0.220236950190606
40.4015309810638 0.225186090485827
40.601567029953 0.229512322556861
40.8015749454498 0.229307667385965
41.0015370845795 0.237522043481788
41.2015578746796 0.240770950287537
41.4015669822693 0.235174420728419
41.6015300750732 0.246052615092053
41.8015439510345 0.246549250333345
42.0015299320221 0.240083332377672
42.2015490531921 0.246487713486615
42.4015378952026 0.248596927302741
42.6015548706055 0.249902656107265
42.8015260696411 0.260758731061171
43.0015239715576 0.270647841755755
43.2015578746796 0.284980020695233
43.4015519618988 0.281569509097571
43.6015570163727 0.264382297977661
43.801647901535 0.262364413103736
44.0015239715576 0.262001186038889
44.2015180587769 0.252972703620167
44.401517868042 0.247684015221956
44.6015560626984 0.259089559388636
44.8015270233154 0.244874177747387
45.001601934433 0.247133948900926
45.201593875885 0.249060626792136
45.4015209674835 0.254337104358989
45.6015260219574 0.259126318137135
45.8015539646149 0.248603125267332
46.0015530586243 0.262092151588694
46.2015240192413 0.276537042694052
46.4015250205994 0.310801439751345
46.6015388965607 0.303532804180975
46.8015260696411 0.353378506555213
47.0016350746155 0.339099519186388
47.2015249729156 0.308223318657222
};
\addlegendentry{ToF}
\addplot [semithick, blue]
table {%
1.20157790184021 0.277142691091876
1.40152406692505 0.331805668960073
1.60156798362732 0.268032351696871
1.80152893066406 0.274408143454336
2.00156497955322 0.296590678876775
2.2015368938446 0.23958450287851
2.40154695510864 0.255468416546663
2.60147094726562 0.237851954092456
2.80152297019958 0.223612649289392
3.00154590606689 0.243852446516745
3.2015209197998 0.252674295395913
3.40163493156433 0.23792838278772
3.6015989780426 0.252030412335085
3.80152106285095 0.234259464947081
4.00155997276306 0.255936302858387
4.20151686668396 0.268786713572214
4.40157389640808 0.258028350621545
4.60153102874756 0.256393198576848
4.80152606964111 0.257345652894698
5.00152206420898 0.259269057190078
5.20151901245117 0.214325072439429
5.40156388282776 0.240294572596637
5.60152101516724 0.265162036552483
5.80151700973511 0.233192834914657
6.00154590606689 0.235829627453043
6.20160388946533 0.210012544944818
6.40152096748352 0.227683185238047
6.60151791572571 0.215573702816685
6.80139708518982 0.229368055554581
7.00155186653137 0.244102796592381
7.20159292221069 0.250205646512584
7.40152192115784 0.259988161747467
7.60157704353333 0.268035026131951
7.80156207084656 0.282672241517633
8.00152587890625 0.329551641458719
8.20156407356262 0.383711588297715
8.40155601501465 0.355686782716186
8.60144805908203 0.413644670467995
8.80152106285095 0.386797068767113
9.00152087211609 0.443762451935827
9.20155191421509 0.325555796849801
9.4015679359436 0.27216244087842
9.60158085823059 0.245925620342722
9.80155396461487 0.259895918887927
10.0015599727631 0.269564542826182
10.2015218734741 0.245465055732624
10.4016048908234 0.110696512656144
10.6015219688416 0.303915504687323
10.8015179634094 0.283841820327828
11.0015649795532 0.171841680194555
11.2015550136566 0.190730360360517
11.4015688896179 0.336191723526175
11.6015808582306 0.341474871477117
11.8015599250793 0.359124990579707
12.0015599727631 0.338525243575095
12.2015240192413 0.315615538378151
12.4015610218048 0.328194430062503
12.6014440059662 0.348414043177226
12.8015148639679 0.364088544492672
13.0015108585358 0.386503453649564
13.2015199661255 0.400756596616459
13.4015309810638 0.337679117496235
13.6015150547028 0.359450794998807
13.8015089035034 0.371568647414748
14.0015230178833 0.393345176628379
14.2015240192413 0.365717676308531
14.4015629291534 0.381704285009887
14.6016058921814 0.278637119407337
14.8015539646149 0.80452657705633
15.0016119480133 0.856765914952527
15.201623916626 0.335360039149447
15.4015500545502 0.664145830456908
15.601557970047 0.983030935587097
15.80157995224 0.903421490150587
16.0015950202942 0.581023433689451
16.20157289505 1.00318554739304
16.401526927948 0.971342524876511
16.6015539169312 1.05740816414537
16.8015599250793 0.907367220866827
17.0015399456024 0.282804866062596
17.2015309333801 0.297438614693279
17.4015579223633 0.313544513558482
17.6015238761902 0.514866319303976
17.8015310764313 0.359964477003943
18.0015690326691 0.661559047818523
18.2015619277954 0.48219906098153
18.4015538692474 0.454495713038497
18.6015629768372 0.863668326248512
18.8015320301056 0.81452148874719
19.0015320777893 0.6
19.2015578746796 0.6
19.4015419483185 0.960952286534252
19.6015799045563 1.62185729269448
19.8013379573822 0.584327992775906
20.0015239715576 1.32599849589977
20.201514005661 0.6
20.4015259742737 0.494132393454141
20.6015300750732 0.366656021429393
20.8015298843384 0.789878142110486
21.0015280246735 1.02398364365677
21.2015490531921 0.244193528355486
21.4015259742737 0.272784945260055
21.6015188694 0.263639776014092
21.8015079498291 0.260129170549654
22.001373052597 0.255197193789222
22.2015209197998 0.258339830033765
22.4015738964081 0.305322024202579
22.601569890976 0.259522773076831
22.8020169734955 0.269222877860435
23.0015769004822 0.365312001421369
23.2015368938446 0.30842975860486
23.4015209674835 0.358228135508571
23.6015510559082 0.406417504802508
23.8015608787537 0.273443583859668
24.0015199184418 0.28199699491023
24.2015700340271 0.290372925302836
24.4015760421753 0.258670716728359
24.6015560626984 1.15430291852295
24.8015530109406 0.6
25.0015568733215 0.6
25.2015650272369 0.6
25.4015760421753 0.6
25.6015269756317 0.642629387128155
25.8015248775482 0.469313576763339
26.0015149116516 0.582031746187464
26.2015118598938 0.557311776014718
26.4015100002289 0.287789745346243
26.6015138626099 0.534955207423665
26.8015129566193 0.41757270279251
27.0015108585358 0.295708534669786
27.2015180587769 0.24848423004343
27.4015109539032 0.522886496680334
27.6015119552612 0.487662634825443
27.8015110492706 0.6
28.0015199184418 0.266832001497073
28.2015390396118 0.6
28.401526927948 0.571374121642588
28.6015529632568 0.487443590439752
28.8015120029449 0.6
29.0015449523926 0.326567471556739
29.201514005661 0.77745107543428
29.401524066925 0.245586759075692
29.6015288829803 0.229341005437592
29.8015248775482 0.229664292796162
30.001522064209 0.402536165640402
30.2015218734741 0.277118032764142
30.4016280174255 0.326747182430646
30.601567029953 0.636341881041304
30.8015389442444 1.1476894968596
31.0015308856964 1.16835101453195
31.2015249729156 0.515444899158127
31.4015228748322 0.877967507940664
31.6015400886536 0.246236729023184
31.8015170097351 0.495422633817544
32.0015139579773 0.254634629639694
32.2015199661255 0.261279664256526
32.4015200138092 0.247492172028796
32.6015129089355 0.28431372548713
32.8015170097351 0.302381251701484
33.001522064209 0.357487660604221
33.2014420032501 0.375244247671313
33.4015109539032 0.390978978226261
33.6020269393921 0.331124889790151
33.8015909194946 0.341542397412235
34.0015108585358 0.363411897058033
34.2015109062195 0.343507422695741
34.4015119075775 0.358990429094024
34.6015119552612 0.302977980843177
34.8018360137939 0.317072924292148
35.0015280246735 0.307355738175336
35.201523065567 0.361935863934262
35.4015519618988 0.344249210485875
35.6015539169312 0.327943360114687
35.8015189170837 0.355080134834293
36.0015180110931 0.34074905770068
36.2015190124512 0.289108037222201
36.4015839099884 0.317370362987897
36.601557970047 0.242029831466535
36.8015649318695 0.251969248015612
37.0015618801117 0.212362243050122
37.201523065567 0.210795628904828
37.4015300273895 0.248321962688208
37.6015410423279 0.339309233037143
37.801521062851 0.367357662029121
38.0015208721161 0.432837583142225
38.201523065567 0.457270312171214
38.401556968689 0.440047648665127
38.6015639305115 0.421901053050884
38.8015630245209 0.44926932482703
39.0015249252319 0.390445284372365
39.2015390396118 0.336287038343751
39.4015500545502 0.352095217018872
39.6015570163727 0.266637752877932
39.8015298843384 0.256255443084983
40.0015258789062 0.271071294325254
40.2015330791473 0.255767579191682
40.4015309810638 0.224706352657543
40.601567029953 0.271274843729048
40.8015749454498 0.217164657885056
41.0015370845795 0.467700621934878
41.2015578746796 0.441037811479886
41.4015669822693 0.217354682174937
41.6015300750732 0.218955905519619
41.8015439510345 0.22676805863771
42.0015299320221 0.244362612744782
42.2015490531921 0.211150879099531
42.4015378952026 0.228128894349198
42.6015548706055 0.522174499542572
42.8015260696411 0.242505553407473
43.0015239715576 0.258277745784353
43.2015578746796 0.260365110453389
43.4015519618988 0.254341490808176
43.6015570163727 0.225181610305499
43.801647901535 0.290468583190334
44.0015239715576 0.222399123193795
44.2015180587769 0.232074987779382
44.401517868042 0.228100555160556
44.6015560626984 0.235119625544882
44.8015270233154 0.251838478046161
45.001601934433 0.24160036216652
45.201593875885 0.228934983708735
45.4015209674835 0.234193546417396
45.6015260219574 0.246406845306415
45.8015539646149 0.233037755413742
46.0015530586243 0.223101192879912
46.2015240192413 0.265828834497692
46.4015250205994 0.232605995223713
46.6015388965607 0.248536286070294
46.8015260696411 0.276395778020551
47.0016350746155 0.24935298018053
47.2015249729156 0.298532336858187
};
\addlegendentry{TDoA}
\addplot [semithick, green!50.1960784313725!black]
table {%
1.20157790184021 0.515365510818456
1.40152406692505 0.458840335925404
1.60156798362732 0.479173711291266
1.80152893066406 0.393503737776978
2.00156497955322 0.352280282625112
2.2015368938446 0.23722108690427
2.40154695510864 0.245809712897637
2.60147094726562 0.232009352045173
2.80152297019958 0.218147730232177
3.00154590606689 0.238238121665229
3.2015209197998 0.237229628532896
3.40163493156433 0.233729712987784
3.6015989780426 0.225036615607592
3.80152106285095 0.229027995541756
4.00155997276306 0.247374325932178
4.20151686668396 0.266990111177828
4.40157389640808 0.25084724678562
4.60153102874756 0.218696100717417
4.80152606964111 0.253637905797223
5.00152206420898 0.255640632667539
5.20151901245117 0.208815247144459
5.40156388282776 0.23248033766668
5.60152101516724 0.263138767772185
5.80151700973511 0.229987291817748
6.00154590606689 0.230741156213808
6.20160388946533 0.201149239729325
6.40152096748352 0.222269353974724
6.60151791572571 0.207218619307611
6.80139708518982 0.223519095755194
7.00155186653137 0.238027758393051
7.20159292221069 0.221569501076796
7.40152192115784 0.256111616960814
7.60157704353333 0.265271873108907
7.80156207084656 0.278193231187093
8.00152587890625 0.325507709004391
8.20156407356262 0.379975106874493
8.40155601501465 0.350508483597538
8.60144805908203 0.413018115166412
8.80152106285095 0.382478237624187
9.00152087211609 0.442434563985517
9.20155191421509 0.321642136615591
9.4015679359436 0.266427738976945
9.60158085823059 0.240577930303146
9.80155396461487 0.256507972256065
10.0015599727631 0.268945901700362
10.2015218734741 0.242754151212152
10.4016048908234 0.292871191767795
10.6015219688416 0.300788773240987
10.8015179634094 0.245552899622271
11.0015649795532 0.125375455985277
11.2015550136566 0.190128154800579
11.4015688896179 0.330089068473131
11.6015808582306 0.339890863297509
11.8015599250793 0.35798696014726
12.0015599727631 0.337135259372574
12.2015240192413 0.311885981229509
12.4015610218048 0.324811154221478
12.6014440059662 0.344991489647838
12.8015148639679 0.360217110951195
13.0015108585358 0.38369557802887
13.2015199661255 0.39839400593096
13.4015309810638 0.337011619604378
13.6015150547028 0.355785503709625
13.8015089035034 0.368794327670613
14.0015230178833 0.391155819864015
14.2015240192413 0.362522837295506
14.4015629291534 0.380636607211569
14.6016058921814 0.274069289658801
14.8015539646149 0.782476325370448
15.0016119480133 0.855228551616133
15.201623916626 0.329912075166407
15.4015500545502 0.663223987369369
15.601557970047 0.981589332212016
15.80157995224 0.901620844143095
16.0015950202942 0.579556076079072
16.20157289505 1.00257649079162
16.401526927948 1.01843834721538
16.6015539169312 1.05614795130529
16.8015599250793 0.851122176425931
17.0015399456024 0.27910669030708
17.2015309333801 0.293836391472881
17.4015579223633 0.31247251739685
17.6015238761902 0.514613410489002
17.8015310764313 0.358403483584908
18.0015690326691 0.707475477917438
18.2015619277954 0.4756118886674
18.4015538692474 0.304338595386147
18.6015629768372 0.907156627829184
18.8015320301056 0.858868829933425
19.0015320777893 0.275065841908734
19.2015578746796 0.6
19.4015419483185 0.959503095996631
19.6015799045563 0.250975333736395
19.8013379573822 0.58332541891273
20.0015239715576 1.25736576445386
20.201514005661 0.225723824075131
20.4015259742737 0.217185569151638
20.6015300750732 0.362173907938617
20.8015298843384 1.03188138939702
21.0015280246735 0.212148307786265
21.2015490531921 0.239829845642255
21.4015259742737 0.269098157586111
21.6015188694 0.253847062616892
21.8015079498291 0.251147357758005
22.001373052597 0.287739126797027
22.2015209197998 0.267486896081957
22.4015738964081 0.293125907849229
22.601569890976 0.293437498774117
22.8020169734955 0.323354723779014
23.0015769004822 0.346463549498599
23.2015368938446 0.328064566677314
23.4015209674835 0.305895014040673
23.6015510559082 0.313143418105421
23.8015608787537 0.31895937448502
24.0015199184418 0.305347076984073
24.2015700340271 0.335016211128142
24.4015760421753 0.346038401427749
24.6015560626984 0.349187807342155
24.8015530109406 0.34180186432571
25.0015568733215 0.336552565252354
25.2015650272369 0.355292080067647
25.4015760421753 0.356548405810646
25.6015269756317 0.339600689637043
25.8015248775482 0.336375906160016
26.0015149116516 0.339477556701204
26.2015118598938 0.31398084396194
26.4015100002289 0.309675893257009
26.6015138626099 0.334585823127696
26.8015129566193 0.334332484367026
27.0015108585358 0.328775381520162
27.2015180587769 0.329249411055298
27.4015109539032 0.337804398820174
27.6015119552612 0.319172293678056
27.8015110492706 0.318440744650463
28.0015199184418 0.293806110842589
28.2015390396118 0.6
28.401526927948 0.57818405617285
28.6015529632568 0.321626029201906
28.8015120029449 0.6
29.0015449523926 0.356832941185104
29.201514005661 0.389224777348812
29.401524066925 0.242664227174754
29.6015288829803 0.222696277870711
29.8015248775482 0.225128349028353
30.001522064209 0.395165656421056
30.2015218734741 0.272050694914813
30.4016280174255 0.321182212855419
30.601567029953 0.634002698701113
30.8015389442444 1.14562325442657
31.0015308856964 0.926599166668513
31.2015249729156 0.43712625073555
31.4015228748322 0.877488435975554
31.6015400886536 0.241017210032682
31.8015170097351 0.493343098018053
32.0015139579773 0.244741784623762
32.2015199661255 0.254489041498419
32.4015200138092 0.22227890122786
32.6015129089355 0.280035442384649
32.8015170097351 0.298751498736906
33.001522064209 0.352486672403558
33.2014420032501 0.371718634278357
33.4015109539032 0.38524632720418
33.6020269393921 0.325402413458082
33.8015909194946 0.29910398646356
34.0015108585358 0.360897298710271
34.2015109062195 0.341691374360736
34.4015119075775 0.356570380590009
34.6015119552612 0.297711097108646
34.8018360137939 0.312004665092773
35.0015280246735 0.302711651348304
35.201523065567 0.356862943084969
35.4015519618988 0.301096644051275
35.6015539169312 0.324229341838592
35.8015189170837 0.354338160244129
36.0015180110931 0.335309764297082
36.2015190124512 0.287226057711176
36.4015839099884 0.31355987437639
36.601557970047 0.234834901931453
36.8015649318695 0.248065689329342
37.0015618801117 0.208648267143493
37.201523065567 0.223806658858744
37.4015300273895 0.242958353329748
37.6015410423279 0.33557901018259
37.801521062851 0.361696735170579
38.0015208721161 0.427316408375106
38.201523065567 0.455114819487339
38.401556968689 0.437659110491277
38.6015639305115 0.419161076901676
38.8015630245209 0.445686753883505
39.0015249252319 0.387465152722178
39.2015390396118 0.333160732237535
39.4015500545502 0.350633842884432
39.6015570163727 0.264463951793261
39.8015298843384 0.254616872455623
40.0015258789062 0.269250770731125
40.2015330791473 0.252446020866063
40.4015309810638 0.218146628978078
40.601567029953 0.263858408348369
40.8015749454498 0.212663104838943
41.0015370845795 0.391004636074701
41.2015578746796 0.438423921545647
41.4015669822693 0.242540161436615
41.6015300750732 0.252265994647921
41.8015439510345 0.222456756271623
42.0015299320221 0.215026996325626
42.2015490531921 0.202248122582331
42.4015378952026 0.220962558481346
42.6015548706055 0.518389168177187
42.8015260696411 0.234614862289365
43.0015239715576 0.251391193923088
43.2015578746796 0.260491731483597
43.4015519618988 0.246146384205467
43.6015570163727 0.221403953609145
43.801647901535 0.288802898381358
44.0015239715576 0.2193494842282
44.2015180587769 0.208005371374122
44.401517868042 0.221498710448487
44.6015560626984 0.233962162746559
44.8015270233154 0.247027011944564
45.001601934433 0.234262618970407
45.201593875885 0.222603822238611
45.4015209674835 0.22607354481856
45.6015260219574 0.237695997594865
45.8015539646149 0.230575431067103
46.0015530586243 0.214991026993708
46.2015240192413 0.234510552525128
46.4015250205994 0.227136514863594
46.6015388965607 0.214798672951527
46.8015260696411 0.270527967279579
47.0016350746155 0.211981069120537
47.2015249729156 0.329579112746748
};
\addlegendentry{Dynamic}
\end{axis}

\end{tikzpicture}

%% file: tex/box_boxplot.tex
\begin{tikzpicture}

\definecolor{color0}{rgb}{1,0.752941176470588,0.796078431372549}
\definecolor{color1}{rgb}{0.67843137254902,0.847058823529412,0.901960784313726}
\definecolor{color2}{rgb}{0.564705882352941,0.933333333333333,0.564705882352941}

\definecolor{color3}{rgb}{1,0.647058823529412,0}

\begin{axis}[
    height=\figureheight,
    width=\figurewidth,
    axis line style={white},
    tick align=outside,
    tick pos=left,
    x grid style={white!69.0196078431373!black},
    xtick style={color=black},
    y grid style={white!90!black},
    ymajorgrids,
    ytick style={color=black},
    scaled y ticks = false,
    tick align=outside,
    tick pos=left,
    %
    %
    %
    %
    %
    %
xmin=0.5, xmax=3.5,
xtick={1,2,3},
xticklabels={ToF,TDoA,Dynamic},
ylabel={Error (m)},
ymin=-0.0906859787097236, ymax=1.90440555290419,
ytick={-0.25,0,0.25,0.5,0.75,1,1.25,1.5,1.75,2},
yticklabels={0.0,0.00,0.25,0.50,0.75,1.00,1.25,1.50,1.75,}
]
\addplot [black, forget plot]
table {%
1 0.194012210103757
1 0.0933514328323061
};
\addplot [black, forget plot]
table {%
1 0.3
1 0.397237190936524
};
\addplot [black, forget plot]
table {%
0.875 0.0933514328323061
1.125 0.0933514328323061
};
\addplot [black, forget plot]
table {%
0.875 0.397237190936524
1.125 0.397237190936524
};
\addplot [mark=0, mark=+, mark size=1.2, color0, only marks, forget plot, fill opacity=0.2, draw opacity=0.666]
table {%
1 0.0838208978862997
1 0.0788589916495371
1 0.0815683750994743
1 0.0774624856517871
1 0.0735965152634554
1 0.0704423411994635
1 0.0758281510849077
1 0.0808006016986359
1 0.0847476520262247
1 0.0697249630050882
1 0.0850312348632537
1 0.0737872856815186
1 0.0854494332098914
1 0.0675773354196574
1 0.0696209925822729
1 0.0719760154796998
1 0.0759494181090441
1 0.0706072453128572
1 0.0829917699807928
1 0.0868563503792721
1 0.407258068727031
1 0.406492529750086
1 0.44080225964678
1 0.434101526825116
1 0.449119415474546
1 0.497211630466811
1 0.447213043069976
1 0.53489488368818
1 0.488693477797804
1 0.509098462097637
1 0.4616328771228
1 0.420339808319963
1 0.422893119701653
};
\addplot [black, forget plot]
table {%
2 0.234134911150744
2 0.130087951422268
};
\addplot [black, forget plot]
table {%
2 0.513843362013424
2 0.778766169229772
};
\addplot [black, forget plot]
table {%
1.875 0.130087951422268
2.125 0.130087951422268
};
\addplot [black, forget plot]
table {%
1.875 0.778766169229772
2.125 0.778766169229772
};
\addplot [mark=0, mark=+, mark size=1.2, color1, only marks, forget plot, fill opacity=0.2, draw opacity=0.666]
table {%
2 1.03105735833855
2 1.00256727042668
2 1.43270496869659
2 1.61811673257332
2 0.927259153404625
2 1.1526663584315
2 0.941015495607598
2 0.912783995167855
2 0.860394913990668
2 1.81371957419447
2 1.11307917750309
2 1.36421239053213
2 0.814089386648519
2 1.70237260743769
2 0.90091689348798
2 1.70311176355806
2 0.861595858701533
2 0.87538378879905
2 0.891280158845533
2 0.891280158845533
};
\addplot [black, forget plot]
table {%
3 0.227345424578596
3 0.141079184750012
};
\addplot [black, forget plot]
table {%
3 0.390405342629341
3 0.552641873196684
};
\addplot [black, forget plot]
table {%
2.875 0.141079184750012
3.125 0.141079184750012
};
\addplot [black, forget plot]
table {%
2.875 0.552641873196684
3.125 0.552641873196684
};
\addplot [mark=0, mark=+, mark size=1.2, color2, only marks, forget plot, fill opacity=0.2, draw opacity=0.666]
table {%
3 1.02508298202205
3 1.80572152068118
3 0.6
3 0.6
3 0.94749936033612
3 0.951030704279367
3 0.714909653739824
3 0.90827522687083
3 0.864326720835335
3 0.729559020245823
3 0.821519676699147
3 0.881590718365197
3 0.892279989596471
3 0.646218147776799
3 0.627208120552226
3 0.616627985999211
3 0.634649434230949
3 0.622751995965575
3 0.657261360348055
3 0.600313205911154
3 0.587632234134204
3 0.576157847980588
3 0.56931537916859
3 0.589502977803178
3 1.05965122527643
3 1.0326848821635
3 0.737453526378245
3 1.14487797810141
3 0.786028478413317
3 1.07974438465346
3 1.10949383922011
3 1.0823634498095
3 0.983280061570397
3 1.071530902107
3 1.14164744331368
3 1.30388029020873
3 1.15539016381945
3 1.05562519026293
3 1.0298616995003
3 0.961037797378821
3 0.945188647823211
3 1.00979942099294
3 0.941211235357545
3 0.911074853208157
3 0.828866584592437
3 0.794002471730374
3 0.802985476975375
3 0.6296036658077
3 0.805238588406328
3 0.608845241083809
3 0.617833570662242
3 0.616703245436309
};
\addplot [semithick, color0]
table {%
0 0
};
\addplot [semithick, color1]
table {%
0 0
};
\addplot [semithick, color2]
table {%
0 0
};
\path [draw=black, fill=color0]
(axis cs:0.75,0.194012210103757)
--(axis cs:1.25,0.194012210103757)
--(axis cs:1.25,0.3)
--(axis cs:0.75,0.3)
--(axis cs:0.75,0.194012210103757)
--cycle;
\path [draw=black, fill=color1]
(axis cs:1.75,0.234134911150744)
--(axis cs:2.25,0.234134911150744)
--(axis cs:2.25,0.513843362013424)
--(axis cs:1.75,0.513843362013424)
--(axis cs:1.75,0.234134911150744)
--cycle;
\path [draw=black, fill=color2]
(axis cs:2.75,0.227345424578596)
--(axis cs:3.25,0.227345424578596)
--(axis cs:3.25,0.390405342629341)
--(axis cs:2.75,0.390405342629341)
--(axis cs:2.75,0.227345424578596)
--cycle;
\addplot [thick, black, forget plot]
table {%
0.75 0.248163405547624
1.25 0.248163405547624
};
\addplot [thick, black, forget plot]
table {%
1.75 0.313672580080072
2.25 0.313672580080072
};
\addplot [thick, black, forget plot]
table {%
2.75 0.286770847936838
3.25 0.286770847936838
};
\end{axis}

\end{tikzpicture}

%% file: tex/box_boxplot_2.tex
\begin{tikzpicture}

\definecolor{color0}{rgb}{1,0.752941176470588,0.796078431372549}
\definecolor{color1}{rgb}{0.67843137254902,0.847058823529412,0.901960784313726}
\definecolor{color2}{rgb}{0.564705882352941,0.933333333333333,0.564705882352941}

\begin{axis}[
    height=\figureheight,
    width=\figurewidth,
    axis line style={white},
    tick align=outside,
    tick pos=left,
    x grid style={white!69.0196078431373!black},
    xtick style={color=black},
    y grid style={white!90!black},
    ymajorgrids,
    ytick style={color=black},
    scaled y ticks = false,
    legend cell align={left},
    legend style={
      fill opacity=0.8,
      draw opacity=0.666,
      text opacity=1,
      at={(0.03,0.97)},
      anchor=north west,
      draw=white!80!black
    },
    tick align=outside,
    tick pos=left,
    %
    %
    %
    %
    %
    %
xmin=0.5, xmax=3.5,
xtick={1,2,3},
xticklabels={ToF,TDoA,Dynamic},
ylabel={Error (m)},
ymin=-0.0254075068782638, ymax=0.53355764444354,
ytick={-0.1,0,0.1,0.2,0.3,0.4,0.5,0.6},
yticklabels={0.0,0.0,0.1,0.2,0.3,0.4,0.5,}
]
\addplot [black, forget plot]
table {%
1 0.0522668238756055
1 0.00283241228824509
};
\addplot [black, forget plot]
table {%
1 0.133510247079895
1 0.213141368348981
};
\addplot [black, forget plot]
table {%
0.875 0.00283241228824509
1.125 0.00283241228824509
};
\addplot [black, forget plot]
table {%
0.875 0.213141368348981
1.125 0.213141368348981
};
\addplot [mark=0, mark=+, mark size=1.2, color0, only marks, forget plot, fill opacity=0.2, draw opacity=0.666]
table {%
1 0.218118166012873
1 0.23351769886843
1 0.2475212978454
1 0.258925838126102
1 0.267241367826496
1 0.270690566203286
1 0.271292488725528
1 0.268466648353801
1 0.263916928981641
1 0.256370653194361
1 0.2462871086452
1 0.232936848206186
1 0.218167325121993
};
\addplot [black, forget plot]
table {%
2 0.0421752174720608
2 0.000422612921004725
};
\addplot [black, forget plot]
table {%
2 0.173487016822106
2 0.29842360497029
};
\addplot [black, forget plot]
table {%
1.875 0.000422612921004725
2.125 0.000422612921004725
};
\addplot [black, forget plot]
table {%
1.875 0.29842360497029
2.125 0.29842360497029
};
\addplot [mark=0, mark=+, mark size=1.2, color1, only marks, forget plot, fill opacity=0.2, draw opacity=0.666]
table {%
2 0.310859351228497
2 0.325096797989185
2 0.336539160473845
2 0.347248684474937
2 0.354760652905797
2 0.362549696376633
2 0.370020464778333
2 0.375338106016334
2 0.378880352730831
2 0.380882583235019
2 0.381986079823986
2 0.381012134454229
2 0.378204291876558
2 0.375025165322236
2 0.370994611630075
2 0.367626199924643
2 0.364492061667003
2 0.361092481810813
2 0.35845553763114
2 0.355725097049634
2 0.354701199918727
2 0.355405238777892
2 0.357205296913729
2 0.360767800197538
2 0.365238134189308
2 0.3711433205784
2 0.376202761427694
2 0.381280071196249
2 0.385792685736865
2 0.390353282379872
2 0.395719233199459
2 0.402962249505723
2 0.410852235297034
2 0.419170524654682
2 0.427274775748002
2 0.437289870892324
2 0.450505449321106
2 0.463634035167087
2 0.477769879620343
2 0.494055597187478
2 0.508150137565276
2 0.500753262482383
2 0.473412271366704
2 0.446387244075037
2 0.419797117563668
2 0.396972216942883
2 0.375829529477798
2 0.356028504576701
2 0.338022619843172
2 0.321800304465285
2 0.307647312842471
2 0.307647312842471
2 0.307647312842471
2 0.307647312842471
};
\addplot [black, forget plot]
table {%
3 0.0617373457182936
3 0.00293228804392914
};
\addplot [black, forget plot]
table {%
3 0.173533572518017
3 0.283176680150716
};
\addplot [black, forget plot]
table {%
2.875 0.00293228804392914
3.125 0.00293228804392914
};
\addplot [black, forget plot]
table {%
2.875 0.283176680150716
3.125 0.283176680150716
};
\addplot [mark=0, mark=+, mark size=1.2, color2, only marks, forget plot, fill opacity=0.2, draw opacity=0.666]
table {%
3 0.299887823607455
3 0.306095524798874
3 0.305336961361775
3 0.29804651251666
3 0.285605286428949
3 0.301833454948216
3 0.319339572476356
3 0.334590436779
3 0.354248894035364
3 0.370335690440868
3 0.387216943706949
3 0.404642814679031
3 0.419825045242244
3 0.436580131003241
3 0.43024556016816
3 0.418071743814209
3 0.399048832909555
3 0.382606767373617
3 0.359786348653254
3 0.338431464089038
3 0.319807452951931
3 0.297132023400832
};
\addplot [semithick, color0]
table {%
0 0
};
\addplot [semithick, color1]
table {%
0 0
};
\addplot [semithick, color2]
table {%
0 0
};
\path [draw=black, fill=color0]
(axis cs:0.75,0.0522668238756055)
--(axis cs:1.25,0.0522668238756055)
--(axis cs:1.25,0.133510247079895)
--(axis cs:0.75,0.133510247079895)
--(axis cs:0.75,0.0522668238756055)
--cycle;
\path [draw=black, fill=color1]
(axis cs:1.75,0.0421752174720608)
--(axis cs:2.25,0.0421752174720608)
--(axis cs:2.25,0.173487016822106)
--(axis cs:1.75,0.173487016822106)
--(axis cs:1.75,0.0421752174720608)
--cycle;
\path [draw=black, fill=color2]
(axis cs:2.75,0.0617373457182936)
--(axis cs:3.25,0.0617373457182936)
--(axis cs:3.25,0.173533572518017)
--(axis cs:2.75,0.173533572518017)
--(axis cs:2.75,0.0617373457182936)
--cycle;
\addplot [thick, black, forget plot]
table {%
0.75 0.0998320731776465
1.25 0.0998320731776465
};
\addplot [thick, black, forget plot]
table {%
1.75 0.090410248257308
2.25 0.090410248257308
};
\addplot [thick, black, forget plot]
table {%
2.75 0.114307761104489
3.25 0.114307761104489
};
\end{axis}

\end{tikzpicture}

%% file: tex/box_boxplot_3.tex
\begin{tikzpicture}

\definecolor{color0}{rgb}{1,0.752941176470588,0.796078431372549}
\definecolor{color1}{rgb}{0.67843137254902,0.847058823529412,0.901960784313726}
\definecolor{color2}{rgb}{0.564705882352941,0.933333333333333,0.564705882352941}

\begin{axis}[
    height=\figureheight,
    width=\figurewidth,
    axis line style={white},
    tick align=outside,
    tick pos=left,
    x grid style={white!69.0196078431373!black},
    xtick style={color=black},
    y grid style={white!90!black},
    ymajorgrids,
    ytick style={color=black},
    scaled y ticks = false,
    legend cell align={left},
    legend style={
      fill opacity=0.8,
      draw opacity=0.666,
      text opacity=1,
      at={(0.03,0.97)},
      anchor=north west,
      draw=white!80!black
    },
    tick align=outside,
    tick pos=left,
    %
    %
    %
    %
    %
    %
xmin=0.5, xmax=3.5,
xtick={1,2,3},
xticklabels={ToF, TDoA, Dynamic},
ylabel={Error (m)},
ymin=-0.0254075068782638, ymax=0.53355764444354,
ytick={-0.1,0,0.1,0.2,0.3,0.4,0.5,0.6},
yticklabels={0.0,0.0,0.1,0.2,0.3,0.4,0.5,}
]
\addplot [black, forget plot]
table {%
1 0.111381591789731
1 0.100394276609544
};
\addplot [black, forget plot]
table {%
1 0.147135002024952
1 0.182658336304876
};
\addplot [black, forget plot]
table {%
0.875 0.100394276609544
1.125 0.100394276609544
};
\addplot [black, forget plot]
table {%
0.875 0.182658336304876
1.125 0.182658336304876
};
\addplot [mark=0, mark=+, mark size=1.2, color0, only marks, forget plot, fill opacity=0.2, draw opacity=0.666]
table {%
1 0.0661059998136851
1 0.0275255097986387
1 0.0126349815935149
1 0.0510009131972164
1 0.185179236607815
1 0.188785675896006
1 0.192533537639185
1 0.197542165090912
1 0.20365130743368
1 0.213141368348981
};
\addplot [black, forget plot]
table {%
2 0.304698910555834
2 0.24349406611513
};
\addplot [black, forget plot]
table {%
2 0.382937731302206
2 0.450505449321106
};
\addplot [black, forget plot]
table {%
1.875 0.24349406611513
2.125 0.24349406611513
};
\addplot [black, forget plot]
table {%
1.875 0.450505449321106
2.125 0.450505449321106
};
\addplot [mark=0, mark=+, mark size=1.2, color1, only marks, forget plot, fill opacity=0.2, draw opacity=0.666]
table {%
2 0.0832789533831834
2 0.0866536831244653
2 0.0879608011384715
2 0.0907886387544689
2 0.095017577262626
2 0.103293967146732
2 0.113161625515584
2 0.124970775386578
2 0.13703464295994
2 0.150736496287419
2 0.166397884649522
2 0.184441862637151
2 0.204856920725179
2 0.22486589599348
2 0.463634035167087
2 0.477769879620343
2 0.494055597187478
2 0.508150137565276
2 0.500753262482383
2 0.473412271366704
};
\addplot [black, forget plot]
table {%
3 0.0802348424571935
3 0.042889009996225
};
\addplot [black, forget plot]
table {%
3 0.239633047786164
3 0.387216943706949
};
\addplot [black, forget plot]
table {%
2.875 0.042889009996225
3.125 0.042889009996225
};
\addplot [black, forget plot]
table {%
2.875 0.387216943706949
3.125 0.387216943706949
};
\addplot [mark=0, mark=+, mark size=1.2, color2, only marks, forget plot, fill opacity=0.2, draw opacity=0.666]
table {%
3 0.404642814679031
3 0.419825045242244
3 0.436580131003241
3 0.43024556016816
3 0.418071743814209
3 0.399048832909555
};
\addplot [semithick, color0]
table {%
0 0
};
\addplot [semithick, color1]
table {%
0 0
};
\addplot [semithick, color2]
table {%
0 0
};
\path [draw=black, fill=color0]
(axis cs:0.75,0.111381591789731)
--(axis cs:1.25,0.111381591789731)
--(axis cs:1.25,0.147135002024952)
--(axis cs:0.75,0.147135002024952)
--(axis cs:0.75,0.111381591789731)
--cycle;
\path [draw=black, fill=color1]
(axis cs:1.75,0.304698910555834)
--(axis cs:2.25,0.304698910555834)
--(axis cs:2.25,0.382937731302206)
--(axis cs:1.75,0.382937731302206)
--(axis cs:1.75,0.304698910555834)
--cycle;
\path [draw=black, fill=color2]
(axis cs:2.75,0.0802348424571935)
--(axis cs:3.25,0.0802348424571935)
--(axis cs:3.25,0.239633047786164)
--(axis cs:2.75,0.239633047786164)
--(axis cs:2.75,0.0802348424571935)
--cycle;
\addplot [thick, black, forget plot]
table {%
0.75 0.129439609968302
1.25 0.129439609968302
};
\addplot [thick, black, forget plot]
table {%
1.75 0.360930141004176
2.25 0.360930141004176
};
\addplot [thick, black, forget plot]
table {%
2.75 0.102869979476003
3.25 0.102869979476003
};
\end{axis}

\end{tikzpicture}

%% file: sec/06_Conclusion.tex

\section{Conclusion}
\label{sec:conclusion}

We have presented a novel approach to solving the scalability problem in UWB-based relative positioning based on dynamic allocation of active and passive localization roles. With our approach, relative positioning can be estimated within a multi-robot system at a desired frequency, by adaptively selecting the nodes that perform active ToF ranging in order to minimize the expected localization error in passive TDoA nodes. Through a series of experiments using an aerial robot, we demonstrate the applicability of our approach and the improved performance compared to a standard TDoA approach.

In future work, we will look into other optimization methods for the role allocation (e.g, data-driven approaches) while performing larger-scale experiments with multiple mobile ground and aerial robots.